\documentclass[onecolumn,journal]{IEEEtran}

\IEEEoverridecommandlockouts                              

\usepackage{amsmath,amsfonts}
\usepackage{algorithmic}
\usepackage{array}
\usepackage[caption=false,font=normalsize,labelfont=sf,textfont=sf]{subfig}
\usepackage{textcomp} 
\usepackage{stfloats}
\usepackage{url}
\usepackage{verbatim}
\usepackage{graphicx}
\usepackage{amssymb}
\usepackage{multirow}
\usepackage{makecell}
\usepackage{threeparttable}
\usepackage{enumitem}
\usepackage{placeins}
\usepackage{mathtools}
\usepackage{booktabs}
\usepackage{authblk}

\def\BibTeX{{\rm B\kern-.05em{\sc i\kern-.025em b}\kern-.08em
		T\kern-.1667em\lower.7ex\hbox{E}\kern-.125emX}}
\usepackage{balance}

\begin{document}
	
	\title{Dual-IMU State Estimation for \\ Relative Localization of Two Mobile Agents \\ \vspace{0.5cm}
	\large Supplementary Material for  Paper ``Dual-IMU State Estimation for Relative Localization of Two Mobile Agents'' Published at IEEE International Conference on Robotics and Automation 2024 }  
	\author{Wenqian Lai, Ruonan Guo, and Kejian J. Wu}	
	\affil{\small XREAL, Inc. Email: \{wqlai,rnguo,kejian\}@xreal.com}
	\maketitle
	
	\begin{abstract}
		In this paper, we address the problem of relative localization of two mobile agents. Specifically, we consider the Dual-IMU system, where each agent is equipped with one IMU, and employs relative pose observations between them. Previous works, however, typically assumed known ego motion and ignored biases of the IMUs. Instead, we study the most general case of unknown biases for both IMUs. Besides the derivation of dynamic model equations of the proposed system, we focus on the observability analysis, for the observability under general motion and the unobservable directions arising from various special motions. Through numerical simulations, we validate our key observability findings and examine their impact on the estimation accuracy and consistency. Finally, the system is implemented to achieve effective relative localization of an HMD with respect to a vehicle moving in the real world.
	\end{abstract}
	

	\section{Introduction}
%
Relative localization between two mobile agents is a fundamental problem in the field of robotics, autonomous vehicles, and virtual/augmented reality (V/AR). It finds applications in various scenarios such as multi-robot tasking~\cite{co-muti-robot}, tracking of moving targets \cite{ref1}, formation flight of aircraft~\cite{ref2}, and the use of V/AR head-mounted displays (HMD) in moving vehicles~\cite{ref3}. In such cases, rather than focusing solely on the absolute poses of each agent in the global coordinate system, it is of great interest to obtain the relative pose and motion information of one agent with respect to the coordinate system of the other agent.

%
One commonly used sensor for localization is the Inertial Measurement Unit (IMU) due to its small size and light weight. An IMU provides measurements of angular velocity and linear acceleration of the platform, and has been successfully employed for single-agent localization~\cite{INSS}. However, in the context of relative localization, the use of IMU has two limitations: Firstly, it only offers motion information with respect to an inertial frame, whereas the agent serving as the reference frame is often in motion, creating a non-inertial frame. This leads to discrepancies between the measurements from a single IMU and the desired relative motion information. To address this issue, two IMUs can be employed, with one equipped on each agent (Dual-IMU). The other limitation lies in the prolonged use of IMU, which often results in pose drift due to biases and noise. To achieve higher localization accuracy, IMU data can be complemented with additional information, such as global observations (e.g. GPS~\cite{gps_ins_inte}), or relative observations between the agents~\cite{zhou3d} as global information may not always be available.

%
For the state estimation problem of a Dual-IMU system with relative observations, existing works have several limitations. Firstly, there is no prior work that fully addresses all four biases of the two IMUs. However, this is necessary in practice, because IMU biases are often unknown and time-varying, hence need to be estimated online. Moreover, the observability properties of this system remain unknown. Observability~\cite{nolinear}, being a fundamental aspect of state estimation systems, determines whether certain state directions are solvable or not, affecting the system's estimation accuracy and consistency~\cite{consistency}.

%
In this paper, we introduce a general Dual-IMU state estimation system with relative pose observations, for achieving relative localization between two mobile agents. We present the dynamic model equations of the proposed system, which jointly estimate the relative pose, velocity, and all four IMU biases from both agents, the corresponding state transition matrix, and measurements of relative position and orientation. Subsequently, we focus on the observability analysis, addressing both the observability under general motion of the two agents and the unobservable directions resulting from special motion constraints commonly found in practical moving platforms. In summary, the main contributions of this paper are:
\begin{itemize}
	\def\labelenumi{\arabic{enumi}.}
	\item
	We introduce a Dual-IMU estimation system with the most general case of all biases unknown, and derive the continuous and discrete-time system model equations.
	\item
	We prove that the system is locally weakly observable under the condition of general motion.
	\item
	We analytically determine the unobservable directions of the system that arise when two agents undergo various special motions, and identify motion cases where relative pose estimates are either degraded or unaffected.
	\item
	We verify the impact of the unobservable directions on the estimation results through numerical simulations. Furthermore, in a real-world example, the system is implemented to achieve effective relative localization of an HMD with respect to a moving vehicle.
\end{itemize}

\section{Related Work}

A prevalent method to achieve the localization of a single agent in the global coordinate system involves the integration of IMU data with additional observations. Traditional navigation systems combine IMU with GNSS to achieve global localization, e.g. Shin~\cite{gpssingleagent}, Wen et al.~\cite{T-GNSSINS}. Recently, Mourikis et al.~\cite{refmsckf}, Qin et al.~\cite{vins}, and Campos et al.~\cite{orbslam3} have successfully fused IMU with visual information from cameras. Huang~\cite{viosummary} provided a summary of this domain. Another popular approach is to combine IMU with Lidar data, as demonstrated in the works of Shan et al.~\cite{liosam} and Xu et al.~\cite{FAST-LIO2}.

As for multi-robot localization, the problem is often referred to as cooperative localization (CL) or cooperative SLAM (C-SLAM). A distributed multi-robot localization algorithm that utilizes relative observations between robots for update has been proposed by Roumeliotis et al.~\cite{ref10}. Martinelli et al.~\cite{ref13}, and Zhou et al.~\cite{zhou3d} utilized distance and bearing measurements between two agents to analytically compute their relative pose. Kim et al.~\cite{kim2010multiple} introduced multi-robot cooperative mapping based on relative pose graphs. Choudhary et al.~\cite{choudhary2017distributed} reported a distributed cooperative mapping framework.

Meanwhile, there exist previous studies of IMU-based multi-agent systems with relative observations for estimation. Jao et al.~\cite{two-foots} conducted real-world experiments to demonstrate that relative position observations between two IMUs can mitigate the global position divergence of the IMU. However, the system emphasized global pose states. Some studies focused on directly estimating the relative pose states, e.g. the work of Fosbury et al.~\cite{ref7} used line-of-sight measurements between two air vehicles as relative observations. Similarly, Aghili et al.~\cite{icp} employed the ICP algorithm to obtain relative translation and rotation between two agents as observations.

On the other hand, there exist extensive studies in the literature that delve into the observability properties of single-agent localization systems. Martinelli~\cite{Martinelli-vins-unobsw} and Hesch et al.~\cite{VINS-consistency} proved the four unobservable directions of visual-inertial systems under the case of general motion. Martinelli~\cite{obs-s-motion} and Wu et al.~\cite{wu-vins-on-wheels} investigated the additional unobservable directions of such systems under certain special motions. Huang et al. in their work~\cite{FEJ} and~\cite{huang-new}, analyzed the underlying cause of system inconsistency stemming from unobservable directions. Based on their key findings, the first-estimates Jacobian Extended Kalman Filter (FEJ-EKF) was introduced to mitigate the estimation inconsistency issue.
Similarly, for observability analysis of the multi-agent localization problem, Huang et al.~\cite{huang-muti} showed that when considering only relative observations, the absolute pose becomes unobservable. Cristofaro et al.~\cite{mutirobot-obs} used the example of two robots to analyze the observability under both global and relative observations.

A common limitation of the aforementioned works on multi-agent localization, however, is that they assumed known ego motion of the robots, or ignored biases from some of the IMUs.
In contrast, in this work, we consider the most general case of unknown biases from both agents' IMUs, and jointly estimate them. More importantly, we analytically study the observability properties of such system, under both general and special motion scenarios.

	\section{Dual-IMU State Estimation}

\begin{figure}[ht]
		\centering
		\includegraphics[width=3.5in]{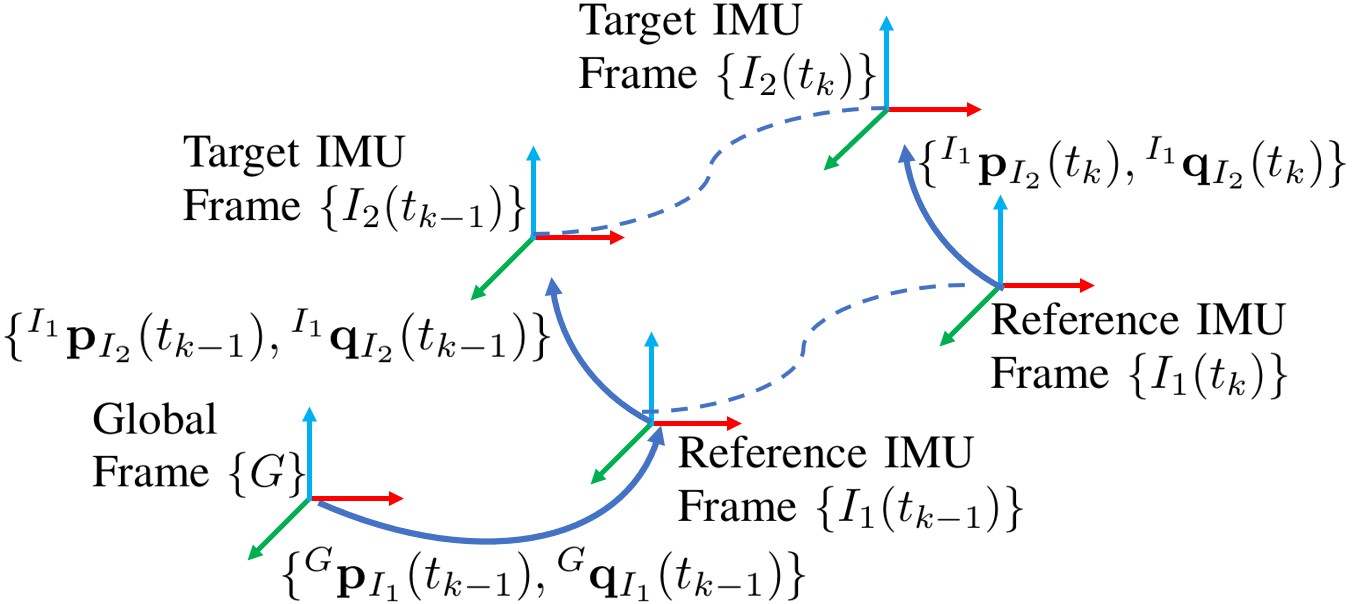}
		\caption{The system comprises a static global coordinate frame \{\(G\)\}, and the moving reference and target IMU coordinate frames \{\( I_1\)\} and \{\( I_2\)\}. \(\prescript{I_1}{}{\mathbf{p}}_{I_2}\) and \(\prescript{I_1}{}{\mathbf{q}_{I_2}}\) are the relative position and orientation of the target IMU with respect to the reference IMU.}
		\label{fig_frame}
	\end{figure}

This section describes the dynamic and measurement models of our Dual-IMU state estimation system. This system comprises two
IMUs, the reference and target IMU, which are rigidly attached to the moving reference and target agent, respectively~(Fig.~\ref{fig_frame}). 
	
	\subsection{System State and Dynamic Model \label{21-system-state-and-propagation-model}}
	
	The system state is a \(22 \times 1\) vector, as:
	\begin{align}
		\mathbf{x} = \begin{bmatrix} \prescript{I_1}{}{\mathbf{p}}^T_{I_2}& \prescript{I_1}{}{\mathbf{v}}^T_{I_2}& \prescript{I_1}{}{\mathbf{q}}^T_{I_2}& \mathbf{b}_{g_1}^T& \mathbf{b}_{g_2}^T& \mathbf{b}_{a_1}^T& \mathbf{b}_{a_2}^T \end{bmatrix}^T \label{base_state}
	\end{align}
	where 
	\(\prescript{I_1}{}{\mathbf{p}}_{I_2}\), \(\prescript{I_1}{}{\mathbf{v}}_{I_2}\), \(\prescript{I_1}{}{\mathbf{q}}_{I_2}\) are the position, linear velocity, and unit (Hamilton) quaternion of orientation of target IMU \{\(I_2\)\} with respect to reference IMU \{\(I_1\)\} (see Fig.~\ref{fig_frame}), \(\mathbf{b}_{g_1}\), \(\mathbf{b}_{a_1}\) 
	are the gyroscope and accelerometer biases of \(I_1\), and
	\(\mathbf{b}_{g_2}\), \(\mathbf{b}_{a_2}\) 
	are the biases of \(I_2\), respectively.
	
	The IMU measurement model of the angular velocity
\(^{I_n}\boldsymbol{\omega}_m\)
and the linear acceleration 
\(^{I_n}\mathbf{a}_m\), \(n=1,2\), 
are:
	\begin{align}
		^{I_n}\boldsymbol{\omega}_m(t) & 
		=\prescript{I_n}{}{\boldsymbol{\omega}}(t) + \mathbf{b}_{g_n}(t) + \mathbf{n}_{g_n}(t)   \label{ww}\\
		^{I_n}\mathbf{a}_m(t)  &
		=\prescript{I_n}{}{\mathbf{a}}(t) + \mathbf{b}_{a_n}(t) + \mathbf{n}_{a_n}(t) \label{aa}
	\end{align}
	where 
	\(^{I_n}\boldsymbol{\omega}(t)\) 
	and  
	\(^{I_n}\mathbf{a}(t)\) are the true angular velocity and linear acceleration in IMU frame,  \(^{I_n}\mathbf{a}(t)=\mathbf{C}^{T}(\prescript{G}{}{\mathbf{q}}_{I_n}(t))(^G\mathbf{a}_{I_n}(t) -\prescript{G}{}{\mathbf{g}})\), with \(\mathbf{C}(\prescript{G}{}{\mathbf{q}}_{I_n}(t))\)  the rotation matrix of IMU frame \{\(I_n\)\}  at time \(t\)  with respect to global frame~\{\(G\)\}, \(^G\mathbf{g}\) the known gravitational acceleration, and \(\mathbf{n}_{g_n}(t)\) and
	\(\mathbf{n}_{a_n}(t)\) are modeled as zero-mean, white Gaussian noise processes.
	
	From \eqref{ww}-\eqref{aa}, we derive continuous-time system model:
	\begin{align} 
	^{I_1}\dot{\mathbf{p}}_{I_2}(t) =&~ \prescript{I_1}{}{\mathbf{v}}_{I_2}(t)   \label{dotp}\\
	^{I_1}\dot{\mathbf{v}}_{I_2}(t) =&~\mathbf{C}(^{I_1}\mathbf{q}_{I_2}(t) )\prescript{I_2}{}{\mathbf{a}}(t) -\prescript{I_1}{}{\mathbf{a}}(t)  \nonumber\\
	&-2\prescript{I_1}{}{\boldsymbol{\omega}}(t) \times \prescript{I_1}{}{\mathbf{v}}_{I_2}(t) - \prescript{I_1}{}{\dot{\boldsymbol{\omega}}}(t) \times\prescript{I_1}{}{\mathbf{p}}_{I_2}(t)   \nonumber  \\
	& -\prescript{I_1}{}{\boldsymbol{\omega}}(t)  \times(\prescript{I_1}{}{\boldsymbol{\omega}}(t) \times \prescript{I_1}{}{\mathbf{p}}_{I_2}(t) )  \label{dotv}  \\
	^{I_1}\dot{\mathbf{q}}_{I_2}(t) 
	=&~\frac{1}{2}(^{I_1}\mathbf{q}_{I_2}(t) \otimes\prescript{I_2}{}{\bar{\boldsymbol{\omega}}}(t) - \prescript{I_1}{}{\bar{\boldsymbol{\omega}}}(t) \otimes\prescript{I_1}{}{\mathbf{q}}_{I_2}(t))  \label{dotq}\\
		\dot{\mathbf{b}}_{g_1}(t) =&~\mathbf{n}_{w{g_1}},\ \ \ \ \ \ \ \dot{\mathbf{b}}_{g_2}(t) = \mathbf{n}_{w{g_2}} \\
		\dot{\mathbf{b}}_{a_1}(t) =&~\mathbf{n}_{w{a_1}},\ \ \ \ \ \ \ \dot{\mathbf{b}}_{a_2}(t) = \mathbf{n}_{w{a_2}} \label{1-2}
	\end{align}
	where
	\(\prescript{I_n}{}{\bar{\boldsymbol{\omega}}} \triangleq \begin{bmatrix} 0 & \prescript{I_n}{}{\boldsymbol{\omega}}^T \end{bmatrix}^T \), \(n=1,2\). 
	The IMU biases are modeled as random walks driven by zero-mean, white Gaussian noise \(\mathbf{n}_{wg_n}\) and \(\mathbf{n}_{wa_n}\). Derivation can be found in Appendix A. 
	
	Linearizing at the current estimates and applying the expectation operator on both sides \(\eqref{dotp}\)-\(\eqref{1-2}\), we obtain the state estimate system equations:
	\begin{align} 
		^{I_1}\dot{\hat{\mathbf{p}}}_{I_2}(t) =&~\prescript{I_1}{}{\hat{\mathbf{v}}}_{I_2}(t) \label{2-1} \\
		^{I_1}\dot{\hat{\mathbf{v}}}_{I_2}(t) =&~\mathbf{C}(^{I_1}\hat{\mathbf{q}}_{I_2}(t) )\prescript{I_2}{}{\hat{\mathbf{a}}}(t) -\prescript{I_1}{}{\hat{\mathbf{a}}}(t) \nonumber \\
		&-2\prescript{I_1}{}{\hat{\boldsymbol{\omega}}}(t) \times\prescript{I_1}{}{\hat{\mathbf{v}}}_{I_2}(t) \nonumber \\
		&-\prescript{I_1}{}{\hat{\boldsymbol{\omega}}}(t)  \times(\prescript{I_1}{}{\hat{\boldsymbol{\omega}}}(t) \times \prescript{I_1}{}{\hat{\mathbf{p}}}_{I_2}(t) ) - \prescript{I_1}{}{\dot{\hat{\boldsymbol{\omega}}}}(t) \times\prescript{I_1}{}{\hat{\mathbf{p}}}_{I_2}(t) \\
		^{I_1}\dot{\hat{\mathbf{q}}}_{I_2}(t) 
		=&~\frac{1}{2}(^{I_1}\hat{\mathbf{q}}_{I_2}(t) \otimes\prescript{I_2}{}{\bar{\hat{\boldsymbol{\omega}}}}(t) - \prescript{I_2}{}{\bar{\hat{\boldsymbol{\omega}}}}(t) \otimes \prescript{I_1}{}{\hat{\mathbf{q}}}_{I_2}(t))  \\
		\dot {\hat{\mathbf{b}}}_{g_1}(t) =&~\mathbf{0}_{3 \times 1},\ \ \ \ \ \ \ \dot {\hat{\mathbf{b}}}_{g_2}(t) = \mathbf{0}_{3 \times 1} \\
		\dot {\hat{\mathbf{b}}}_{a_1}(t) =&~\mathbf{0}_{3 \times 1},\ \ \ \ \ \ \ \dot {\hat{\mathbf{b}}}_{a_2}(t) = \mathbf{0}_{3 \times 1} \label{2-2}
	\end{align}
	where
	\(\prescript{I_n}{}{\hat{\boldsymbol{\omega}}} =\prescript{I_n}{}{\boldsymbol{\omega}}_m - \hat{\mathbf{b}}_{g_n}\)
	and
	\(\prescript{I_n}{}{\hat{\mathbf{a}}} =\prescript{I_n}{}{\mathbf{a}}_m - \hat{\mathbf{b}}_{a_n}\).
	
	\subsection{Continuous-Time Error-State Equation}	
	Error state of \eqref{base_state} is defined as following, a \(21 \times 1\) vector:
	\begin{equation}
		\tilde{\mathbf{x}} = \begin{bmatrix} \prescript{I_1}{}{\tilde{\mathbf{p}}}^T_{I_2}& \prescript{I_1}{}{\tilde{\mathbf{v}}}^T_{I_2}&\prescript{I_1}{}{\delta \boldsymbol{\theta}}^T_{I_2}& \tilde{\mathbf{b}}_{g_1}^T& \tilde{\mathbf{b}}_{g_2}^T& \tilde{\mathbf{b}}_{a_1}^T& \tilde{\mathbf{b}}_{a_2}^T \end{bmatrix}^T \label{base_error_state}
	\end{equation}
	for the position, velocity, and bias states, the additive error model is used, i.e., \(\tilde{\mathbf{x}} = \mathbf{x} - \hat{ \mathbf{x}}\) is the error of the estimate \(\hat{ \mathbf{x}}\) of the state \(\mathbf{x}\). As for the quaternion state \(\mathbf{q}\), the \(3 \times 1\) angle error \(\delta \boldsymbol{\theta}\) is defined as \(
	\delta \mathbf{q} = {\hat{\mathbf{q}}}^{-1} \otimes {\mathbf{q}}
	= \begin{bmatrix}
		1 &  \frac{1}{2}\delta \boldsymbol{\theta}^T
	\end{bmatrix}^T \) following \cite{eskf}.
	
	The linearized continuous-time error-state equation is:
	\begin{equation}
		\dot{\tilde{\mathbf{x}}}(t) = \mathbf{F}(t) \tilde{\mathbf{x}}(t) + \mathbf{G}(t)\mathbf{n}(t)  \label{FGFG}
	\end{equation}
	The continuous-time error-state transition matrix, \(\mathbf{F}\), is:
	\begin{align}
		\mathbf{F} = &\begin{bmatrix}
			\mathbf{0}_3 & \mathbf{I}_3 & \mathbf{0}_3 & \mathbf{0}_3 & \mathbf{0}_3 & \mathbf{0}_3 & \mathbf{0}_3 \\
			-\mathbf{U} & -2[\prescript{I_1}{}{\hat{\boldsymbol{\omega}}} _\times ] & -\hat{\mathbf{C}}[\prescript{I_2}{}{\hat{\mathbf{a}}}_{\times}] 
			& -\mathbf{K} & \mathbf{0}_3 & \mathbf{I}_3 & -\hat{\mathbf{C}}\\
			\mathbf{0}_3 & \mathbf{0}_3 & -[\prescript{I_2}{}{\hat{\boldsymbol{\omega}}}_\times]  & \hat{\mathbf{C}}^T & - \mathbf{I}_3 & \mathbf{0}_3 & \mathbf{0}_3\\
			\mathbf{0}_3 & \mathbf{0}_3 & \mathbf{0}_3 & \mathbf{0}_3 & \mathbf{0}_3 & \mathbf{0}_3 & \mathbf{0}_3 \\
			\mathbf{0}_3 & \mathbf{0}_3 & \mathbf{0}_3 & \mathbf{0}_3 & \mathbf{0}_3 & \mathbf{0}_3 & \mathbf{0}_3\\
			\mathbf{0}_3 & \mathbf{0}_3 & \mathbf{0}_3 & \mathbf{0}_3 & \mathbf{0}_3 & \mathbf{0}_3 & \mathbf{0}_3\\
			\mathbf{0}_3 & \mathbf{0}_3 & \mathbf{0}_3 & \mathbf{0}_3 & \mathbf{0}_3 & \mathbf{0}_3 & \mathbf{0}_3\\
		\end{bmatrix}    \nonumber \\ 
		\mathrm{with}&\ \hat{\mathbf{C}} \triangleq \mathbf{C}(\prescript{I_1}{}{\hat{\mathbf{q}}}_{I_2}), \ \ \ \ \ 	\mathbf{U} \triangleq  [{\prescript{I_1}{}{\hat{\boldsymbol{\omega}}}}_\times ]^2 + [^{I_1}{\dot{\hat{\boldsymbol{\omega}}}}_\times ]  \label{FF} \\
		&\  \mathbf{K} \triangleq [\prescript{I_1}{}{ \hat{\boldsymbol{\omega}}}_\times ][{\prescript{I_1}{}{\hat{\mathbf{p}}}_{I_2}}_\times] + [(\prescript{I_1}{}{ \hat{\boldsymbol{\omega}}} \times \prescript{I_1}{}{\hat{\mathbf{p}}}_{I_2} + 2{\prescript{I_1}{}{\hat{\mathbf{v}}}_{I_2}}   )_\times] \nonumber
	\end{align}
	and \(\mathbf{n}\) is the noise, and \(\mathbf{G}\) is the noise Jacobian matrix. For derivation see Appendix B.
	
	Additional, the continuous time system noise covariance matrix \(	\mathbf{Q}_c\left(t\right) = E\left[\mathbf{n}\left(t+\tau\right)\mathbf{n}^T\left(t\right) \right] \).	
	
	\subsection{Discrete-Time Error-State Equation \label{discrete-time-implementation}}
	
	Given that our IMU measurements are acquired at discrete time steps, with both IMUs synchronized at the same time interval of \(\delta t= t_{k}-t_{k-1}\), the linearized discrete-time error-state equation is:
	\begin{equation}
		\tilde{\mathbf{x}}_{k} = \mathbf{\Phi}_{k,k-1} \tilde{\mathbf{x}}_{k-1} + \mathbf{w}_{k,k-1}  \label{FGFG}
	\end{equation}
	where the \(\mathbf{\Phi}_{k,k-1}\) is the discrete-time error-state transition matrix and the zero-mean, white Gaussian noise \(\mathbf{w}_{k,k-1}\) follow the discrete-time system noise covariance matrix \(\mathbf{Q}_{k,k-1}\):
	\begin{align}
		\mathbf{\Phi}_{k,k-1} &= \mathbf{\Phi} \left(t_k,t_{k-1}\right) = \exp\left[ \int_{t_{k-1}}^{t_k} \mathbf{F}\left(\tau \right) d\tau \right] \label{phi_1}\\
		\mathbf{Q}_{k,k-1} &=\int_{t_{k-1}}^{t_{k}} \mathbf{\Phi}\left(\tau \right)  \mathbf{G}\left(\tau\right)\mathbf{Q}_c\mathbf{G}^T\left(\tau\right) \mathbf{\Phi}^T(\tau) d \tau \label{Q_1}
	\end{align}
	
	The structure of the error-state transition matrix is as follows:
	\begin{equation}
		\mathbf{\Phi} = \left[\begin{array}{ccccccccccc}
			\mathbf{\Phi}_{11} & \mathbf{\Phi}_{12} & \mathbf{\Phi}_{13} & \mathbf{\Phi}_{14} & \mathbf{\Phi}_{15} & \mathbf{\Phi}_{16} &  \mathbf{\Phi}_{17}\\
			\mathbf{\Phi}_{21} & \mathbf{\Phi}_{22} & \mathbf{\Phi}_{23} & \mathbf{\Phi}_{24} & \mathbf{\Phi}_{25} & \mathbf{\Phi}_{26} &  \mathbf{\Phi}_{27} \\
			\mathbf{0}_3 & \mathbf{0}_3 & \mathbf{\Phi}_{33} & \mathbf{\Phi}_{34} & \mathbf{\Phi}_{35} & \mathbf{0}_3 & \mathbf{0}_3 \\
			\mathbf{0}_3 & \mathbf{0}_3 & \mathbf{0}_3 & \mathbf{I}_3 & \mathbf{0}_3 & \mathbf{0}_3 & \mathbf{0}_3  \\
			\mathbf{0}_3 & \mathbf{0}_3 & \mathbf{0}_3 & \mathbf{0}_3 & \mathbf{I}_3 & \mathbf{0}_3 & \mathbf{0}_3 \\
			\mathbf{0}_3 & \mathbf{0}_3 & \mathbf{0}_3 & \mathbf{0}_3 & \mathbf{0}_3 & \mathbf{I}_3 & \mathbf{0}_3  \\
			\mathbf{0}_3 & \mathbf{0}_3 & \mathbf{0}_3 & \mathbf{0}_3 & \mathbf{0}_3 & \mathbf{0}_3 & \mathbf{I}_3 \\ 
		\end{array}\right] \label{Phi}
	\end{equation}
	The detailed derivation and expression of each term in \(\mathbf{\Phi}\) are provided in Appendix C.
         	\subsection{Relative Pose Measurement Model \label{measurment-mode}}
	
	We consider two types of measurements in this paper: relative position (\(\mathbf{dp}\)) and relative orientation (\(\mathbf{dq}\)) measurements.
	
	\subsubsection{Relative Position \( \mathbf{dp}\) \label{231-relative-position}}~{}

	The measurement model:
	\begin{equation}
		\mathbf{dp} = \prescript{{I_1}}{}{\mathbf{p}}_{I_2} + \boldsymbol{\eta}_p
	\end{equation}
	where \(\boldsymbol{\eta}_p\) is zero-mean, white Gaussian noise.
	
	The linearized error model and measurement Jacobian \(\mathbf{H}_p\):
	\begin{equation}
		\tilde{\mathbf{dp}} =\mathbf{dp} - \prescript{I_1}{}{\hat{\mathbf{p}}}_{I_2} \simeq \prescript{{I_1}}{}{\tilde{\mathbf{p}}}_{I_2} +\boldsymbol{\eta}_p  = \mathbf{H}_p \tilde{\mathbf{x}}+\boldsymbol{\eta}_p
	\end{equation}
	\begin{equation}
		\mathbf{H}_p =\begin{bmatrix} \mathbf{I}_{3 \times 3} & \mathbf{0}_{3 \times 18} \end{bmatrix} \label{H1}
	\end{equation}
	where \(\tilde{\mathbf{x}}\) is the error
	states \(\eqref{base_error_state}\).
	
	\subsubsection{Relative Orientation \( \mathbf{dq}\) \label{232-relative-orientation}}~{}
	
	The measurement is an orientation with zero-mean, white Gaussian noise \(\boldsymbol{\eta}_q\) in the reference IMU's coordinate system.
	\begin{equation}
		\mathbf{dq} = \begin{bmatrix} 1 \\ \frac{1}{2}\boldsymbol{\eta}_q \end{bmatrix} \otimes \prescript{{I_1}}{}{\mathbf{q}}_{I_2}
	\end{equation}
	
	The linearized error model and measurement Jacobian \(\mathbf{H}_q\):
	\begin{equation}
		\tilde{\mathbf{dq}} = 2\left[ \mathbf{dq} \otimes\prescript{{I_1}}{}{\hat{\mathbf{q}}}^{-1}_{I_2}\right]_{2:4} \simeq \mathbf{C}\left(\prescript{{I_1}}{}{\hat{\mathbf{q}}}_{I_2}\right) \prescript{{I_1}}{}{\delta\boldsymbol{\theta}}_{I_2} + \boldsymbol{\eta}_q = \mathbf{H}_{q} \tilde{\mathbf{x}}+\boldsymbol{\eta}_q
	\end{equation}
	\begin{equation}
		\mathbf{H}_q = \begin{bmatrix} \mathbf{0}_{3 \times 6} & \mathbf{C}\left(\prescript{{I_1}}{}{  \hat{\mathbf{q}}_{I_2}}\right) & \mathbf{0}_{3 \times 12} \end{bmatrix} \label{H2}
	\end{equation}
	
	We can apply the derived system dynamic and measurement models~\eqref{dotp}-\eqref{H2} to various estimation frameworks, such as the Extended Kalman Filter (EKF)~\cite{refmsckf} or optimization-based methods~\cite{g2o}, to obtain the state estimates.

	\section{Nonlinear Observability Analysis under General Motion \label{3-nolinear-observability-analysis-under-general-motion}}
	
	Next, we conduct an observability analysis of our Dual-IMU system. We employ the nonlinear observability analysis method in~\cite{nolinear} to prove that our system, under the condition of general motion, is locally weakly observable.

	\subsection{System Model Equation \label{31-adjust-continue-time-system}}
	
	For expression simplicity, we transform the state \eqref{base_state} to:
	\begin{equation}
		\mathbf{x}' =\begin{bmatrix} \prescript{I_1}{}{}\mathbf{p}_{I_2}^T & \ \mathbf{v'}^T & \prescript{{I_1}}{}{\mathbf{q}}_{I_2}^T & \mathbf{b}_{g_1}^T& \mathbf{b}_{g_2}^T& \mathbf{b}_{a_1}^T& \mathbf{b}_{a_2}^T \end{bmatrix}^T \label{nolinear-state}
	\end{equation}
	where \(\mathbf{v}' =\prescript{{I_1}}{}{\mathbf{v}}_{I_2} +\prescript{I_1}{}{\boldsymbol{\omega}} \times \prescript{I_1}{}{\mathbf{p}}_{I_2}\). Since \(\mathbf{v}'\) is simply a change of variable as compared to \(\prescript{I_1}{}{\mathbf{v}}_{I_2}\), the subsequent observability results with \eqref{nolinear-state} are equivalent to those with \eqref{base_state}.
	
	Now, we rewrite \eqref{dotp}-\eqref{1-2} in input-affine form as:
		\begin{align} 
		\dot{\mathbf{x}}'  =& \  \mathbf{f}_0 + \mathbf{f}_1 \prescript{{I_1}}{}{\boldsymbol{\omega}}_m + \mathbf{f}_2 \prescript{{I_2}}{}{\boldsymbol{\omega}}_m + \mathbf{f}_3 \prescript{{I_1}}{}{\mathbf{a}}_m + \mathbf{f}_4 \prescript{I_2}{}{\mathbf{a}}_m \nonumber \\
		\mathbf{f}_0 =& \begin{bmatrix} \mathbf{f}_{p}^T & 
			\mathbf{f}_{v}^T & 
			\mathbf{f}_{q}^{T} & 
			\mathbf{0}_{1\times12}  \end{bmatrix}^T \nonumber \\
		\mathbf{f}_1 =& \begin{bmatrix} [{\prescript{I_1}{}{\mathbf{p}}_{I_2}}_\times]^T & [\mathbf{v}'_\times]^T & -\frac{1}{2} [\prescript{I_1}{}{\mathbf{q}}_{I_2}]^{T}_{R(:,2:4)} & \mathbf{0}_{3 \times 12} \end{bmatrix}^T \nonumber \\
		\mathbf{f}_2 =& \begin{bmatrix}
			\mathbf{0}_3 & \mathbf{0}_3 & \frac{1}{2}[\prescript{I_1}{}{\mathbf{q}}_{I_2}]_{L(:,2:4)}^T & \mathbf{0}_{3\times12}
		\end{bmatrix}^T \nonumber \\
		\mathbf{f}_3 =& \begin{bmatrix}
			\mathbf{0}_3 & -\mathbf{I}_3 & \mathbf{0}_{3\times4}& \mathbf{0}_{3\times12}
		\end{bmatrix} ^T \nonumber \\
		\mathbf{f}_4 =& \begin{bmatrix}
			\mathbf{0}_3 & \mathbf{C}^T(\prescript{I_1}{}{\mathbf{q}}_{I_2}) & \mathbf{0}_{3\times4}& \mathbf{0}_{3\times12}
		\end{bmatrix} ^T \label{nolinear-tran}\\
		\mathrm{with}&\    \mathbf{f}_{p} =\mathbf{v}' +  \mathbf{b}_{g_1} \times \prescript{{I_1}}{}{\mathbf{p}}_{I_2} \nonumber \\
		& \ \mathbf{f}_{v} = -\mathbf{C}( \prescript{{I_1}}{}{\mathbf{q}}_{I_2} ) \mathbf{b}_{a_2} + \mathbf{b}_{a_1} + \mathbf{b}_{g_1}\times \mathbf{v}'  \nonumber \\
		& \ \mathbf{f}_q   = -\frac{1}{2}([\prescript{I_1}{}{\mathbf{q}}_{I_2}]_{L} \bar{\mathbf{b}}_{g_2}  - [\prescript{I_1}{}{\mathbf{q}}_{I_2}]_R \bar{\mathbf{b}}_{g_1} ) \nonumber
	\end{align}
	where \([\prescript{I_1}{}{\mathbf{q}}_{I_2}]_L\) and \([ \prescript{I_1}{}{\mathbf{q}}_{I_2}]_R\) are the left and right quaternion-multiplication matrices of \(\prescript{{I_1}}{}{\mathbf{q}}_{I_2}\), \(\bar{\mathbf{b}}_{g_n} \triangleq \begin{bmatrix} 0 & \mathbf{b}_{g_n}^T \end{bmatrix}^T\),~\(n=1,2\). For derivation see Appendix D-1.

	As for system observations, the unit norm of \(\prescript{I_1}{}{\mathbf{q}}_{I_2}\) is treated as a measurement denoted by \(\mathbf{h}_0\).
	Additionally, we include the \(\mathbf{dp}\) measurement in our analysis, denoted by \(\mathbf{h}_1\):
\begin{align}
	\mathbf{h}_0 &= 1 = ||\prescript{{I_1}}{}{\mathbf{q}}_{I_2}||^2_2 \\
	\mathbf{h}_1 &= \mathbf{dp} =\prescript{{I_1}}{}{\mathbf{p}}_{I_2} \label{h11}
\end{align}
	
	\subsection{Nonlinear Observability Analysis under General Motion \label{32-observability-analysis-of-general-motion}}
	
	Following the approach outlined in \cite{nolinear}, when the observation matrix \(\mathcal{O}\) formed by the Lie derivatives elements achieves full column rank, the system is locally weakly observable:
	\begin{equation}
		\mathcal{O} = \begin{bmatrix} \boldsymbol{L}^0h ^T & \boldsymbol{L}_{f_i}^1h ^T & \boldsymbol{L}_{f_if_j}^2h ^T & \boldsymbol{L}_{f_if_jf_k}^3h^T & \cdots \end{bmatrix} ^{T}
	\end{equation}
	
	In order to establish the full column rank of matrix  \(\mathcal{O}\), a subset of Lie derivatives elements can be selected to form a submatrix. If this submatrix attains full column rank, it indicates that the matrix \(\mathcal{O}\) also achieves full column rank. One such subset we found is as follows:
\begin{align}
	& \boldsymbol{L}^0\boldsymbol{h_0},\ 
	\boldsymbol{L}^0\boldsymbol{h_1},\ 
	\boldsymbol{L}^1_{f_0}\boldsymbol{h_0},\  
	\boldsymbol{L}_{f_0}^1\boldsymbol{h_1},\  
	\boldsymbol{L}_{f_0f_0}^2\boldsymbol{h_1}, \  \boldsymbol{L}^2_{f_0f_{41}}\boldsymbol{h_1}, \nonumber\\ & 
	\boldsymbol{L}^2_{f_0f_{42}}\boldsymbol{h_1},\ \boldsymbol{L}^2_{f_0f_{43}} \boldsymbol{h_1}, \  \boldsymbol{L}_{f_0f_0f_{31}}^3\boldsymbol{h_1}, \  \boldsymbol{L}_{f_0f_0f_{32}}^3\boldsymbol{h_1},\ \boldsymbol{L}^3_{{f_0}f_{41}f_0}\boldsymbol{h_1},   \nonumber\\ & \boldsymbol{L}^3_{{f_0}f_{42}f_0}\boldsymbol{h_1},\ \boldsymbol{L}^3_{{f_0}f_{43}f_0}\boldsymbol{h_1}, \    \boldsymbol{L}_{f_0f_0f_{21}}^3\boldsymbol{h_1}, \  \boldsymbol{L}_{f_0f_0f_{22}}^3\boldsymbol{h_1} \label{XX}
\end{align}	
	For detailed proof see Appendix D-2.
	
	In summary, we have shown that the Dual-IMU state estimation system with only \(\mathbf{dp}\) measurement is observable. Hence, the case with both  \(\mathbf{dp}\) and \(\mathbf{dq}\) measurements is also observable, trivially. This means that, if the two agents keep moving generally, the estimation results can be reliable.

	\section{Unobservable Directions under Special Motions \label{unobservable-directions-under-special-motions}}

	In the preceding section, we have established that under general motion conditions, as long as \(\mathbf{dp}\) measurement is present, all states are observable.
	Here, general motion refers to the case when both agents move freely, rotationally and translationally, in the 3D space.
	In practice, however, agents cannot persistently sustain rich 3D motions. For example, in the case of ground-moving robots, they mostly undergo special motion profiles, such as stationary, linear, or planar motions. Also, in the case of a person wearing an HMD and sitting inside a moving vehicle, the HMD may often be stationary or purely rotating with respect to the vehicle. In this section, we present the unobservable directions arising from different types of special motions, and explain their physical meanings, for the case when both  \(\mathbf{dp}\) and \(\mathbf{dq}\) measurements are used, as well as for the case when only  \(\mathbf{dp}\) measurement is used. Following the methodology in~\cite{refobsm,VINS-consistency}, unobservable directions can be found as the null space of the observability matrix of the linearized system.

	\subsection{Unobservable Directions with Both \(\mathbf{dp} \) and \(\mathbf{dq} \) Measurements \label{unobs-dpq}}

	In our estimation system with both \(\mathbf{dp}\) and \(\mathbf{dq}\) measurements, since, these are directly measurements of the relative state \(\prescript{I_1}{}{\mathbf{p}}_{I_2}\) and  \(\prescript{I_1}{}{\mathbf{q}}_{I_2}\), these two directions are observable, regardless of the motion. Additionally, some other quantities corresponding to the first or second-order derivatives of measurements, are also informative. For example, the relative velocity (\(\prescript{I_1}{}{\mathbf{v}}_{I_2}\)), which corresponds to the first-order derivatives of \(\mathbf{dp}\) is observable. Also, the relative gyroscope bias \(\mathbf{C}\left(\prescript{I_1}{}{\mathbf{q}}_{I_2}\right) \mathbf{b}_{g_2}- \mathbf{b}_{g_1}\), which equals to the first-order derivative of \(\mathbf{dq}\) minus terms consisting of known measurements and \(\prescript{I_1}{}{\mathbf{q}}_{I_2}\), hence, it is observable. Similarly, the system also has information on the relative accelerometer bias \(\mathbf{C}\left(\prescript{I_1}{}{\mathbf{q}}_{I_2}\right) \mathbf{b}_{a_2}- \mathbf{b}_{a_1}\), which corresponds to a part of the second-order derivative of \(\mathbf{dp}\). We use \(\mathbf{b}_{g-}\) and \(\mathbf{b}_{a-}\) to denote the directions of ``relative'' gyroscope and accelerometer bias:
	\begin{align}
			\left[ {\mathbf{b}_{a-}} \ | \ {\mathbf{b}_{g-}} \right]_{\left[21\times 3,\  21 \times 3\right]}   &= \begin{bmatrix}
				\mathbf{0}_3  & \mathbf{0}_3   \\
				\mathbf{0}_3 & \mathbf{0}_3 \\
				\mathbf{0}_3 &  \mathbf{0}_3 \\
				\mathbf{0}_3 &  -\mathbf{I}_3    \\
				\mathbf{0}_3   & \mathbf{C}^{T}( \prescript{{I_1}}{}{\mathbf{q}}_{I_2} \left(t\right)) \\
				-\mathbf{I}_3 & \mathbf{0}_3 \\
				\mathbf{C}^{T}( \prescript{{I_1}}{}{\mathbf{q}}_{I_2} \left(t\right))& \mathbf{0}_3  \\
			\end{bmatrix} \label{bag-}
		\end{align}

	Meanwhile, the remaining directions of the state vector \eqref{base_error_state} form potential unobservable directions of the system. These directions correspond to the ``composite'' gyroscope and accelerometer bias, denoted as \(\mathbf{b}_{g+}\) and \(\mathbf{b}_{a+}\):  
	\begin{align}
		&\left[ {\mathbf{b}_{a+}} \ | \ {\mathbf{b}_{g+}} \right]_{\left[21\times 3,\  21 \times 3\right]}  
= \begin{bmatrix}
	\mathbf{0}_3  & \mathbf{0}_3   \\
	\mathbf{0}_3  & \mathbf{0}_3 \\
	\mathbf{0}_3  &  \mathbf{0}_3 \\
	\mathbf{0}_3  & \mathbf{I}_3    \\
	\mathbf{0}_3  & \mathbf{C}^T( \prescript{I_1}{}{\mathbf{q}}_{I_2} (t_0)) \\
	\mathbf{I}_3  & \mathbf{0}_3 \\
	\mathbf{C}^T(\prescript{I_1}{}{\mathbf{q}}_{I_2} (t_0))& \mathbf{0}_3 \\
\end{bmatrix} \label{Npq} 
	\end{align}
	where \(t_0\) represents the initial moment.
	
	For all types of special motions we consider here, the unobservable directions of the system with both \(\mathbf{dp} \) and \(\mathbf{dq} \) measurements are summarized in Table~(\ref{table-dpq}). See Appendix E-1 for the proof. 
	Hence, for each type of motion of the two IMUs, we can query Table~(\ref{table-dpq}) to obtain the corresponding unobservable state directions. Note that, the composite gyroscope and accelerometer biases, \(\mathbf{b}_{g+}\) and \(\mathbf{b}_{a+}\), form the union of all unobservable directions in Table~(\ref{table-dpq}). In what follows, we will explain the physical meaning of each direction under its corresponding special motion types.	

\begin{itemize}
	\item {The unobservable direction \( {\mathbf{b}_{a+}} \):}
	
\begin{itemize}
	
	\item[-] \textit{Physical Quantity:} \( {\mathbf{b}_{a+}} \) is 3 dof of the composite accelerometer bias.
	
	\item[-] \textit{Sufficient Conditions:} There are no relative angular velocity between the two IMUs (\(\prescript{I_1}{}{\boldsymbol{\omega}}_{I_2} = \mathbf{0}_{3\times1}\)), e.g., two robots simultaneously under linear motion.

	\item[-] \textit{Interpretation:} There are no global measurements for the two IMUs. Hence, we cannot directly obtain information about the composite accelerometer bias from measurements.
\end{itemize}

	\item {The unobservable direction \( {\mathbf{b}_{a+}^{\boldsymbol{\omega}}}\):}
	\begin{itemize}
		
	 \item[-] \textit{Physical Quantity:} \(\mathbf{b}_{a+}^{\boldsymbol{\omega}}  = \mathbf{b}_{a+} \cdot \prescript{I_1}{}{\boldsymbol{\omega}}_{I_2}\) is 1 dof of the composite accelerometer bias along \( \prescript{I_1}{}{}{\boldsymbol{\omega}}_{I_2} \) direction. 
	
	\item[-] \textit{Sufficient Conditions:} 			The relative angular velocity  \(\prescript{I_1}{}{}{\boldsymbol{\omega}}_{I_2} \) direction remains constant, i.e., \(\prescript{I_1}{}{\boldsymbol{\omega}}_{I_2}(t) = k(t) \cdot \boldsymbol{\xi}  \), where, \(\boldsymbol{\xi}\) is unit vector of constant direction of \(\prescript{I_1}{}{\boldsymbol{\omega}}_{I_2}\), e.g., two ground robots move in the same plane.
	
	\item[-] \textit{Interpretation:} Same as \( {\mathbf{b}_{a+}} \), there are no global measurements. Furthermore, the time-varying relative orientation leads to a change in the direction of \(\mathbf{b}_{a-}\) (see the \(\mathbf{C}\left(\prescript{I_1}{}{}{\mathbf{q}}_{I_2}\left(t\right)\right)\) term in \eqref{bag-}), where some new directions of \(\mathbf{b}_{a-}\) is including in the subspace of \(\mathbf{b}_{a+}\). Hence, some directions of \(\mathbf{b}_{a+}\) becomes informative. But, in this case, the relative angular velocity direction remains constant, and \(\mathbf{b}_{a+}^{\boldsymbol{\omega}} \) always remains orthogonal to \(\mathbf{b}_{a-}\). Therefore, there is no information transmission from \(\mathbf{b}_{a-}\) to \(\mathbf{b}_{a+}\) and \(\mathbf{b}_{a+}^{\boldsymbol{\omega}}\) remains unobservable.
\end{itemize}

		\item {The unobservable direction  \({\mathbf{b}_{g+}}\):}
\begin{itemize}
		\item[-]\textit{Physical Quantity:} \({\mathbf{b}_{g+}}\) is 3 dof of the composite gyroscope bias.
		\item[-]\textit{Sufficient Conditions:}
		\begin{enumerate}[left=26pt]
			\item [c1.1]
			\(\prescript{I_1}{}{}{\boldsymbol{\omega}}_{I_2} = \mathbf{0}_{3 \times 1} \);
			\item [and c1.2]
			\(\prescript{I_1}{}{}{\mathbf{p}}_{I_2} = \mathbf{0}_{3 \times 1} \) or \( \prescript{I_1}{}{}{\boldsymbol{\omega}} = \mathbf{0}_{3 \times 1} \);  \label{bgg-2}
			\item [and c1.3]
			\(\prescript{I_1}{}{}{\mathbf{v}}_{I_2}= \mathbf{0}_{3 \times 1}\); \label{bgg-3}
		\end{enumerate}
		e.g., two IMUs that overlap and are at rest relative to each other.
		
		\item[-]\textit{Interpretation:} Same as \( {\mathbf{b}_{a+}} \), we cannot directly obtain information about the composite gyroscope, due to the lack of global measurement for two IMUs. In addition, as a non-inertial system, the absolute gyroscope biases error \(\tilde{\mathbf{b}}_{g_1}\) of the reference IMU is related to the observable quantity, which is the relative velocity error \(\prescript{I_1}{}{\tilde{\mathbf{v}}}_{I_2}\). \(\mathbf{F}_{24} = \mathbf{K}\) is the Jacobian of \(\prescript{I_1}{}{\tilde{\mathbf{v}}}_{I_2}\) with respect to  \(\tilde{\mathbf{b}}_{g_1}\) \eqref{FF}. But, in this case, conditions c1.2 and c1.3 results in that the \(\mathbf{K} = \mathbf{0}_3\). Hence, information cannot be transmitted from relative velocity to the gyroscope biases.
\end{itemize}

	\item {The unobservable direction  \({\mathbf{b}_{g+}^{\boldsymbol{\omega}}} \):}
\begin{itemize}	
	\item[-]\textit{Physical Quantity:} \({\mathbf{b}_{g+}^{\boldsymbol{\omega}}} = \mathbf{b}_{g+}\cdot \prescript{I_1}{}{\boldsymbol{\omega}}_{I_2} \) is 1 dof of the composite gyroscope bias along \( \prescript{I_1}{}{}{\boldsymbol{\omega}}_{I_2} \) direction.
	\item[-]\textit{Sufficient Conditions:}
	
	\begin{enumerate}[left=26pt]
		\item [c2.1] \label{bgw-1}
			The relative angular velocity  \(\prescript{I_1}{}{}{\boldsymbol{\omega}}_{I_2} \) direction remains constant, i.e., \(\prescript{I_1}{}{\boldsymbol{\omega}}_{I_2}(t) = k(t) \cdot \boldsymbol{\xi}  \), where, \(\boldsymbol{\xi}\) is unit vector of constant direction of \(\prescript{I_1}{}{\boldsymbol{\omega}}_{I_2}\);
		\item [and c2.2]
		\(\prescript{I_1}{}{}{\mathbf{p}}_{I_2} \parallel  \prescript{I_1}{}{}{\boldsymbol{\omega}} \) or  \(\prescript{I_1}{}{}{\mathbf{p}}_{I_2} = \mathbf{0}_{3 \times 1} \) or \( \prescript{I_1}{}{}{\boldsymbol{\omega}} = \mathbf{0}_{3 \times 1}\);
		\item [and c2.3]
		\(\prescript{I_1}{}{}{\mathbf{p}}_{I_2} \parallel  \prescript{I_1}{}{}{\boldsymbol{\omega}}_{I_2} \) or  \(\prescript{I_1}{}{}{\mathbf{p}}_{I_2} = \mathbf{0}_{3 \times 1} \) or \( \prescript{I_1}{}{}{\boldsymbol{\omega}} = \mathbf{0}_{3 \times 1}\);
		\item [and c2.4]
		\(\prescript{I_1}{}{}{\mathbf{v}}_{I_2} \parallel  \prescript{I_1}{}{}{\boldsymbol{\omega}}_{I_2} \) or  \(\prescript{I_1}{}{}{\mathbf{v}}_{I_2}= \mathbf{0}_{3 \times 1} \);
	\end{enumerate}
	e.g., an HMD rotating around the direction of gravity inside a stationary vehicle.

	\item[-]\textit{Interpretation:} There are also no global measurements for this composite. And, the condition c2.1 leads to information that cannot be transmitted from \(\mathbf{b}_{g-}\) to \({\mathbf{b}_{g+}^{\boldsymbol{\omega}}} \), due to the same reasons as in \( {\mathbf{b}_{a+}^{\boldsymbol{\omega}}} \). Furthermore, the conditions c2.2, c2.3 and c2.4 results in the matrix \(\mathbf{K}\), which is the Jacobian of \(\prescript{I_1}{}{\tilde{\mathbf{v}}}_{I_2}\) with respect to  \(\tilde{\mathbf{b}}_{g_1}\), loss a rank. And the nullspace of \(\mathbf{K}\) is \(\prescript{I_1}{}{}{\boldsymbol{\omega}}_{I_2}\). Hence, the information of relative velocity cannot be transmitted gyroscope biases in \({\mathbf{b}_{g+}^{\boldsymbol{\omega}}} \) direction.
	
\end{itemize}
	
	\item {The unobservable direction  \({\mathbf{b}_{g+}^{\boldsymbol{\alpha}}} \)}
	\begin{itemize}
	\item[-]\textit{Physical Quantity:}  \({\mathbf{b}_{g+}^{\boldsymbol{\alpha}}} = \mathbf{b}_{g+} \cdot \boldsymbol{\alpha} \) is 1 dof of the composite gyroscope bias along \(\boldsymbol{\alpha}\) direction, where \(\boldsymbol{\alpha}\) is defined in Table~(\ref{table-dpq}).
	
	\item[-]\textit{Sufficient Conditions:}
	\begin{enumerate}[itemindent=26pt]
		\item [c3.1]
		\(\prescript{I_1}{}{}{\boldsymbol{\omega}}_{I_2} \parallel  \boldsymbol{\alpha} \);
		\item [and c3.2]
		\(\prescript{I_1}{}{}{\mathbf{p}}_{I_2} \parallel  \prescript{I_1}{}{}{\boldsymbol{\omega}} \) or  \(\prescript{I_1}{}{}{\mathbf{p}}_{I_2} =  \mathbf{0}_{3\times1} \) or \( \prescript{I_1}{}{}{\boldsymbol{\omega}} =   \mathbf{0}_{3\times1} \);
		\item [and c3.3]
		\(\prescript{I_1}{}{}{\mathbf{p}}_{I_2} \parallel  \boldsymbol{\alpha}\) or  \(\prescript{I_1}{}{}{\mathbf{p}}_{I_2} =  \mathbf{0}_{3\times1} \) or \( \prescript{I_1}{}{}{\boldsymbol{\omega}} =   \mathbf{0}_{3\times1} \);
		\item [and c3.4]
		\(\prescript{I_1}{}{}{\mathbf{v}}_{I_2} \parallel  \boldsymbol{\alpha} \) or  \(\prescript{I_1}{}{}{\mathbf{v}}_{I_2}= \mathbf{0}_{3 \times 1} \); 
	\end{enumerate}
	e.g., two elevators, which only move along the direction of gravity.
	
	\item[-]\textit{Interpretation:} Same as \({\mathbf{b}_{g+}^{\boldsymbol{\omega}}} \), the condition c3.1 leads to information that cannot be transmitted from \(\mathbf{b}_{g-}\) to \({\mathbf{b}_{g+}^{\boldsymbol{\alpha}}} \). Furthermore, the conditions c3.2, c3.3 and c3.4 eliminate the correlation between relative velocity and \(\mathbf{b}_{g_1}\) along \(\boldsymbol{\alpha}\) direction. Hence, \({\mathbf{b}_{g+}^{\boldsymbol{\alpha}}} \) direction remains unobservable.
\end{itemize}

\end{itemize}

	\FloatBarrier

\begin{table*}[!t]
	\scriptsize
	\caption{Unobservable Directions with Both \(\mathbf{dp}\) and \(\mathbf{dq}\) Measurements}
	\vspace{-0.6cm}
	\center{
		\renewcommand{\arraystretch}{1.3}
		\resizebox{1.0\textwidth}{!}{
			\begin{threeparttable}
				\begin{tabular}{|p{1.0cm}<{\centering}|p{1.0cm}<{\centering}|p{2.0cm}<{\centering}|p{0.25cm}<{\centering}|p{1.0cm}<{\centering}|p{1.2cm}<{\centering}|p{1.6cm}<{\centering}|p{1.1cm}<{\centering}|p{1.3cm}<{\centering}|c|c|}
					\hline
					\multicolumn{3}{|c|}{\multirow{3}{*}{\makecell*[c]{Motion Constraints \\ of Target IMU w.r.t. \\ Reference IMU\(^{\ddagger}\)}}} &
					&
					\multicolumn{7}{c|}{Motion Constraints of Reference IMU\(^{\dag}\)} \\ \cline{5-11} 
					\multicolumn{3}{|l|}{} &  & \multicolumn{3}{c|}{\makecell*[c]{\(\prescript{I_1}{}{\boldsymbol{\omega}} = \mathbf{0}\)}} & \multicolumn{2}{c|}{\(\prescript{I_1}{}{\boldsymbol{\omega}} \parallel \boldsymbol{\alpha} \)} & \multicolumn{1}{c|}{\(\prescript{I_1}{}{\boldsymbol{\omega}} \) dir. const. \(\prescript{I_1}{}{\boldsymbol{\omega}} \nparallel \boldsymbol{\alpha} \)} & \multicolumn{1}{c|}{\(\prescript{I_1}{}{\boldsymbol{\omega}} \) unconstrained } 					 \\ \cline{5-11} 
					\multicolumn{3}{|l|}{} & &  \(\prescript{I_1}{}{\mathbf{a}}\) = \( \boldsymbol{\alpha} \) 
					& \(\prescript{I_1}{}{\mathbf{v}} \parallel \boldsymbol{\alpha} \)  
					& \makecell*[c]{ \(\prescript{I_1}{}{\mathbf{v}}\)  dir. const. \\  \(\prescript{I_1}{}{\mathbf{v}} \nparallel \boldsymbol{\alpha} \)  } 
					&   \(\prescript{I_1}{}{\mathbf{a}}\) = \( \boldsymbol{\alpha} \) 
					& \(\prescript{I_1}{}{\mathbf{v}} \perp  \prescript{I_1}{}{\boldsymbol{\omega}}\) 
					& \(\prescript{I_1}{}{\mathbf{v}} \perp  \prescript{I_1}{}{\boldsymbol{\omega}}\)   
					& \makecell[c]{ \(\prescript{I_1}{}{\mathbf{v}}\) unconstrained}
					\\ \hline
					
					\(\prescript{I_1}{}{\mathbf{p}}_{I_2}\) & \(\prescript{I_1}{}{\mathbf{v}}_{I_2}\) & \(\prescript{I_1}{}{\boldsymbol{\omega}}_{I_2}\) &
					&  I & II & III & IV & V & VI &  VII
					\\ 	\hline 		
					\multirow{3}{*}{\(\mathbf{p}=\mathbf{0}\)} &
					\multirow{3}{*}{\(\mathbf{v}=\mathbf{0}\)} &
					\(\boldsymbol{\omega}=\mathbf{0}\)&
					A &
					\multicolumn{3}{c|}{\({\mathbf{b}_{a+}}\),\({\mathbf{b}_{g+}}\)} &
					\multicolumn{2}{c|}{\({\mathbf{b}_{a+}}\),\({\mathbf{b}_{g+}}\)} &
					\multicolumn{1}{c|}{\({\mathbf{b}_{a+}}\),\({\mathbf{b}_{g+}}\)} &
					\multicolumn{1}{c|}{\({\mathbf{b}_{a+}}\),\({\mathbf{b}_{g+}}\)} 
					\\ \cline{3-11} 
					& & \multicolumn{1}{c|}{\(\boldsymbol{\omega}_x=\mathbf{0}\) \(\boldsymbol{\omega}_y=\mathbf{0}\)} &B
					&
					\multicolumn{3}{c|}{\(\mathbf{b}_{a+}^{\boldsymbol{\omega}}\),\({\mathbf{b}_{g+}^{\boldsymbol{\omega}}}\)} &
					\multicolumn{2}{c|}{\(\mathbf{b}_{a+}^{\boldsymbol{\omega}}\),\({\mathbf{b}_{g+}^{\boldsymbol{\omega}}}\)} &		
					\multicolumn{1}{c|}{\(\mathbf{b}_{a+}^{\boldsymbol{\omega}}\),\({\mathbf{b}_{g+}^{\boldsymbol{\omega}}}\)} &
					\multicolumn{1}{c|}{\(\mathbf{b}_{a+}^{\boldsymbol{\omega}}\),\({\mathbf{b}_{g+}^{\boldsymbol{\omega}}}\)} 
					\\ \cline{3-11} 
					& & Unconstrained    & C &
					\multicolumn{3}{c|}{0} & 
					\multicolumn{2}{c|}{0} & 
					\multicolumn{1}{c|}{0} &
					\multicolumn{1}{c|}{0} 
					\\ 		\hline 
					{\multirow{6}{*}{\makecell*[c]{\(\mathbf{p}_x=\mathbf{0}\) \\ \(\mathbf{p}_y=\mathbf{0}\) \\ \(\mathbf{p}_z \neq \mathbf{0}\)}}} &
					{\multirow{3}{*}{\(\mathbf{v}=\mathbf{0}\)}} & \(\boldsymbol{\omega}=\mathbf{0}\)&
					D &
					\multicolumn{3}{c|}{\({\mathbf{b}_{a+}}\),\({\mathbf{b}_{g+}}\)} &
					\multicolumn{2}{c|}{\({\mathbf{b}_{a+}}\),\(\mathbf{b}_{g+}^{\boldsymbol{\alpha}}\)} &
					\multicolumn{1}{c|}{\({\mathbf{b}_{a+}}\)} &
					\multicolumn{1}{c|}{\({\mathbf{b}_{a+}}\)}
					\\ \cline{3-11} 
					& & \(\boldsymbol{\omega}_x=\mathbf{0}\) \(\boldsymbol{\omega}_y=\mathbf{0}\) & E &
					\multicolumn{3}{c|}{\(\mathbf{b}_{a+}^{\boldsymbol{\omega}}\),\({\mathbf{b}_{g+} ^{\boldsymbol{\omega}}}\)} &
					\multicolumn{2}{c|}{\(\mathbf{b}_{a+}^{\boldsymbol{\omega}}\),\({\mathbf{b}_{g+}^{\boldsymbol{\omega}}}\)} &
					\multicolumn{1}{c|}{\(\mathbf{b}_{a+}^{\boldsymbol{\omega}} \)} &
					\multicolumn{1}{c|}{\(\mathbf{b}_{a+}^{\boldsymbol{\omega}} \)} 
					\\ \cline{3-11} 
					& & Unconstrained    & F &
					\multicolumn{3}{c|}{0} & 
					\multicolumn{2}{c|}{0} & 
					\multicolumn{1}{c|}{0} &
					\multicolumn{1}{c|}{0} 
					\\ 
					\cline{2-11}  &
					\multirow{3}{*}{\makecell*[c]{\(\mathbf{v}_x=\mathbf{0}\) \\ \(\mathbf{v}_y=\mathbf{0}\)}} & \(\boldsymbol{\omega}=\mathbf{0}\)&
					G & 
					\multicolumn{3}{c|}{\({\mathbf{b}_{a+}}\),\(\mathbf{b}_{g+}^{\boldsymbol{\alpha}}\)} &
					\multicolumn{2}{c|}{\({\mathbf{b}_{a+}}\),\(\mathbf{b}_{g+}^{\boldsymbol{\alpha}}\)} &
					\multicolumn{1}{c|}{\({\mathbf{b}_{a+}}\)} &
					\multicolumn{1}{c|}{\({\mathbf{b}_{a+}}\)} 
					\\ \cline{3-11} 
					& & \(\boldsymbol{\omega}_x=\mathbf{0}\) \(\boldsymbol{\omega}_y=\mathbf{0}\) & H &
					\multicolumn{3}{c|}{\(\mathbf{b}_{a+}^{\boldsymbol{\omega}}\),\(\mathbf{b}_{g+}^{\boldsymbol{\omega}}\)} &
					\multicolumn{2}{c|}{\(\mathbf{b}_{a+}^{\boldsymbol{\omega}} \),\(\mathbf{b}_{g+}^{\boldsymbol{\omega}}\)}  &
					\multicolumn{1}{c|}{\(\mathbf{b}_{a+}^{\boldsymbol{\omega}} \)}  &
					\multicolumn{1}{c|}{\(\mathbf{b}_{a+}^{\boldsymbol{\omega}} \)} 
					\\ \cline{3-11} 
					& & Unconstrained    & J &
					\multicolumn{3}{c|}{0} & 
					\multicolumn{2}{c|}{0} & 
					\multicolumn{1}{c|}{0} &
					\multicolumn{1}{c|}{0} 
					\\  \hline 
					\multirow{9}{*}{\makecell*[c]{\(\mathbf{p}_x \neq \mathbf{0}\) \\ \(\mathbf{p}_y \neq \mathbf{0}\) \\ \(\mathbf{p}_z \neq \mathbf{0}\) }} &
					\multirow{3}{*}{\(\mathbf{v}=\mathbf{0}\)} &
					\(\boldsymbol{\omega}=\mathbf{0}\)
					& K &
					\multicolumn{3}{c|}{\({\mathbf{b}_{a+}}\),\({\mathbf{b}_{g+}}\)} & 
					\multicolumn{2}{c|}{\({\mathbf{b}_{a+}}\)} & 
					\multicolumn{1}{c|}{\({\mathbf{b}_{a+}}\)} &
					\multicolumn{1}{c|}{\({\mathbf{b}_{a+}}\)} 
					\\ \cline{3-11} 
					& & \(\boldsymbol{\omega}_x=\mathbf{0}\) \(\boldsymbol{\omega}_y=\mathbf{0}\) & L &
					\multicolumn{3}{c|}{\(\mathbf{b}_{a+}^{\boldsymbol{\omega}}\),\( \mathbf{b}_{g+}^{\boldsymbol{\omega}} \) } & 
					\multicolumn{2}{c|}{\(\mathbf{b}_{a+}^{\boldsymbol{\omega}} \)} & 
					\multicolumn{1}{c|}{\(\mathbf{b}_{a+}^{\boldsymbol{\omega}} \)} &
					\multicolumn{1}{c|}{\(\mathbf{b}_{a+}^{\boldsymbol{\omega}} \)} 
					\\ \cline{3-11} 
					& & Unconstrained    & M &
					\multicolumn{3}{c|}{0} & 
					\multicolumn{2}{c|}{0} & 
					\multicolumn{1}{c|}{0} &
					\multicolumn{1}{c|}{0} 
					\\ 
					\cline{2-11} 
					& \multirow{3}{*}{\makecell*[c]{\(\mathbf{v}_x=\mathbf{0}\) \\ \(\mathbf{v}_y=\mathbf{0}\)}} & \(\boldsymbol{\omega}=\mathbf{0}\)&
					N & 
					\multicolumn{3}{c|}{\({\mathbf{b}_{a+}}\),\(\mathbf{b}_{g+}^{\boldsymbol{\alpha}}\)} & 
					\multicolumn{2}{c|}{\({\mathbf{b}_{a+}}\)} & 
					\multicolumn{1}{c|}{\({\mathbf{b}_{a+}}\)} &
					\multicolumn{1}{c|}{\({\mathbf{b}_{a+}}\)} 
					\\ \cline{3-11} 
					& & \(\boldsymbol{\omega}_x=\mathbf{0}\) \(\boldsymbol{\omega}_y=\mathbf{0}\) & O &
					\multicolumn{3}{c|}{\(\mathbf{b}_{a+}^{\boldsymbol{\omega}}\),\(\mathbf{b}_{g+}^{\boldsymbol{\omega}}\)} & 
					\multicolumn{2}{c|}{\(\mathbf{b}_{a+}^{\boldsymbol{\omega}} \)} & 
					\multicolumn{1}{c|}{\(\mathbf{b}_{a+}^{\boldsymbol{\omega}} \)} &
					\multicolumn{1}{c|}{\(\mathbf{b}_{a+}^{\boldsymbol{\omega}} \)} 
					\\ \cline{3-11} 
					& & Unconstrained    & P &
					\multicolumn{3}{c|}{0} & 
					\multicolumn{2}{c|}{0} & 
					\multicolumn{1}{c|}{0} &
					\multicolumn{1}{c|}{0} 
					\\ 				 \cline{2-11} 
					& \multirow{3}{*}{\makecell*[c]{Uncon-\\strained}} &
					\(\boldsymbol{\omega}=\mathbf{0}\)
					& Q &
					\multicolumn{3}{c|}{\({\mathbf{b}_{a+}}\)} & 
					\multicolumn{2}{c|}{\({\mathbf{b}_{a+}}\)} & 
					\multicolumn{1}{c|}{\({\mathbf{b}_{a+}}\)} & 
					\multicolumn{1}{c|}{\({\mathbf{b}_{a+}}\)}
					\\ \cline{3-11} 
					& & \(\boldsymbol{\omega}_x=\mathbf{0}\) \(\boldsymbol{\omega}_y=\mathbf{0}\) & R & 					
					\multicolumn{3}{c|}{\(\mathbf{b}_{a+}^{\boldsymbol{\omega}} \)} & 			
					\multicolumn{2}{c|}{\(\mathbf{b}_{a+}^{\boldsymbol{\omega}} \)} & 
					\multicolumn{1}{c|}{\(\mathbf{b}_{a+}^{\boldsymbol{\omega}} \)} &
					\multicolumn{1}{c|}{\(\mathbf{b}_{a+}^{\boldsymbol{\omega}} \)}				
					\\ \cline{3-11} 
					& & Unconstrained    & S &
					\multicolumn{3}{c|}{0} & 
					\multicolumn{2}{c|}{0} & 
					\multicolumn{1}{c|}{0} &
					\multicolumn{1}{c|}{0} 
					\\ \hline
				\end{tabular}
				\small
	For the motions we consider here, they are classified according to motion constraints of reference IMU (Cols~I-VII) and motion constraints of target IMU with respect to reference IMU (Rows~A-S). Here, Cell VII-S represents general motion, while all other cells correspond to various special motions.  \(\mathbf{b}_{a+}\), \(\mathbf{b}_{g+}\), \(\mathbf{b}_{a+}^{\boldsymbol{\omega}}\), \(\mathbf{b}_{g+}^{\boldsymbol{\omega}}\), and \(\mathbf{b}_{g+}^{\boldsymbol{\alpha}}\) denote the corresponding  unobservable directions. In this table, for case Col-I, \(\boldsymbol{\alpha}\) is the constant local acceleration of reference IMU, while for the rest cases Cols~II-VII, \(\boldsymbol{\alpha}\) equals the constant negative local gravity in the reference IMU at initial moment (\(-\prescript{I_1}{}{\mathbf{g}}(t_0)\)).
			\end{threeparttable}
		}
	}\label{table-dpq}
	\normalsize
\end{table*}
\begin{table*}[htp]
	\scriptsize
	\caption{\textbf{Additional} Unobservable Directions with Only \(\mathbf{dp}\) Measurement}
	\vspace{-0.6cm}
	\center{
		\renewcommand{\arraystretch}{1.3}
		\resizebox{1.0\textwidth}{!}{
			\begin{threeparttable}
				\begin{tabular}{|p{1.0cm}<{\centering}|p{1.0cm}<{\centering}|p{2.0cm}<{\centering}|p{0.25cm}<{\centering}|p{1.0cm}<{\centering}|p{1.1cm}<{\centering}|p{1.6cm}<{\centering}|p{1.1cm}<{\centering}|p{1.3cm}<{\centering}|c|c|}
					\hline
					\multicolumn{3}{|c|}{\multirow{3}{*}{\makecell*[c]{Motion Constraints \\ of Target IMU w.r.t. \\ Reference IMU\(^{\ddagger}\)}}} &
					&
					\multicolumn{7}{c|}{Motion Constraints of Reference IMU\(^{\dag}\)} \\ \cline{5-11} 
					\multicolumn{3}{|l|}{} &  & \multicolumn{3}{c|}{\makecell*[c]{\(\prescript{I_1}{}{\boldsymbol{\omega}} = \mathbf{0}\)}} & \multicolumn{2}{c|}{\(\prescript{I_1}{}{\boldsymbol{\omega}} \parallel \boldsymbol{\alpha} \)} & \multicolumn{1}{c|}{\(\prescript{I_1}{}{\boldsymbol{\omega}} \) dir. const. \(\prescript{I_1}{}{\boldsymbol{\omega}} \nparallel \boldsymbol{\alpha} \)} & \multicolumn{1}{c|}{\(\prescript{I_1}{}{\boldsymbol{\omega}} \) unconstrained } 					 \\ \cline{5-11} 
					\multicolumn{3}{|l|}{} & &  \(\prescript{I_1}{}{\mathbf{a}}\) = \( \boldsymbol{\alpha} \) 
					& \(\prescript{I_1}{}{\mathbf{v}} \parallel \boldsymbol{\alpha} \)  
					& \makecell*[c]{ \(\prescript{I_1}{}{\mathbf{v}}\)  dir. const. \\  \(\prescript{I_1}{}{\mathbf{v}} \nparallel \boldsymbol{\alpha} \)  } 
					&   \(\prescript{I_1}{}{\mathbf{a}}\) = \( \boldsymbol{\alpha} \) 
					& \(\prescript{I_1}{}{\mathbf{v}} \perp  \prescript{I_1}{}{\boldsymbol{\omega}}\) 
					& \(\prescript{I_1}{}{\mathbf{v}} \perp  \prescript{I_1}{}{\boldsymbol{\omega}}\)   
					& \makecell[c]{ \(\prescript{I_1}{}{\mathbf{v}}\) unconstrained }
					\\ \hline
					
					\(\prescript{I_1}{}{\mathbf{p}}_{I_2}\) & \(\prescript{I_1}{}{\mathbf{v}}_{I_2}\) & \(\prescript{I_1}{}{\boldsymbol{\omega}}_{I_2}\) &
					&  I & II & III & IV & V & VI &  VII
					\\ 	\hline 	
					\multirow{3}{*}{\(\mathbf{p}=\mathbf{0}\)} &
					\multirow{3}{*}{\(\mathbf{v}=\mathbf{0}\)} &
					\(\boldsymbol{\omega}=\mathbf{0}\)&
					A &
					\(\boldsymbol{\theta}\), \(\mathbf{b}_{g_1}^{\boldsymbol{\alpha}}\) &
					\(\boldsymbol{\theta}^{\boldsymbol{\alpha}}\),\(\mathbf{b}_{g_1}^{\boldsymbol{\alpha}}\) &
					\(\boldsymbol{\theta}^{\boldsymbol{\beta}_1}\)&
					\(\boldsymbol{\theta}^{\boldsymbol{\alpha}}\),\(\mathbf{b}_{g_1}^{\boldsymbol{\alpha}}\) &
					0 &
					0 &
					0  \\ \cline{3-11} 
					& & \multicolumn{1}{c|}{\(\boldsymbol{\omega}_x=\mathbf{0}\) \(\boldsymbol{\omega}_y=\mathbf{0}\)} &B
					&\(\boldsymbol{\theta}\), \(\mathbf{b}_{g_1}^{\boldsymbol{\alpha}}\) &
					\(\boldsymbol{\theta}^{\boldsymbol{\alpha}}\),\(\mathbf{b}_{g_1}^{\boldsymbol{\alpha}}\) &
					\(\boldsymbol{\theta}^{\boldsymbol{\beta}_1}\)&
					\(\boldsymbol{\theta}^{\boldsymbol{\alpha}}\),\(\mathbf{b}_{g_1}^{\boldsymbol{\alpha}}\) &
					0 &
					0 &
					0   \\ \cline{3-11} 
					& & Unconstrained    & C
					&\(\boldsymbol{\theta}\), \(\mathbf{b}_{g_1}^{\boldsymbol{\alpha}}\) &
					\(\boldsymbol{\theta}^{\boldsymbol{\alpha}}\),\(\mathbf{b}_{g_1}^{\boldsymbol{\alpha}}\) &
					\(\boldsymbol{\theta}^{\boldsymbol{\beta}_1}\)&
					\(\boldsymbol{\theta}^{\boldsymbol{\alpha}}\),\(\mathbf{b}_{g_1}^{\boldsymbol{\alpha}}\) &
					0 &
					0 &
					0  
					\\ \hline
					{\multirow{6}{*}{\makecell*[c]{\(\mathbf{p}_x = \mathbf{0}\) \\ \(\mathbf{p}_y = \mathbf{0}\) \\ \(\mathbf{p}_z \neq \mathbf{0}\) }}} &
					{\multirow{3}{*}{\(\mathbf{v}=\mathbf{0}\)}} & \(\boldsymbol{\omega}=\mathbf{0}\)&
					D &\(\boldsymbol{\theta}\), \(\mathbf{b}_{g_1}^{\boldsymbol{\alpha}}\) &
					\(\boldsymbol{\theta}^{\boldsymbol{\alpha}}\),\(\mathbf{b}_{g_1}^{\boldsymbol{\alpha}}\) &
					\(\boldsymbol{\theta}^{\boldsymbol{\beta}_1}\)&
					\(\boldsymbol{\theta}^{\boldsymbol{\alpha}}\),\(\mathbf{b}_{g_1}^{\boldsymbol{\alpha}}\) &
					0 &
					0 &
					0   \\ \cline{3-11} 
					& & \(\boldsymbol{\omega}_x=\mathbf{0}\) \(\boldsymbol{\omega}_y=\mathbf{0}\) & E
					&\(\boldsymbol{\theta}\), \(\mathbf{b}_{g_1}^{\boldsymbol{\alpha}}\) &
					\(\boldsymbol{\theta}^{\boldsymbol{\alpha}}\),\(\mathbf{b}_{g_1}^{\boldsymbol{\alpha}}\) &
					\(\boldsymbol{\theta}^{\boldsymbol{\beta}_1}\)&
					\(\boldsymbol{\theta}^{\boldsymbol{\alpha}}\),\(\mathbf{b}_{g_1}^{\boldsymbol{\alpha}}\) &
					0 &
					0 &
					0  \\ \cline{3-11} 
					& & Unconstrained    & F &
					\(\boldsymbol{\theta}\), \(\mathbf{b}_{g_1}^{\boldsymbol{\alpha}}\) &
					\(\boldsymbol{\theta}^{\boldsymbol{\alpha}}\),\(\mathbf{b}_{g_1}^{\boldsymbol{\alpha}}\) &
					\(\boldsymbol{\theta}^{\boldsymbol{\beta}_1}\)&
					\(\boldsymbol{\theta}^{\boldsymbol{\alpha}}\),\(\mathbf{b}_{g_1}^{\boldsymbol{\alpha}}\) &
					0 &
					0 &
					0   \\ \cline{2-11}  			 &
					\multirow{3}{*}{\makecell*[c]{\(\mathbf{v}_x=\mathbf{0}\) \\ \(\mathbf{v}_y=\mathbf{0}\)}} & \(\boldsymbol{\omega}=\mathbf{0}\)&
					G &\( {\boldsymbol{\theta}}^{\boldsymbol{\alpha}} \), \(\mathbf{b}_{g_1}^{\boldsymbol{\alpha}}\) &
					\(\boldsymbol{\theta}^{\boldsymbol{\alpha}}\),\(\mathbf{b}_{g_1}^{\boldsymbol{\alpha}}\) &
					0&
					\(\boldsymbol{\theta}^{\boldsymbol{\alpha}}\),\(\mathbf{b}_{g_1}^{\boldsymbol{\alpha}}\) &
					0&
					0 &
					0  \\  \cline{3-11} 
					& & \(\boldsymbol{\omega}_x=\mathbf{0}\) \(\boldsymbol{\omega}_y=\mathbf{0}\) & H
					&\( {\boldsymbol{\theta}}^{\boldsymbol{\alpha}} \), \(\mathbf{b}_{g_1}^{\boldsymbol{\alpha}}\) &
					\(\boldsymbol{\theta}^{\boldsymbol{\alpha}}\),\(\mathbf{b}_{g_1}^{\boldsymbol{\alpha}}\) &
					0&
					\(\boldsymbol{\theta}^{\boldsymbol{\alpha}}\),\(\mathbf{b}_{g_1}^{\boldsymbol{\alpha}}\) &
					0 &
					0 &
					0  
					\\ \cline{3-11} 
					& & Unconstrained    & J 
					&\( {\boldsymbol{\theta}}^{\boldsymbol{\alpha}} \), \(\mathbf{b}_{g_1}^{\boldsymbol{\alpha}}\) &
					\(\boldsymbol{\theta}^{\boldsymbol{\alpha}}\),\(\mathbf{b}_{g_1}^{\boldsymbol{\alpha}}\) &
					0&
					\(\boldsymbol{\theta}^{\boldsymbol{\alpha}}\),\(\mathbf{b}_{g_1}^{\boldsymbol{\alpha}}\) &
					0 &
					0 &
					0
					\\ \hline
					\multirow{9}{*}{\makecell*[c]{\(\mathbf{p}_x \neq \mathbf{0}\) \\ \(\mathbf{p}_y \neq \mathbf{0}\) \\ \(\mathbf{p}_z \neq \mathbf{0}\) } } &
					\multirow{3}{*}{\(\mathbf{v}=\mathbf{0}\)} &
					\(\boldsymbol{\omega}=\mathbf{0}\)
					& K &\(\boldsymbol{\theta}\), \(\mathbf{b}_{g_1}^{\boldsymbol{\alpha}}\) &
					\(\boldsymbol{\theta}^{\boldsymbol{\alpha}}\),\(\mathbf{b}_{g_1}^{\boldsymbol{\alpha}}\) &
					\(\boldsymbol{\theta}^{\boldsymbol{\beta}_1}\)&
					0 &
					0 &0 &
					0  
					\\ \cline{3-11} 
					& & \(\boldsymbol{\omega}_x=\mathbf{0}\) \(\boldsymbol{\omega}_y=\mathbf{0}\) & L &\(\boldsymbol{\theta}\), \(\mathbf{b}_{g_1}^{\boldsymbol{\alpha}}\) &
					\(\boldsymbol{\theta}^{\boldsymbol{\alpha}}\),\(\mathbf{b}_{g_1}^{\boldsymbol{\alpha}}\) &
					\(\boldsymbol{\theta}^{\boldsymbol{\beta}_1}\) &
					0 &
					0 &
					0 &
					0  \\ \cline{3-11} 
					& & Unconstrained    & M &\(\boldsymbol{\theta}\), \(\mathbf{b}_{g_1}^{\boldsymbol{\alpha}}\) &
					\(\boldsymbol{\theta}^{\boldsymbol{\alpha}}\),\(\mathbf{b}_{g_1}^{\boldsymbol{\alpha}}\) &
					\(\boldsymbol{\theta}^{\boldsymbol{\beta}_1}\)&
					0 &
					0 &
					0 &
					0   \\ \cline{2-11} 
					& \multirow{3}{*}{\makecell*[c]{\(\mathbf{v}_x=\mathbf{0}\) \\ \(\mathbf{v}_y=\mathbf{0}\)}} & \(\boldsymbol{\omega}=\mathbf{0}\)&
					N &\( {\boldsymbol{\theta}}^{\boldsymbol{\alpha}} \), \(\mathbf{b}_{g_1}^{\boldsymbol{\alpha}}\) &
					\(\boldsymbol{\theta}^{\boldsymbol{\alpha}}\),\(\mathbf{b}_{g_1}^{\boldsymbol{\alpha}}\) &
					0&
					0 &
					0 &
					0 &
					0 
					\\ \cline{3-11} 
					& & \(\boldsymbol{\omega}_x=\mathbf{0}\) \(\boldsymbol{\omega}_y=\mathbf{0}\) & O
					&\( {\boldsymbol{\theta}}^{\boldsymbol{\alpha}} \), \(\mathbf{b}_{g_1}^{\boldsymbol{\alpha}}\) &
					\(\boldsymbol{\theta}^{\boldsymbol{\alpha}}\),\(\mathbf{b}_{g_1}^{\boldsymbol{\alpha}}\) &
					0&
					0 &
					0 &
					0 &
					0 
					\\ 
					\cline{3-11} 
					& & Unconstrained    & P &\( {\boldsymbol{\theta}}^{\boldsymbol{\alpha}} \), \(\mathbf{b}_{g_1}^{\boldsymbol{\alpha}}\) &
					\(\boldsymbol{\theta}^{\boldsymbol{\alpha}}\),\(\mathbf{b}_{g_1}^{\boldsymbol{\alpha}}\) &
					0&
					0 &
					0 &
					0 &
					0 
					\\ 
					\cline{2-11} 
					\specialrule{0em}{-1pt}{0pt}
					& \multirow{3}{*}{\makecell*[c]{Uncon-\\strained}} &
					\(\boldsymbol{\omega}=\mathbf{0}\)
					& Q &
					0 &0 &0 &0 &0 &0 &0 
					\\ 	\specialrule{0em}{0pt}{-1pt}  \cline{3-11} 
					\specialrule{0em}{-1pt}{0pt}
					& & \(\boldsymbol{\omega}_x=\mathbf{0}\) \(\boldsymbol{\omega}_y=\mathbf{0}\) & R &
					0 &0 &0 &0 &0 &0 &0
					\\ 					\specialrule{0em}{0pt}{-1pt} \cline{3-11} 
					\specialrule{0em}{-1pt}{0pt}
					& & Unconstrained    & S &
					0 &0 &0 &0 &0 &0 &0
					\\ 					 \hline 
				\end{tabular}
\small
The special motions we consider here, as well as their classifications, are the same with Table~(\ref{table-dpq}).  \( \boldsymbol{\theta}\) (composed of  \(\boldsymbol{\theta}^{\boldsymbol{\beta}_1}\), \(\boldsymbol{\theta}^{\boldsymbol{\beta}_2}\), and  \(\boldsymbol{\theta}^{\boldsymbol{\alpha}} \)), \( \boldsymbol{\theta}^{\boldsymbol{\alpha}}\), \( \boldsymbol{\theta}^{\boldsymbol{\beta}_1}\), and \( \mathbf{b}_{g_1}^{\boldsymbol{\alpha}}\) denote the additional unobservable directions. In this table, \(\boldsymbol{\alpha}\) follows the same definition with Table~(\ref{table-dpq}), and for case Col-I,  \(\boldsymbol{\beta}_1\) and  \(\boldsymbol{\beta}_2\) constitute an orthogonal basis perpendicular to \(\boldsymbol{\alpha}\), while for case Col-III, \(\boldsymbol{\beta}_1\) is the constant direction of \(\prescript{I_1}{}{\mathbf{v}}\). 
Note that, the total unobservable directions with only \(\mathbf{dp}\) measurement are obtained by adding together Table~(\ref{table-dpq}) and Table~(\ref{table-dp}).	 
\\
\\
\small
\(^{\dagger}\)Cols~I-VII represent motion constraints of the reference IMU, i.e., the local \(\prescript{I_1}{}{\boldsymbol{\omega}}\), \(\prescript{I_1}{}{\mathbf{a}}\), \(\prescript{I_1}{}{\mathbf{v}}\) of \(\{I_1\}\). Based on the definition of \(\boldsymbol{\alpha}\), examples of Cols~I-VII are: stationary, uniform linear or uniform acceleration linear motion with no rotation (Col-I); linear motion along gravity direction with no rotation (Col-II);  linear motion in a non-gravity direction with no rotation (Col-III); rotating around gravity in place or with uniform linear translation (Col-IV); planar motion on a horizontal surface (Col-V); planar motion on an inclined surface (Col-VI); free 3D motion (Col-VII).	
\vspace{0.3cm}

\small
\(^{\ddagger}\)Rows~A-S represent motion constraints of the target IMU wrt the reference IMU, i.e., the relative \(\prescript{I_1}{}{\mathbf{p}}_{I_2}\), \(\prescript{I_1}{}{\mathbf{v}}_{I_2}\), \(\prescript{I_1}{}{\boldsymbol{\omega}}_{I_2}\) between \(\{I_1\}\) and \(\{I_2\}\). Without loss of generality, we align z-axis of the reference IMU at initial moment with \(\boldsymbol{\alpha}\), and x, y-axis are the perpendicular directions. We use \(\mathbf{p}_{j} = 0\), \(\mathbf{v}_{j} = 0\), and \(\boldsymbol{\omega}_{j} = 0\), \(j=x,y,z\), to represent that the relative position (\(\prescript{I_1}{}{\mathbf{p}}_{I_2}\)), relative linear velocity (\(\prescript{I_1}{}{\mathbf{v}}_{I_2}\)), and relative angular velocity (\(\prescript{I_1}{}{\boldsymbol{\omega}}_{I_2} = \mathbf{C}(\prescript{I_1}{}{\mathbf{q}}_{I_2}) \prescript{I_2}{}{\boldsymbol{\omega}} - \prescript{I_1}{}{\boldsymbol{\omega}} \)) has constant zero for its \(j\)-th element, respectively, while \(\neq 0\) to denote this element is non-zero for certain time steps. 
		\end{threeparttable}}\label{table-dp}
	}
\end{table*}

\FloatBarrier
		
	\subsection{Unobservable Directions with Only \(\mathbf{dp}\) Measurement \label{unobs-dp}}
	Under the same types of special motions, the unobservable directions in the case with both \(\mathbf{dp}\) and \(\mathbf{dq}\) measurements, as shown in Table~(\ref{table-dpq}), will also be present in the case with only \(\mathbf{dp}\) measurement. Furthermore, the absence of \(\mathbf{dq}\) measurement will lead to some additional unobservable direction corresponding to relative orientation and some absolute bias:
	\begin{align}
 	&\left[ \underbrace{\overbrace{\boldsymbol{\theta}^{\boldsymbol{\beta}_1} \ | \    \boldsymbol{\theta}^{\boldsymbol{\beta}_2}}^{{\boldsymbol{\theta}^{\boldsymbol{\beta}}}_{21\times2}} \ | \  \boldsymbol{\theta}^{\boldsymbol{\alpha}} }_{{\boldsymbol{\theta}}_{21 \times 3}} \ | \  {\mathbf{b}_{g_1}^{\boldsymbol{\alpha}}} \right]_{\left[21 \times 1, \ 21 \times 1,\ 21 \times 1,\ 21 \times 1 \right]}
 \nonumber \\
= & \begin{bmatrix}
	\mathbf{0}_{3\times2}   &\mathbf{0}_{3\times1}  & \mathbf{0}_{3\times1}  \\
	\mathbf{0}_{3\times2} &\mathbf{0}_{3\times1}  & \mathbf{0}_{3\times1}  \\
	\mathbf{C}^T(\prescript{I_1}{}{\mathbf{q}}_{I_2}(t_0))\begin{bmatrix}
		\boldsymbol{\beta}_1 & \boldsymbol{\beta}_2
	\end{bmatrix}  & \mathbf{C}^T(\prescript{I_1}{}{\mathbf{q}}_{I_2}(t_0)) \boldsymbol{\alpha} &  \mathbf{0}_{3\times1} \\
	\mathbf{0}_{3\times2}  &\mathbf{0}_{3\times1} &\boldsymbol{\alpha}    \\
	\mathbf{0}_{3\times2}  &\mathbf{0}_{3\times1}  & \mathbf{0}_{3\times1}\\
	\begin{bmatrix}
		\boldsymbol{\alpha}	\times \boldsymbol{\beta}_1 &  \boldsymbol{\alpha} \times \boldsymbol{\beta}_2  
	\end{bmatrix} &\mathbf{0}_{3\times1}  &  \mathbf{0}_{3\times1} \\
	\mathbf{0}_{3\times2} &\mathbf{0}_{3\times1}  &\mathbf{0}_{3\times1} 
\end{bmatrix} \label{Np}  
	\end{align}
	 where \(\boldsymbol{\alpha}\), \(\boldsymbol{\beta}_1\), \(\boldsymbol{\beta}_2\) are defined in Table~(\ref{table-dp}).

	The \emph{additional} unobservable directions with only \(\mathbf{dp}\) measurement are summarized in Table~(\ref{table-dp}), the proof is shown in Appendix E-2. Therefore, we can obtain the \emph{total} unobservable state directions for the \(\mathbf{dp}\) only case, by querying Table~(\ref{table-dpq}) and Table~(\ref{table-dp}) and adding them together. Note, the composite gyroscope biases \(\mathbf{b}_{g+}\), composite accelerometer biases \(\mathbf{b}_{a+}\), relative oritentation \(\boldsymbol{\theta}\) and absolute gyroscope bias \( \mathbf{b}_{g_1}^{\boldsymbol{\alpha}}\), from \eqref{Npq} and \eqref{Np}, form the union of all unobservable directions in these cases. In what follows, we will explain the physical meaning of each direction under its corresponding special motion types.

\begin{itemize}
	\item 	
	The unobservable directions \(\boldsymbol{\theta}^{\boldsymbol{\alpha}}\) and \({\mathbf{b}_{g_1}^{\boldsymbol{\alpha}}}\)
	\begin{itemize}
		\item[-] \textit{Physical Quantity:} \(\boldsymbol{\theta}^{\boldsymbol{\alpha}}\) is 1 dof of the relative orientation about the \(\boldsymbol{\alpha}\) direction, i.e., the yaw angle with respect to \(\boldsymbol{\alpha}\). And, \({\mathbf{b}_{g_1}^{\boldsymbol{\alpha}}}\) is 1 dof of the absolute gyroscope bias of reference IMU along \(\boldsymbol{\alpha}\) direction.

		\item[-]\textit{Sufficient Conditions:}
		\begin{enumerate}[itemindent=26pt]
		\item [c4.1]
			\( \prescript{I_1}{}{}{\mathbf{v}}_{I_2} \parallel \boldsymbol{\alpha} \) or  \( ^{I_1}\mathbf{v}_{I_2}  =  \mathbf{0}_{3 \times 1} \); 
		\item [and c4.2]
			\(\prescript{I_1}{}{}{\mathbf{a}} \parallel \boldsymbol{\alpha} \);
		\item [and c4.3]
			\(\prescript{I_1}{}{}{\boldsymbol{\omega}} \parallel \boldsymbol{\alpha} \) or \(\prescript{I_1}{}{}{\boldsymbol{\omega}} = \mathbf{0}_{3 \times 1} \);
		\item [and c4.4]
			\(\prescript{I_1}{}{}{\mathbf{p}}_{I_2} \parallel \prescript{I_1}{}{}{\boldsymbol{\omega}} \) or \(\prescript{I_1}{}{}{\mathbf{p}}_{I_2} = \mathbf{0}_{3 \times 1}\) or \(\prescript{I_1}{}{}{\boldsymbol{\omega}}  = \mathbf{0}_{3 \times 1} \);
\end{enumerate}
	e.g., two stationary robots.
	
	\item[-]\textit{Interpretation:} Similar to the system of a single agent with only GPS measurement, in the absence of motion perpendicular to the direction of gravity, heading angle error cannot induce changes in translational measurement \cite{gpssingleagent}. The condition c4.1 corresponds to that no motion perpendicular to gravity. However, the ``gravity'' in a Dual-IMU system, as a non-inertial frame, may vary. Here, we introduce ``non-inertial-gravity'', which is composed of the gravity and the ``inertial force'' caused by reference frame motion, to represent the time-varying gravity. Analogous to gravity on Earth, \(^G\mathbf{g}  = \prescript{G}{}{\mathbf{C}}_{I}\prescript{I}{}{}\mathbf{a} - \prescript{G}{}{\mathbf{a}}_{I}  \),  the non-inertial-gravity \(\prescript{I_1}{}{\mathbf{g}}' = \mathbf{C}\left(\prescript{{I_1}}{}{\mathbf{q}}_{I_2} \right)\prescript{{I_2}}{}{\mathbf{a}} -  \prescript{I_1}{}{}{\dot{\mathbf{v}}}_{I_2}\). According to \eqref{dotv}, the ``non-inertial-gravity'' depend on relative position, relative velocity and acceleration, angular velocity of reference IMU.
	In this case, conditions c4.2, c4.3 and c4.4 remain non-inertial-gravity in a consistent direction \(\boldsymbol{\alpha}\). Hence, the yaw angler with respect to \(\boldsymbol{\alpha}\), denoted as \(\boldsymbol{\theta}^{\boldsymbol{\alpha}}\), is unobservable. Furthermore, the information of gyroscope bias \({\mathbf{b}_{g_1}^{\boldsymbol{\alpha}}}\) cannot be directly obtained from relative orientation, due to the relative orientation in this direction \(\boldsymbol{\theta}^{\boldsymbol{\alpha}}\) is unobservable. %
	Moreover, it also cannot be obtained absolute gyroscope bias of reference IMU from relative velocity, due to the same reason as \(\mathbf{b}_{g+}^{\boldsymbol{\alpha}}\). Therefore, the direction \(\mathbf{b}_{g_1}^{\boldsymbol{\alpha}}\) is also unobservable.

	\end{itemize}
	\item 
	The unobservable directions \(\boldsymbol{\theta}^{\boldsymbol{\beta}} \)
	\begin{itemize}
		\item [-]\textit{Physical Quantity:} \(\boldsymbol{\theta}^{\boldsymbol{\beta}} \) is 2 dof of the relative orientation about the \(\boldsymbol{\beta}\) directions, i.e., the two tilt angles with respect to \(\boldsymbol{\alpha}\), and coupled with the absolute accelerometer bias of reference IMU along \(\boldsymbol{\beta}\) directions.
	
	\item [-]\textit{Sufficient Conditions:} 
	\begin{enumerate}[itemindent=25pt]
	\item [c5.1]
	\(\prescript{I_1}{}{\mathbf{a}} = \boldsymbol{\alpha}\);
	\item [and c5.2]
	\(\prescript{I_1}{}{\boldsymbol{\omega}} = \mathbf{0}_{3\times1}\);	
	\item [and c5.3]
	\(\prescript{I_1}{}{\mathbf{v}}_{I_2} = \mathbf{0}_{3\times1}\);
	\end{enumerate}
	e.g., an HMD undergoing pure rotation inside a stationary vehicle.
	
	\item [-]\textit{Interpretation:} In our system, unlike real gravity, there may be an error in the measurement of non-inertial-gravity, and this error can come from the accelerometer bias of the reference IMU. The accelerometer bias errors perpendicular to the direction of non-inertial-gravity result in the tilting of the non-inertial-gravity vector, causing tilt angle errors. 
	Furthermore, we know that the relative tilting angle error and coupled bias error may propagate to the velocity error, which is observable, from the \(\mathbf{F}\) matrix in \eqref{FF}.
	But, if the error of relative orientation and accelerometer bias follow condition \(-\hat{\mathbf{C}}[\prescript{I_2}{}{{\hat{\mathbf{a}}}}_\times] \delta \boldsymbol{\theta} + \tilde{\mathbf{b}}_{a_1} = \mathbf{0}_{3\times 1}\), the angle error and bias error will not be measured.
	In this case, \(\hat{\mathbf{C}}[\prescript{I_2}{}{{\hat{\mathbf{a}}}}_\times] = -\left[\prescript{I_1}{}{\mathbf{g}}'_{\times}\right]\hat{\mathbf{C}} = \left[\boldsymbol{\alpha}_{\times}\right]\hat{\mathbf{C}} \).
	Relative orientation error along \(\hat{\mathbf{C}}^T \boldsymbol{\beta} \) direction and the coupled absolute bias error along \(\boldsymbol{\alpha} \times \boldsymbol{\beta}\) direction. \(-\hat{\mathbf{C}}[\prescript{I_2}{}{{\hat{\mathbf{a}}}}_\times] \delta \boldsymbol{\theta} + \tilde{\mathbf{b}}_{a_1}\) constant equal to zero. Hence, the \(\boldsymbol{\theta}^{\boldsymbol{\beta}} \) remains unobservable.
	
\end{itemize}
		
	\item {The unobservable direction \(\boldsymbol{\theta}^{\boldsymbol{\beta}_1} \)}
\begin{itemize}
	\item [-]\textit{Physical Quantity:} \(\boldsymbol{\theta}^{\boldsymbol{\beta}_1} \) is 1 dof of the relative orientation about the \(\boldsymbol{\beta}_1\) direction, and coupled with the absolute accelerometer bias of reference IMU along \(\boldsymbol{\alpha}	\times \boldsymbol{\beta}_1\) direction.
	
	\item [-]\textit{Sufficient Conditions:}
	\begin{enumerate}[itemindent=26pt]
		\item [c6.1]
		\(\prescript{I_1}{}{}{\mathbf{v}} \parallel \boldsymbol{\beta}_1\) and \(\boldsymbol{\beta}_1 \nparallel \boldsymbol{\alpha} \);
		\item [and c6.2]
		\(\prescript{I_1}{}{}{\boldsymbol{\omega}}  = \mathbf{0}_{3\times1}\);
		\item [and c6.3]
		\(\prescript{I_1}{}{}{\mathbf{v}}_{I_2} =  \mathbf{0}_{3\times1} \);
	\end{enumerate}
	e.g., an HMD is stationary in a vehicle with the vehicle moving along a straight line.
	
	\item [-]\textit{Interpretation:} Same as \(\boldsymbol{\theta}^{\boldsymbol{\beta}}\), although, the \(\hat{\mathbf{C}}[\prescript{I_2}{}{{\hat{\mathbf{a}}}}_\times] = -\left[\prescript{I_1}{}{\mathbf{g}}'_{\times}\right]\hat{\mathbf{C}} = [(\boldsymbol{\alpha} + k \boldsymbol{\beta}_1 )_{\times}]\hat{\mathbf{C}}^T\) is time-varying in this case, the \(\prescript{I_1}{}{\mathbf{g}}'\) varying only along one direction. The error along \(\boldsymbol{\theta}^{\boldsymbol{\beta}_1} \) direction still satisfies the condition, which is \(-\hat{\mathbf{C}}[\prescript{I_2}{}{{\hat{\mathbf{a}}}}_\times] \delta \boldsymbol{\theta} + \tilde{\mathbf{b}}_{a_1} = \mathbf{0}_{3\times 1}\). Hence, the direction \(\boldsymbol{\theta}^{\boldsymbol{\beta}_1} \) remains unobservable.
	
	\end{itemize}
	
\end{itemize}

	In summary, we have shown that these special motions can introduce multiple unobservable directions to the system. We can classify all these cases into two categories: The first category is when certain components of the relative orientation are unobservable (all non-zero cells in Table~(\ref{table-dp})), e.g., when a vehicle stays still or in uniform linear motion, while an HMD purely rotates inside the vehicle.
The second category is when only biases may be unobservable (all other cells in Table~(\ref{table-dpq}) ), e.g., when either agent undergoes sufficient motion, while the other agent can be moving or stationary. In practice, a mobile agent's motion is typically the combination of general and special motions, with each mode occupying certain periods of time. As shown in~\cite{wu-vins-on-wheels}, ``approximate'' or intermittent special motion profiles may cause estimation inaccuracy issues as well. Therefore, according to our observability analysis, if the two agents follow special motions in the first category, even if it is only approximate, the system with only \(\mathbf{dp}\) measurement may have a large estimation error along the unobservable directions of the relative orientation (e.g., relative yaw error). Otherwise, if the agents follow special motions in the second category, then individual IMU bias estimates may be unreliable, but the relative pose states, since they are observable, can be accurate.

	\section{Numerical Simulation\label{6-numerical-simulation}}
	\FloatBarrier

	In order to verify the unobservable directions under the special motions in the preceding section and show their impact on the state estimates, we conduct Monte Carlo simulations based on EKF.
    Based on
	the distinct motions of the two IMUs, we provide a comprehensive
	presentation of the results for five typical patterns, corresponding to
	Cell~I-S, V-M, V-K, III-K, and I-K in Table~(\ref{table-dpq}) and (\ref{table-dp}). For each pattern, we
	consider two measurement methods: with both 
	\(\mathbf{dp}\) and \(\mathbf{dq}\)
	measurements, and with only \(\mathbf{dp}\) measurement. For each of the 10 cases, we execute 50 independent runs with random noise inputs, and calculate the root mean square error (RMSE)~\cite{refrmse} and 3\(\sigma\) bound for each state per time step, serving as an assessment of the accuracy and consistency of the estimates.
	
	For all the cases, to understand the differences of the first and second order estimates between the observable and unobservable directions, there are several possible behaviors that may appear. As for observable directions, the results tend to be the same, where the error rapidly decreases to a very small value and remains well within the 3\(\sigma\) bound. Meanwhile, as for unobservable directions, there are 3 types of behaviors: 1) The error rapidly increases and significantly exceeds the 3\(\sigma\) bound. 2) The error does not converge and wanders around or approaches the boundaries of the 3\(\sigma\) bound. 3) Although the error does not exceed the 3\(\sigma\) bound, it does not converge, where the error and the 3\(\sigma\) bound continuously increases.
	
	It's worth noting that in some cases, each component of biases might be unobservable, but the  directions of \(\mathbf{b}_{g-}\) and \(\mathbf{b}_{a-}\) in \eqref{bag-} are observable and the directions of \(\mathbf{b}_{g+}\) and \(\mathbf{b}_{a+}\) in \eqref{Npq} are unobservable.

	\subsection{Both \(\mathbf{dp}\) and \(\mathbf{dq}\) Measurements}

	From Fig.~\ref{dpq_static_pxyz_vxvyvz_wxwywz} (Cell I-S in Table~(\ref{table-dpq})) and  Fig.~\ref{dpq_plane_pxyz_wxyz} (Cell V-M in Table~(\ref{table-dpq})), it's quite evident that all states are observable. From Fig.~\ref{dpq_static_pxyz_vxvyvz_wxwywz}, although the reference IMU is stationary, the significant motion of the target IMU ensures that all states are observable.
	From Fig.~\ref{dpq_plane_pxyz_wxyz}, even though there is only relative rotational motion between
	the two IMUs, the sufficient planar motion of the reference IMU enables
	rapid error convergence. 
	In comparison to Fig.~\ref{dpq_plane_pxyz_wxyz}, where the target IMU has no relative rotation
	with respect to the reference IMU, the 3 directions of \(\mathbf{b}_{a+}\)
	become unobservable in Fig.~\ref{dpq_plane_pxyz} (Cell V-K in Table~(\ref{table-dpq})). 
	From Fig.~\ref{dpq_plane_pxyz}(j), we can notice that
	although the errors do not exceed the 3\(\sigma\) bound, both the errors
	and their 3\(\sigma\) bounds are gradually increasing.
	Comparing Fig.~\ref{dpq_straight_pxyz} (Cell III-K in Table~(\ref{table-dpq})) to Fig.~\ref{dpq_plane_pxyz}, it is evident that besides \(\mathbf{b}_{a+}\), the 3 directions of
	\(\mathbf{b}_{g+}\) for Fig.~\ref{dpq_straight_pxyz}(h) are also unobservable that the errors wander around the boundaries of the 3\(\sigma\) bound. The performance of Fig.~\ref{dpq_static_static} (Cell I-K in Table~(\ref{table-dpq})) is the same as Fig.~\ref{dpq_straight_pxyz}, with a total of 6
	directions being unobservable. By analyzing the error results across the five motion patterns, it can
	be observed that when both \(\mathbf{dp}\) and \(\mathbf{dq}\)
	measurements are available, the relative position and rotation are
	observable. Additionally, \(\mathbf{b}_{g-}\) and \(\mathbf{b}_{a-}\) are also observable.

	\FloatBarrier
	\subsection{Only \(\mathbf{dp}\) Measurement}
	
\begin{figure*}[htb]
		\centering
		\includegraphics[scale=0.335]{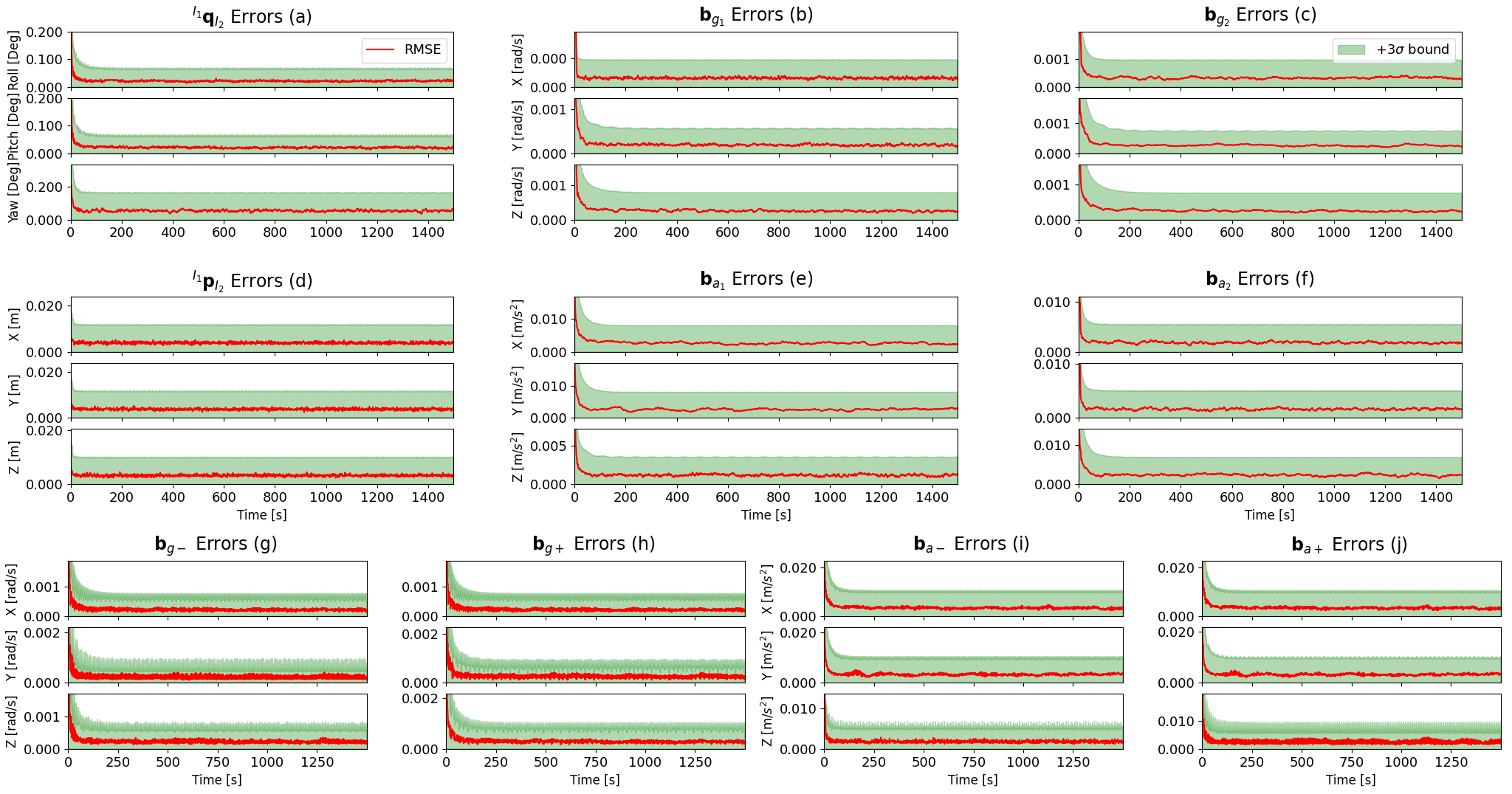}
		\caption{The RMSE and 3\(\sigma\) bound of the motion that the reference IMU remains still and the target IMU moves generally with respect to the reference IMU in~Cell~I-S of Table~(\ref{table-dpq}) with both \(\mathbf{dp}\) and \(\mathbf{dq}\) measurements. The red line represents RMSE, and the green background represents 3\(\sigma\) bound. The definition of each subplot is that: (a) and (d) represent the
			relative orientation and relative position errors, (b) and
			(c) represent the gyroscope bias errors of reference IMU and target IMU, (e) and (f) represent the accelerometer bias errors of
			reference IMU and target IMU, (g) and (h) represent the
			errors of \(\mathbf{b}_{g-}\) and \(\mathbf{b}_{a-}\), (i) and (j) represent
			the errors of \(\mathbf{b}_{g+}\) and \(\mathbf{b}_{a+}\), respectively. All states are
			observable.}
		\label{dpq_static_pxyz_vxvyvz_wxwywz}
	\end{figure*}

\begin{figure*}[htb]
		\centering
		\includegraphics[scale=0.335]{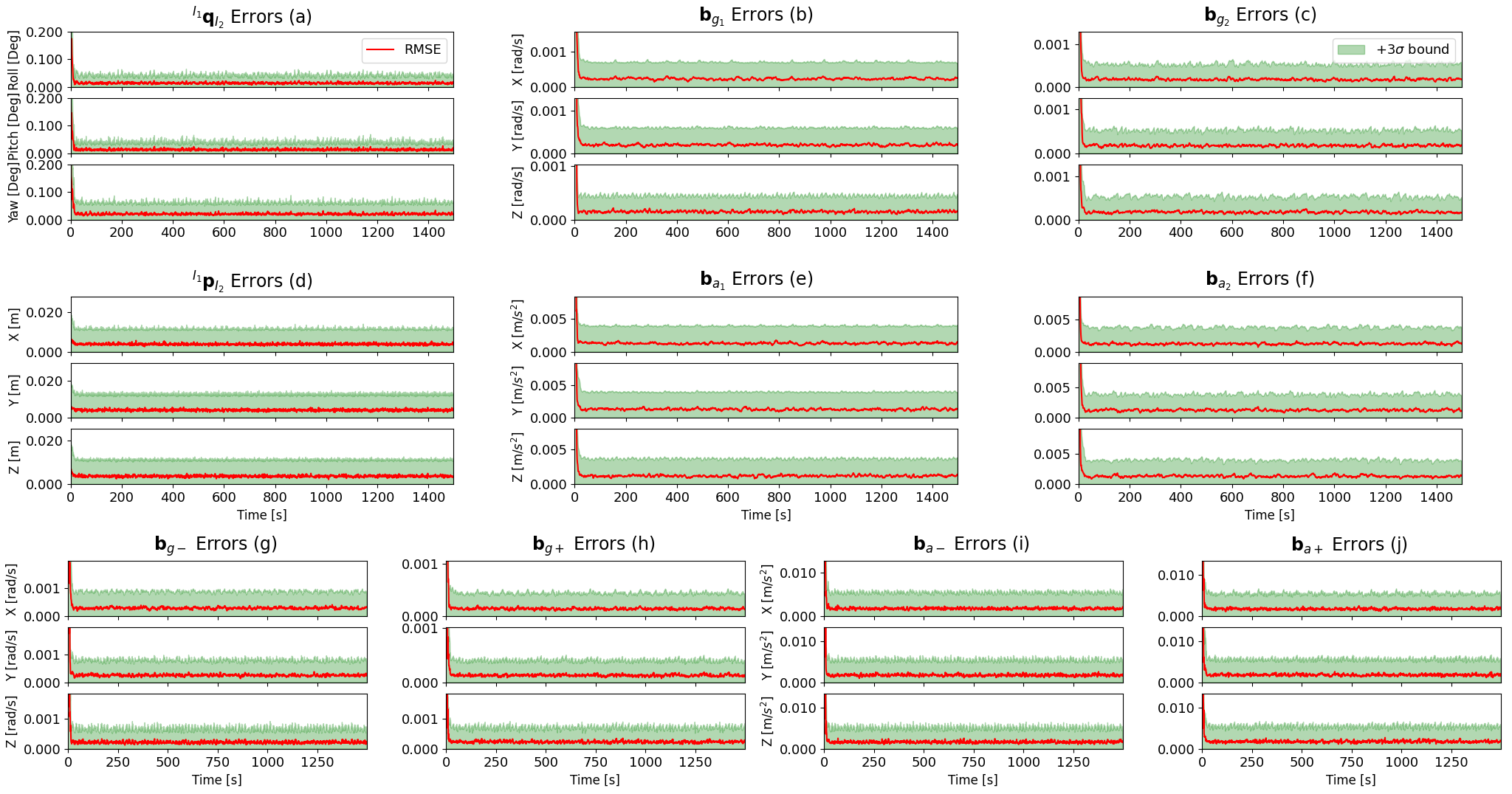}
		\caption{The RMSE and 3\(\sigma\) bound of the motion that the reference IMU moves on a horizontal plane and the target IMU moves only rotates with respect to the reference IMU in~Cell~V-M of Table~(\ref{table-dpq}) with both \(\mathbf{dp}\) and \(\mathbf{dq}\) measurements. The red line represents RMSE, and the green background represents 3\(\sigma\) bound. The definition of each subplot is the same as in Fig.~\ref{dpq_static_pxyz_vxvyvz_wxwywz}. All states are
			observable.}
		\label{dpq_plane_pxyz_wxyz}
	\end{figure*}

\begin{figure*}[htb]
		\centering
		\includegraphics[scale=0.335]{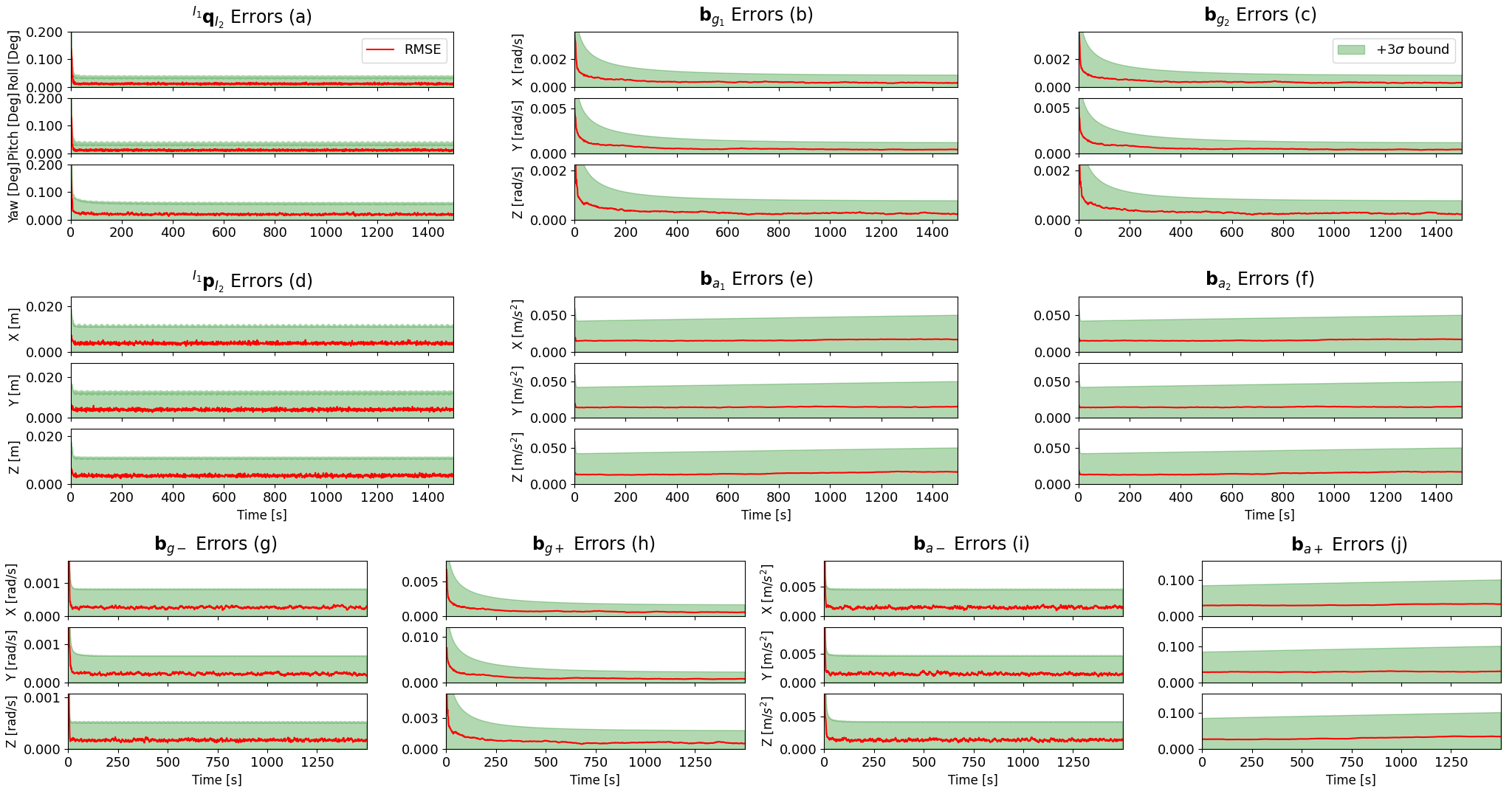}
		\caption{The RMSE and 3\(\sigma\) bound of the motion that the reference IMU moves on a horizontal plane and the target IMU remains still with respect to the reference IMU in~Cell~V-K of Table~(\ref{table-dpq}) with both \(\mathbf{dp}\) and \(\mathbf{dq}\) measurements. The red line represents RMSE, and
			the green background represents 3\(\sigma\) bound. The definition of each subplot is the same as in Fig.~\ref{dpq_static_pxyz_vxvyvz_wxwywz}. There are 3 unobservable directions: 3 directions of \(\mathbf{b}_{a+}\).}
		\label{dpq_plane_pxyz}
	\end{figure*}

\begin{figure*}[htb]
		\centering
		\includegraphics[scale=0.335]{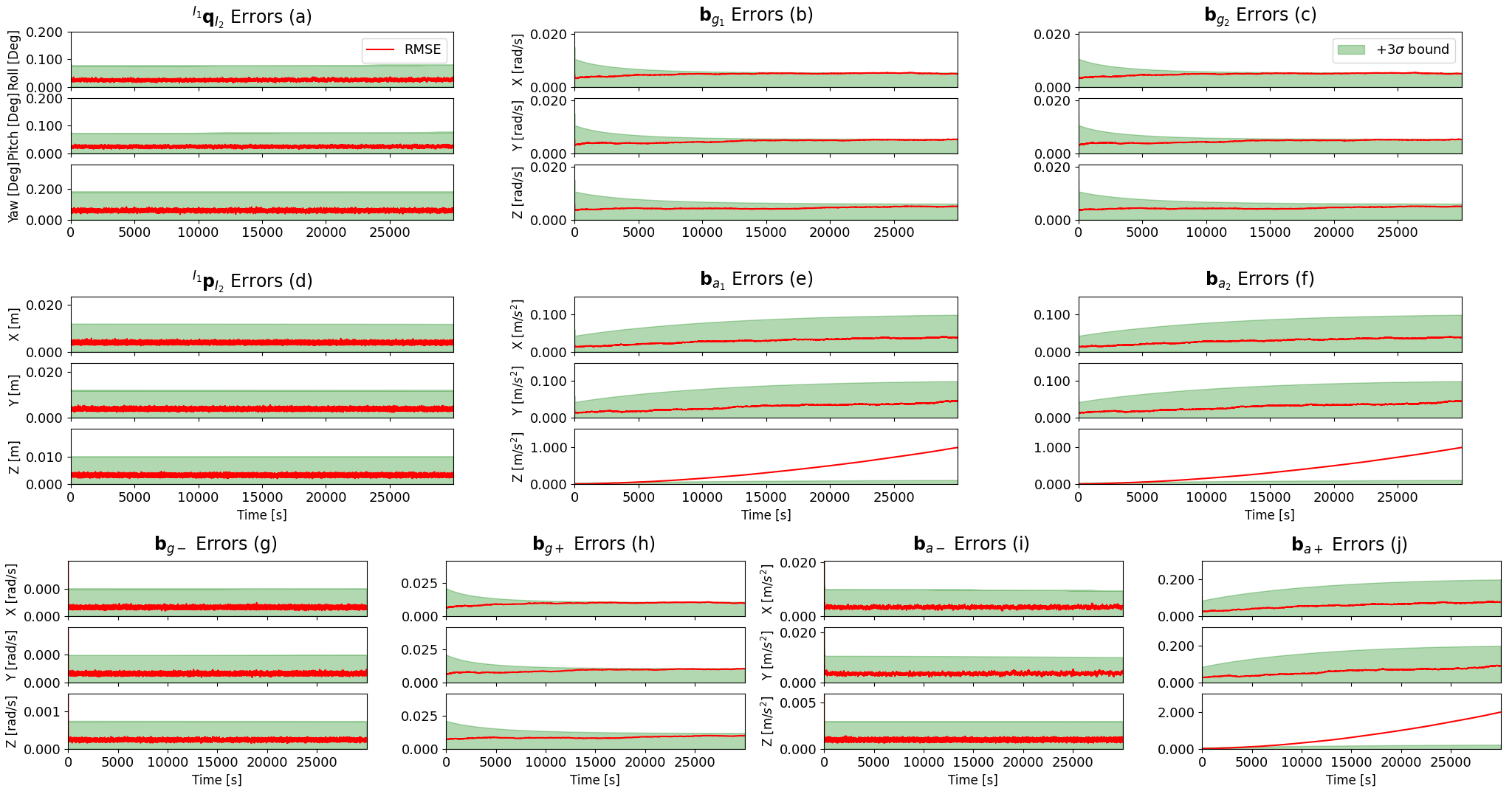}
		\caption{The RMSE and 3\(\sigma\) bound of the motion that the reference IMU moves along a straight line (the X direction) and the target IMU remains still with respect to the reference IMU in~Cell~III-K of Table~(\ref{table-dpq}) with both \(\mathbf{dp}\) and \(\mathbf{dq}\) measurements. The red line represents RMSE, and the
			green background represents 3\(\sigma\) bound. The definition of each subplot is the same as in Fig.~\ref{dpq_static_pxyz_vxvyvz_wxwywz}. There are 6 unobservable directions: 3 directions of \(\mathbf{b}_{g+}\) and 3
			directions of \(\mathbf{b}_{a+}\).}
		\label{dpq_straight_pxyz}
	\end{figure*}

\begin{figure*}[htb]
		\centering
		\includegraphics[scale=0.335]{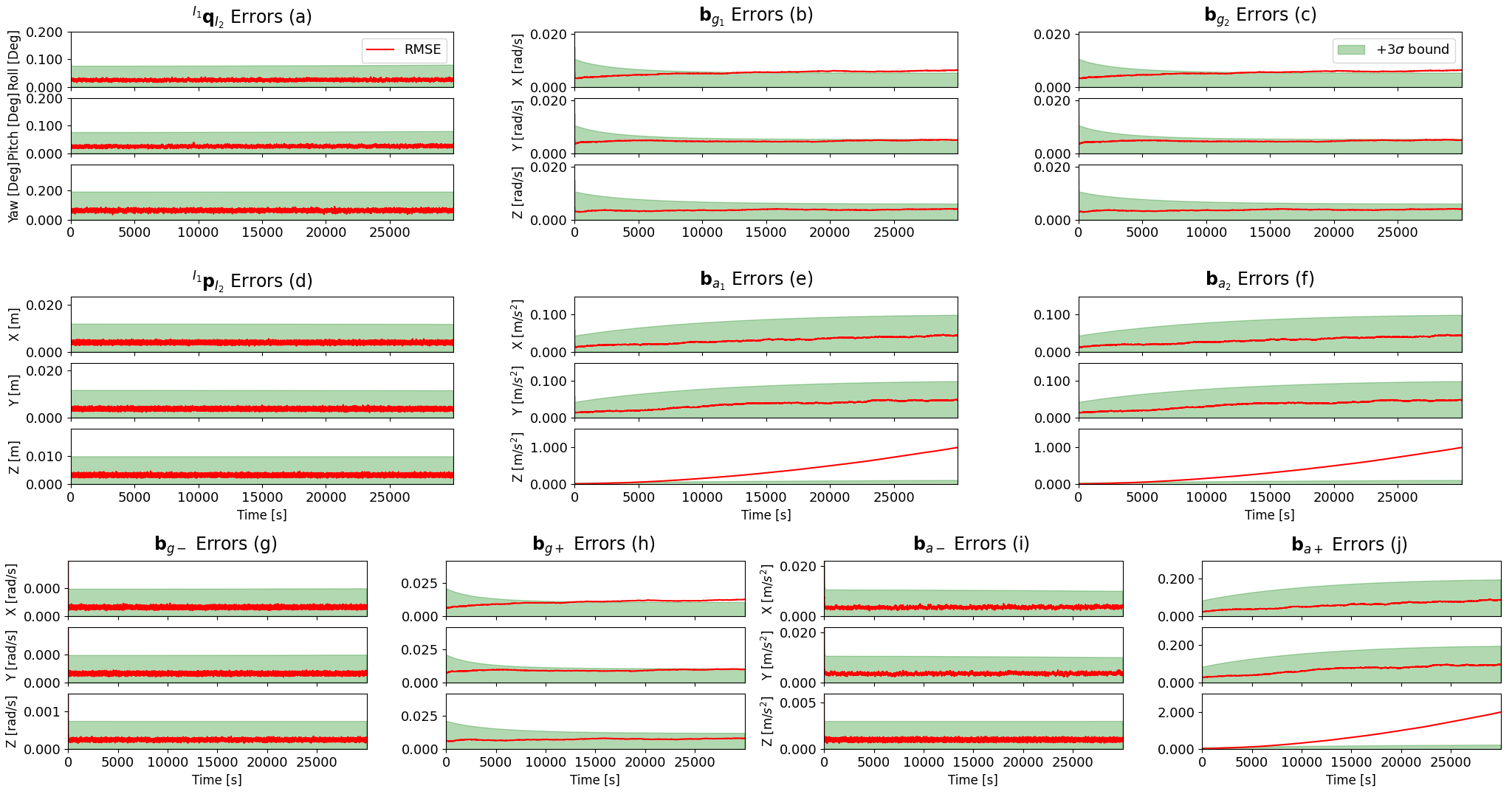}
		\caption{The RMSE and 3\(\sigma\) bound of the motion that the reference IMU remains still and the target IMU remains still with respect to the reference IMU in~Cell~I-K of Table~(\ref{table-dpq}) with both \(\mathbf{dp}\) and \(\mathbf{dq}\) measurements. The red line represents RMSE, and the
			green background represents 3\(\sigma\) bound. The definition of each subplot is the same as in Fig.~\ref{dpq_static_pxyz_vxvyvz_wxwywz}. There are 6 unobservable directions: 3 directions of \(\mathbf{b}_{g+}\) and 3
			directions of \(\mathbf{b}_{a+}\).}
		\label{dpq_static_static}
	\end{figure*}

\begin{figure*}[!htb]
		\centering
		\includegraphics[scale=0.335]{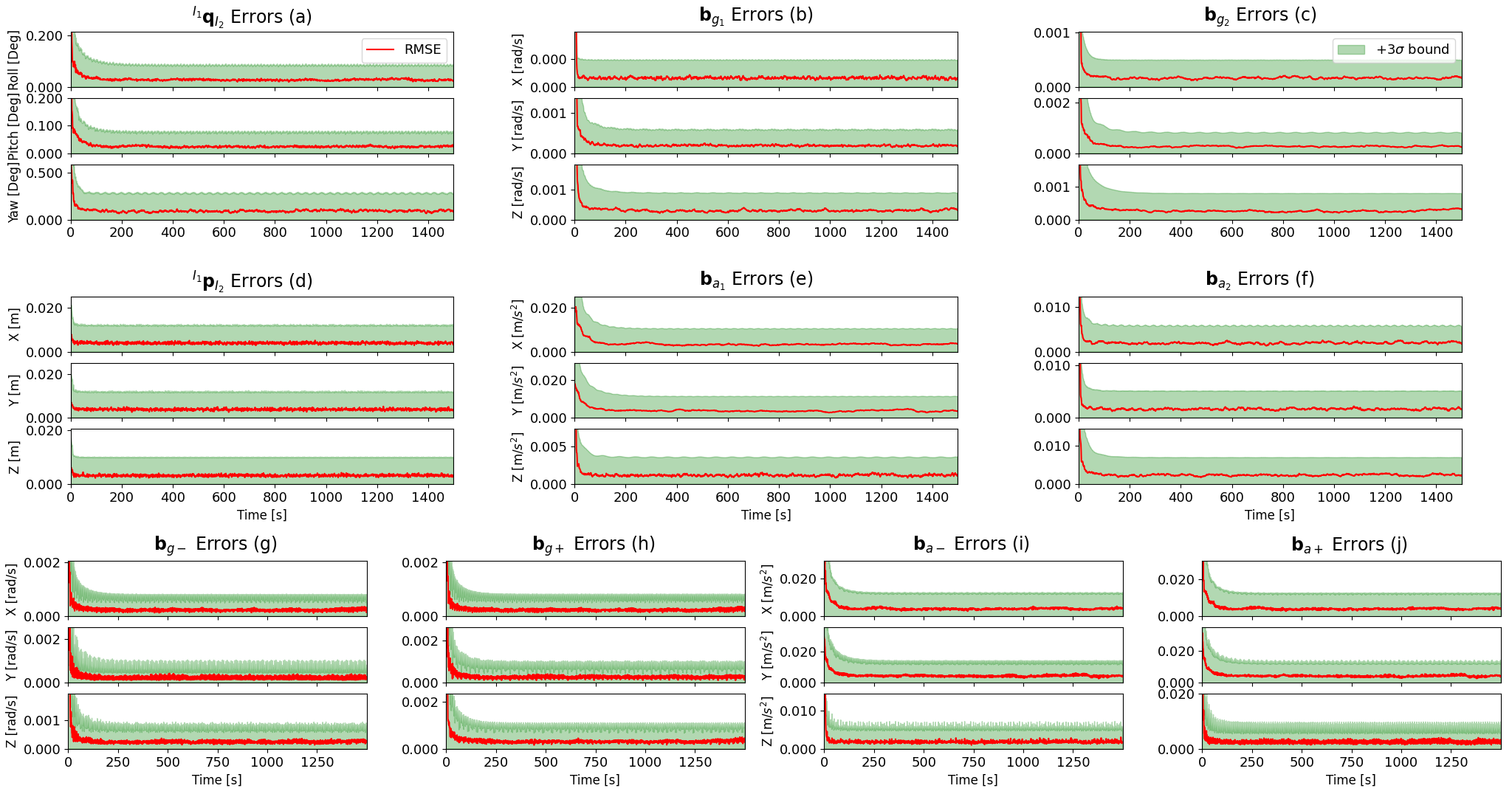}
		\caption{The RMSE and 3\(\sigma\) bound of the motion that the reference IMU remains still and the target IMU moves generally with respect to the reference IMU in~Cell~I-S of both Table~(\ref{table-dpq}) and Table~(\ref{table-dp}) with only \(\mathbf{dp}\) measurement. The red line represents RMSE, and the
			green background represents 3\(\sigma\) bound. The definition of each subplot is the same as in Fig.~\ref{dpq_static_pxyz_vxvyvz_wxwywz}. All states are
			observable.}
		\label{dp_static_pxyz_vxvyvz_wxwywz}
	\end{figure*}

\begin{figure*}
		\centering
		\includegraphics[scale=0.335]{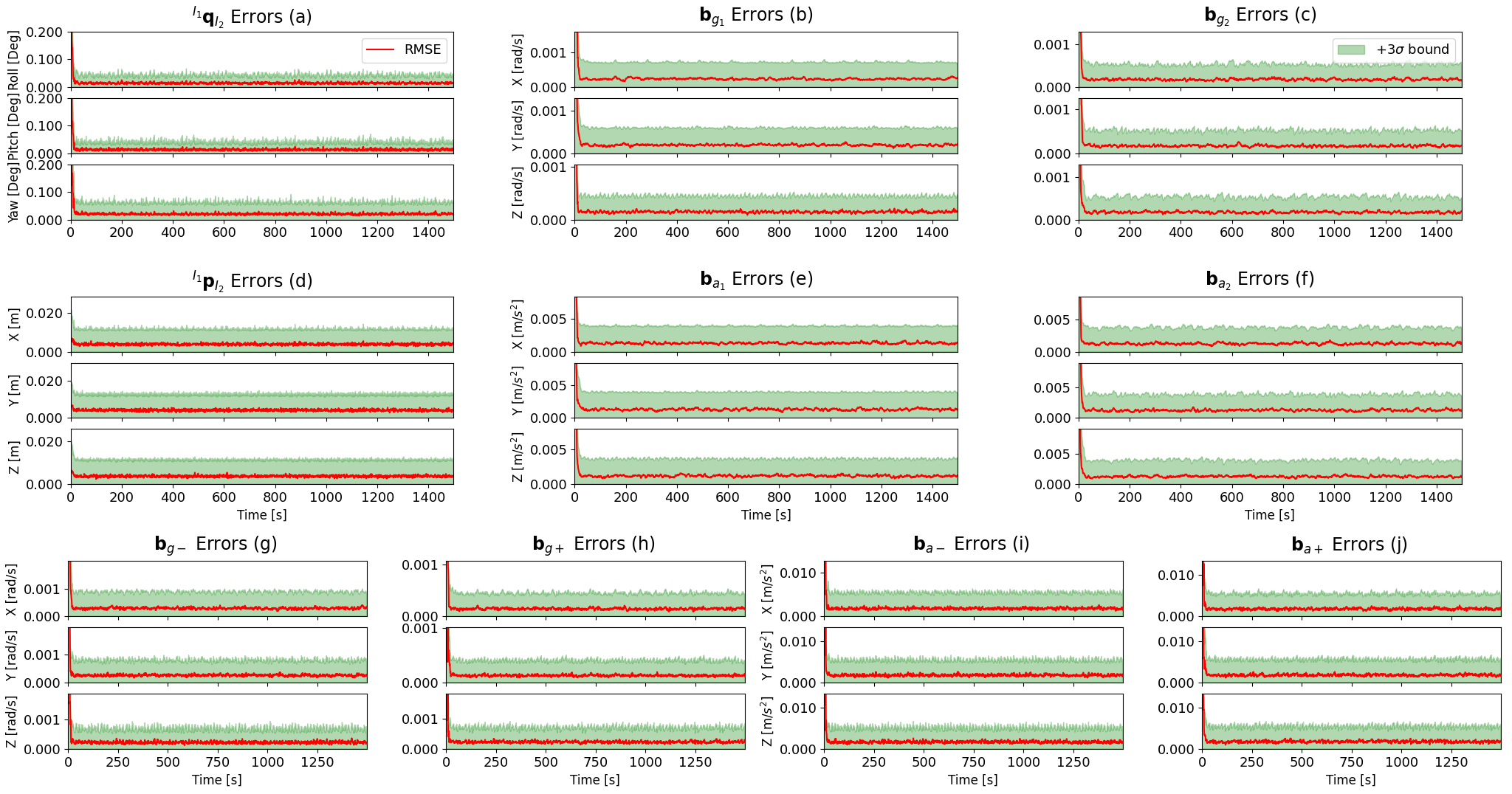}
		\caption{The RMSE and 3\(\sigma\) bound of the motion that the reference IMU moves on a horizontal plane and the target IMU moves only rotates with respect to the reference IMU in~Cell~VI-M of both Table~(\ref{table-dpq}) and Table~(\ref{table-dp}) with only \(\mathbf{dp}\) measurement. The red line represents RMSE, and
			the green background represents 3\(\sigma\) bound. The definition of each subplot is the same as in Fig.~\ref{dpq_static_pxyz_vxvyvz_wxwywz}. All states are
			observable.}
		\label{dp_plane_pxyz_wxyz}
	\end{figure*}

\begin{figure*}
		\centering
		\includegraphics[scale=0.335]{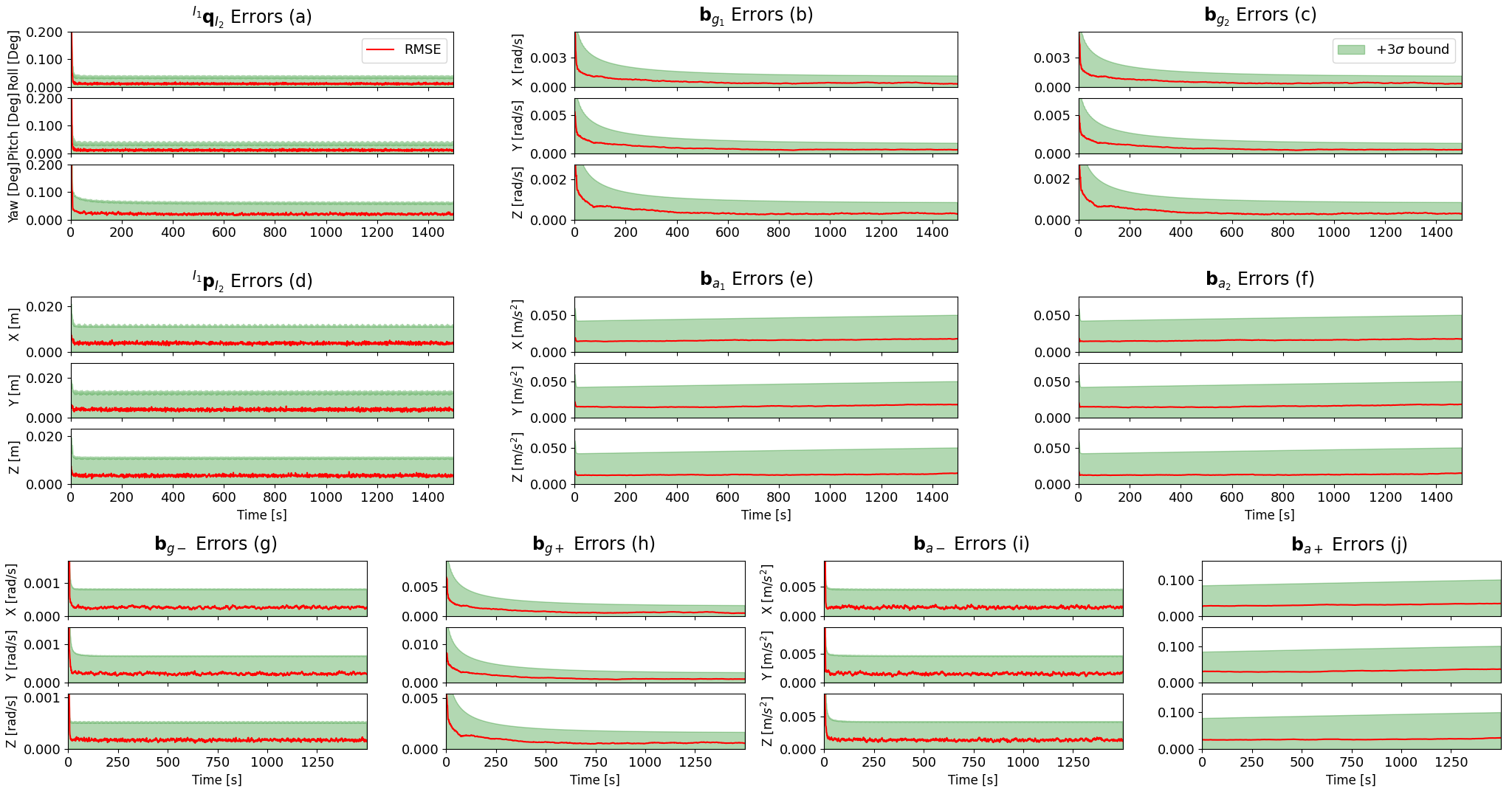}
		\caption{The RMSE and 3\(\sigma\) bound of the motion that the reference IMU moves on a horizontal plane and the target IMU remains still with respect to the reference IMU in~Cell~V-K of both Table~(\ref{table-dpq}) and Table~(\ref{table-dp}) with only \(\mathbf{dp}\) measurement. The red line represents RMSE, and
			the green background represents 3\(\sigma\) bound. The definition of each subplot is the same as in Fig.~\ref{dpq_static_pxyz_vxvyvz_wxwywz}. There are 3 unobservable directions: 3 directions of \(\mathbf{b}_{a+}\).}
		\label{dp_plane_pxyz}
	\end{figure*}

\begin{figure*}
		\centering
		\includegraphics[scale=0.335]{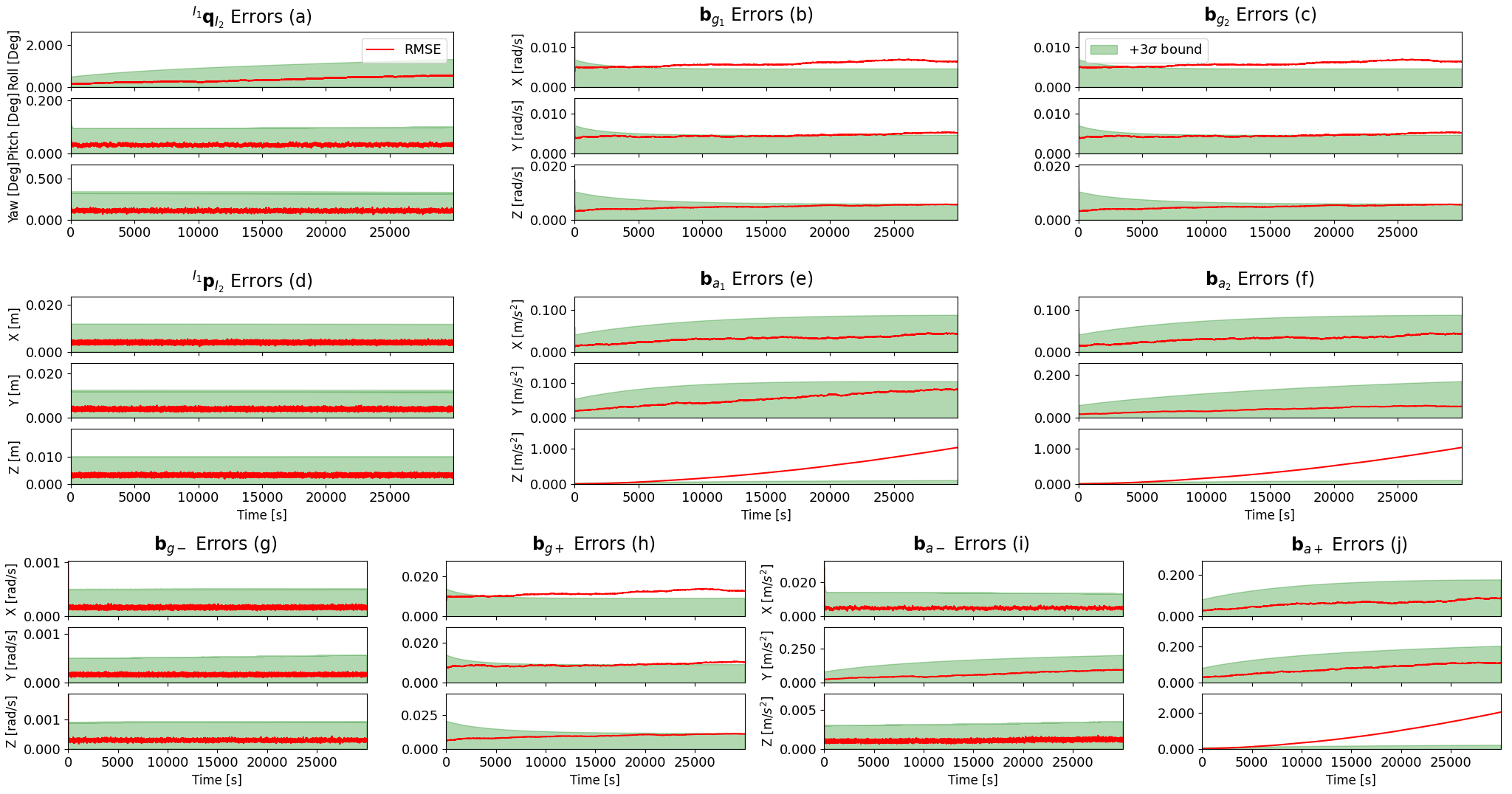}
		\caption{The RMSE and 3\(\sigma\) bound of the motion that the reference IMU moves along a straight line (the X direction) and the target IMU remains still with respect to the reference IMU in~Cell~III-K of both Table~(\ref{table-dpq}) and Table~(\ref{table-dp}) with only \(\mathbf{dp}\) measurement. The red line represents RMSE, and the
			green background represents 3\(\sigma\) bound. The definition of each subplot is the same as in Fig.~\ref{dpq_static_pxyz_vxvyvz_wxwywz}. There are 7 unobservable directions: 3 directions of \(\mathbf{b}_{g+}\), 3
			directions of \(\mathbf{b}_{a+}\) and the \(\boldsymbol{\theta}^{\boldsymbol{\beta}_1}\) direction (the errors of the roll angle and the Y direction of \(\mathbf{b}_{a_1}\) exhibit correlation).
		}
		\label{dp_straight_pxyz}
	\end{figure*}

\begin{figure*}
		\centering
		\includegraphics[scale=0.335]{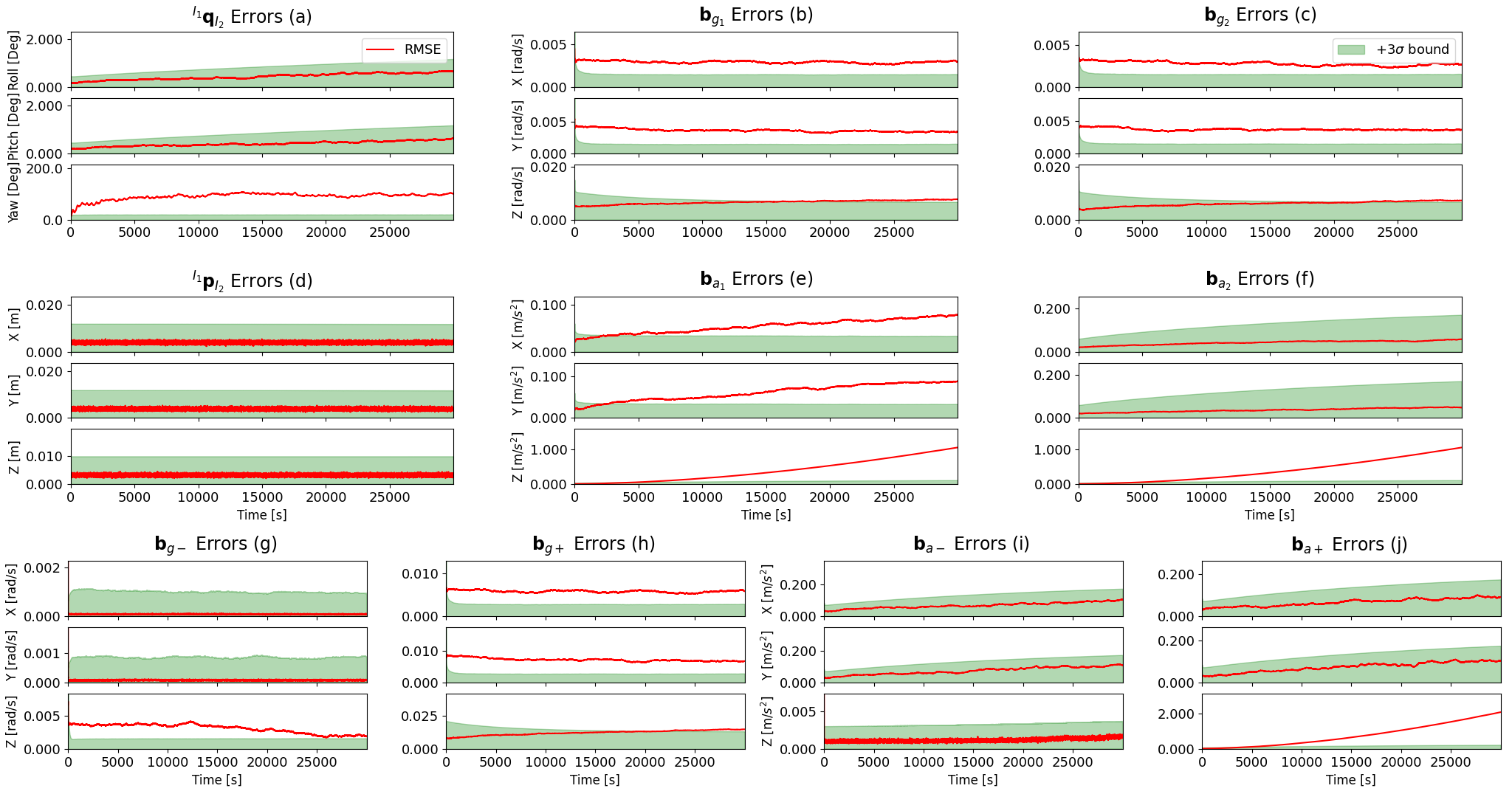}
		\caption{The RMSE and 3\(\sigma\) bound of the motion that the reference IMU remains still and the target IMU remains still with respect to the reference IMU in~Cell~I-K of both Table~(\ref{table-dpq}) and Table~(\ref{table-dp}) with only \(\mathbf{dp}\) measurement. The red line represents RMSE, and
			the green background represents 3\(\sigma\) bound. The definition of each subplot is the same as in Fig.~\ref{dpq_static_pxyz_vxvyvz_wxwywz}. There are 10 unobservable directions: 3 directions of \(\mathbf{b}_{g+}\), 3 directions of \(\mathbf{b}_{a+}\), the \(\boldsymbol{\theta}^{\boldsymbol{\alpha}}\) direction (the yaw), the \(\mathbf{b}_{g_1}^{\boldsymbol{\alpha}}\) direction (the Z direction of \(\mathbf{b}_{g_1}\)) and the
			\(\boldsymbol{\theta}^{\boldsymbol{\beta}_1}\) (the errors of the roll angle and the Y direction of \(\mathbf{b}_{a_1}\) exhibit correlation) and \(\boldsymbol{\theta}^{\boldsymbol{\beta}_2}\) (the errors of the pitch angle and the X direction of \(\mathbf{b}_{a_1}\) exhibit correlation) directions.}
		\label{dp_static_static}
	\end{figure*}

	 	When there is only \(\mathbf{dp}\) measurement, comparing Fig.~\ref{dp_static_pxyz_vxvyvz_wxwywz} (Cell~I-S in Table~(\ref{table-dpq}) and (\ref{table-dp})) and Fig.~\ref{dp_plane_pxyz_wxyz} (Cell~V-M in Table~(\ref{table-dpq}) and (\ref{table-dp})) to Fig.~\ref{dpq_static_pxyz_vxvyvz_wxwywz} and Fig.~\ref{dpq_plane_pxyz_wxyz}, all states remain observable. From Fig.~\ref{dp_plane_pxyz} (Cell~V-K in Table~(\ref{table-dpq}) and (\ref{table-dp})) and Fig.~\ref{dpq_plane_pxyz}, the unobservable directions are the same as well.
	However, separately comparing Fig.~\ref{dp_straight_pxyz} (Cell~III-K in Table~(\ref{table-dpq}) and (\ref{table-dp})) and Fig.~\ref{dp_static_static} (Cell~I-K in Table~(\ref{table-dpq}) and (\ref{table-dp})) to Fig.~\ref{dpq_straight_pxyz}  and Fig.~\ref{dpq_static_static}, some
	additional unobservable directions arise.
	From Fig.~\ref{dp_straight_pxyz}, the additional unobservable direction is \(\boldsymbol{\theta}^{\boldsymbol{\beta}_1}\), and the errors of the roll angle (Fig.~\ref{dp_straight_pxyz}(a)) and the Y direction of \(\mathbf{b}_{a_1}\) (Fig.~\ref{dp_straight_pxyz}(e)) exhibit correlation.
	From Fig.~\ref{dp_static_static}, when both IMUs are still,
	the number of unobservable directions increases significantly. In
	addition to the unobservable directions of \(\mathbf{b}_{g+}\) and \(\mathbf{b}_{a+}\) shown
	in Fig.~\ref{dp_static_static}(h) and Fig.~\ref{dp_static_static}(j), the yaw angle error (\(\boldsymbol{\theta}^{\boldsymbol{\alpha}}\)) grows at a very fast rate and
	far exceeds the 3\(\sigma\) bound (Fig.~\ref{dp_static_static}(a)). And the Z direction of \(\mathbf{b}_{g_1}\) (\(\mathbf{b}_{g_1}^{\boldsymbol{\alpha}}\)) also
	becomes unobservable (Fig.~\ref{dp_static_static}(b)). What's even more interesting is that we can
	observe the errors in roll and pitch angles (Fig.~\ref{dp_static_static}(a)) are correlated with the
	errors in the Y and X directions of \(\mathbf{b}_{a_1}\) (Fig.~\ref{dp_static_static}(e)) (\(\boldsymbol{\theta}^{\boldsymbol{\beta}_1} \) and \(\boldsymbol{\theta}^{\boldsymbol{\beta}_2} \) ), respectively.
	
		In summary, all simulation results are as expected. As a result, the simulation results perfectly validate our observability analysis in Sect. \ref{unobservable-directions-under-special-motions}.
		
		\vspace{10cm}
	
	\FloatBarrier
	\section{Experimental Results \label{7-experimental-results}}
	
	In the preceding section, we have verified that there are some unobservable directions under special motions for the Dual-IMU state estimation through simulations. In this section, we design experiments to showcase the performance of the Dual-IMU system in real-world applications.
	
\begin{figure*}[!]
		\centering
		\includegraphics[scale=0.55]{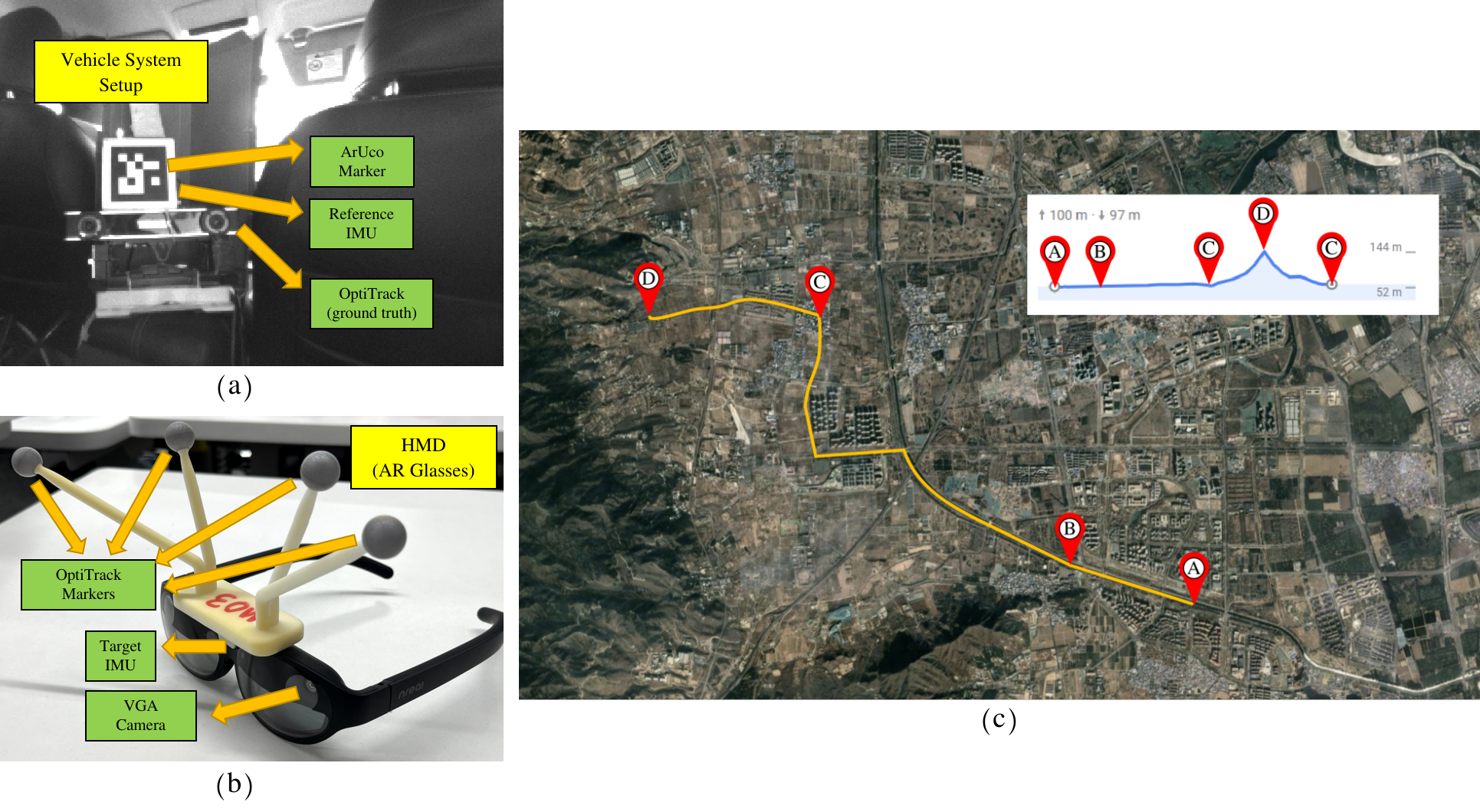}
		
		\caption{The key experimental sections. (a) illustrates the installation arrangement of the reference IMU, the ArUco marker, and the OptiTrack system inside the vehicle. (b) depicts the HMD equipped with OptiTrack markers used in the experiment. (c) is the route of the vehicle's movement. The yellow line
			represents the travel route, the small chart indicates altitude changes
			along the route, and A, B, C, D are key anchor points in the route.}
		\label{experimental_fig}
	\end{figure*}

	In our experiment, both the reference and target IMUs are consumer-grade MEMS-IMUs, where the reference IMU (Fig.~\ref{experimental_fig}(a)) is attached to a moving vehicle and the target IMU comes from an HMD (Augmented Reality (AR) glasses) (Fig.~\ref{experimental_fig}(b)) with two VGA global shutter cameras.
	Images from the HMD’s left camera of a car-fixed ArUco marker (10 cm) are
	processed by OpenCV's built-in algorithms \cite{refopencv} to calculate relative poses, serving as \(\mathbf{dp}\) and \(\mathbf{dq}\) measurements.
	 The ground truth is provided by an OptiTrack \cite{refopti} system (model: V120-DUO) attached to the vehicle, which calculates the HMD's pose by observing multiple markers mounted on the HMD. The ground truth has a rotation accuracy of less than 1 degree and a position accuracy of less than 1.5 mm. Our estimation implementation is based on EKF, where each prediction and update cycle takes 20$\mu s$ (ArUco marker processing time excluded) on an Intel (i7, 2.3GHz) processor.
	
	The vehicle's route is depicted in Fig.~\ref{experimental_fig}(c). In this route taking about 30 minutes, the
	vehicle's motion is highly diverse. The trajectory starts at
	anchor point A, passes through points B, C, and D, and finally returns
	to point C before stopping. First, the vehicle remains still at
	anchor point A for approximately 5 minutes. Next, the vehicle turns around and moves along a straight line from anchor point A to B. Afterward, there is a segment of normal
	driving on a flat road with both left and right turns from anchor point
	B to C. Referring to the small chart, the altitude increases
	continuously from anchor point C to D, indicating that the vehicle is
	going uphill. Finally, after remaining still again at anchor point
	D for approximately 5 minutes, the vehicle goes downhill and returns to
	anchor point C, concluding the trajectory.

	We drive the vehicle along the route twice, and a person
	sitting inside the vehicle and wearing the HMD exhibits
	two different motion patterns:  
	attempting to remain still and moving normally. For these two patterns, the comparison of the HMD's linear 
	and angular velocity with respect to the vehicle, as calculated from the
	ground truth, is presented in Table~(\ref{motion_model}). The velocity differences between the two patterns are significant. When the vehicle is stationary, for the pattern where the HMD remains still, we can observe that the HMD maintains a nearly perfect state of relative stillness with respect to the vehicle, with an average linear velocity of only 7 mm per second, significantly less than the 169 mm per second in the normal motion pattern of the HMD.
	When the vehicle is moving, the HMD unavoidably experiences movement along with the vehicle's sway. Consequently, the statistical values of linear and angular velocities for both patterns increase, but the distinction remains clear.

	\begin{table*}[!htb]
		\caption{Motion Quantities of the HMD in Two Motion Patterns}
		\center{
			\renewcommand{\arraystretch}{1.2}
			\begin{threeparttable}
			\begin{tabular}{|c|c|c|c|}
				\hline
				\textbf{HMD Motion Patterns}          & \textbf{       HMD Motion Quantities}         & \textbf{Vehicle Still} & \textbf{Vehicle Move} \\ \hline
				\multirow{2}{*}{Remains Still}    & Linear Velocity [m/s] & 0.007                  & 0.042                 \\ \cline{2-4} 
				& Angular Velocity [rad/s] & 0.045                  & 0.192                 \\ \hline
				\multirow{2}{*}{Moves Normally} & Linear Velocity [m/s] & 0.169                  & 0.199                 \\ \cline{2-4} 
				& Angular Velocity [rad/s] & 0.788                  & 1.063                 \\ \hline
			\end{tabular}

			 Mean of the HMD's linear and angular velocity with respect to the vehicle in two test cases. Each case corresponds to a different motion model of the HMD: remain still and move normally.
			\end{threeparttable} 
		}

		\label{motion_model}
	\end{table*}

	In each motion model, we explore 3 setups based on
	whether the initial biases are calibrated and whether there is
	\(\mathbf{dq}\) measurement: 1)
	with uncalibrated biases (ub) and both
	\(\mathbf{dp}\) and \(\mathbf{dq}\) measurements (ub-\(\mathbf{dpdq}\)), 2) with uncalibrated biases and only \(\mathbf{dp}\) measurement (ub-\(\mathbf{dp}\)), 3) with calibrated biases (cb) and only \(\mathbf{dp}\) measurement
	(cb-\(\mathbf{dp}\)). The results of setup with calibrated biases and both
	\(\mathbf{dp}\) and \(\mathbf{dq}\) measurements (cb-\(\mathbf{dpdq}\)) are almost the same with ub-\(\mathbf{dpdq}\), since the relative biases (\(\mathbf{b}_{g-}\) and \(\mathbf{b}_{a-}\)) converge immediately with both \(\mathbf{dp}\) and \(\mathbf{dq}\) measurements, and is omitted here. 
	
\begin{figure*}[!]
			\centering
			\includegraphics[scale=0.25]{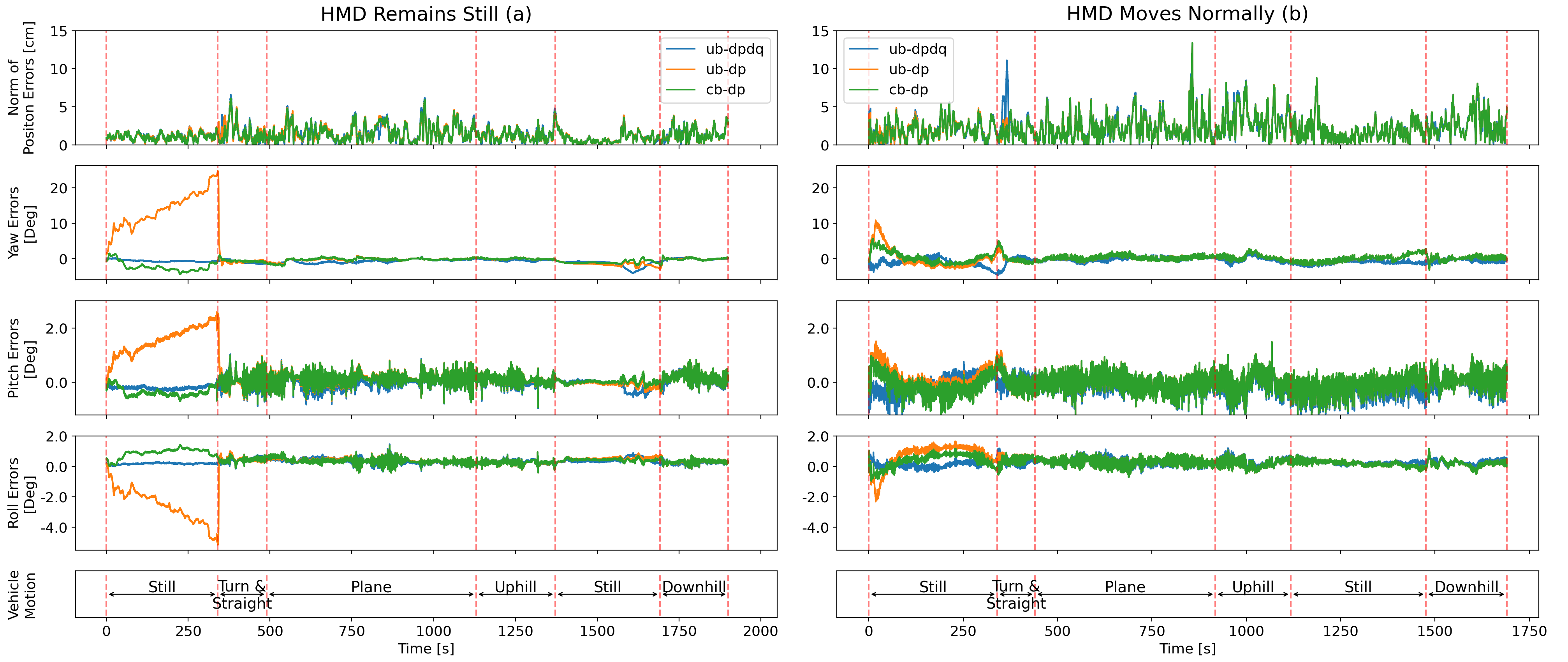}
			\caption{Comparison of relative pose errors under different motions of the HMD and the vehicle, as well as under different estimation setups. 
				Specifically, for the initial biases, we consider uncalibrated biases (ub) and calibrated biases (cb). As for the measurements used, we evaluate the system with both \(\mathbf{dp}\) and \(\mathbf{dq}\) (dpdq) and with \(\mathbf{dp}\) only~(dp). Note that, the results of cb-dpdq are almost identical with ub-dpdq, and are omitted. With both \(\mathbf{dp}\) and \(\mathbf{dq}\), the errors of relative pose are always very small. With only \(\mathbf{dp}\) and uncalibrated biases, when the vehicle is still and if the HMD remains still, the angle errors continue to increase, but if the HMD moves normally, the error of yaw briefly increases initially and then quickly decreases to a reasonable range. And with only \(\mathbf{dp}\) but calibrated biases, the pose errors are also small.}
			\label{error}
		\end{figure*}

	Since relative position does not hold the same
	special significance in different directions as relative rotation does,
	relative position errors are represented using the norm of the errors.
	The performance of relative pose errors under different cases is
	depicted in Fig.~\ref{error}. Simultaneously, the RMSE of the four error terms are summarized in Table~(\ref{RMS}).

	As shown in Fig.~\ref{error}, the errors of relative pose estimates
	with both dp and dq measurements are consistently small that the RMSE of position is less than 3 cm and the RMSE of the yaw angle is approximately 1 degree (Table~(\ref{RMS})),
	because the relative pose is directly measured and hence
	observable from Table~(\ref{table-dpq}).
	%
	Meanwhile, with only dp measurement, the pose errors are largely affected by the motion periods.
	From Fig.~\ref{error}(a), in the initial phase, both the vehicle and HMD remain still, and from Cell I-K in Table~(\ref{table-dp}), the relative orientation are unobservable. Hence, we can see that with uncalibrated biases, the angle errors, especially the yaw (\(\boldsymbol{\theta}^{\boldsymbol{\alpha}}\)) error, continue to increase to a big amount, while
	with calibrated biases, the errors still exist but are smaller.  This also indicates that at this stage, biases (e.g., \({\mathbf{b}_{g_1}^{\boldsymbol{\alpha}}}\)) are not observable.
	%
	Next, the vehicle starts moving (turn and straight), which corresponds to planar motion in~Cell~V-K of Table~(\ref{table-dp}) and hence the relative orientation becomes
	observable, then the angle errors quickly converge.
	Later on, when both IMUs come to the second stationary period, since gyroscope biases have already converged due to previous motions corresponding to Cell~VII-K in Table~(\ref{table-dpq}) and (\ref{table-dp}), the accuracy remains high.
	Moreover, when comparing Fig.~\ref{error}(b) with (a),we can see that with uncalibrated biases, the HMD moving normally (Row-S in Table~(\ref{table-dp})) results in better
	relative orientation estimates than staying still (Row-K in Table~(\ref{table-dp})). The error performance is consistent with Table~(\ref{RMS}).
	
	
	 It's worth mentioning that the position errors of the HMD moving normally are slightly larger than staying still from Fig.~\ref{error} and Table~(\ref{RMS}), because the tracking quality of the ArUco marker degrades (Table~(\ref{RMS})) as the HMD's camera moves faster. For the same reasons, with both \(\mathbf{dp}\) and \(\mathbf{dq} \) measurements, the angle errors when the HMD moves normally are comparable to, or slightly greater than, when the HMD remains still.

	To summarize, the experimental results align well with our observability analysis in Sect. \ref{unobservable-directions-under-special-motions}. Overall, the proposed
	algorithm achieves satisfactory relative localization on real-world data and runs extremely fast (20$\mu s$ per process). To further improve the localization accuracy, we can enlarge the marker's size, design special markers, use outside-in tracking systems, or introduce additional observations to the vehicle.

		\begin{table*}[!]
			\small
			\caption{RMSE under Different Estimation Setups and HMD's Motion Patterns}
			\center{
				\renewcommand{\arraystretch}{1.2}
				\begin{threeparttable}
					
					\begin{tabular}{|c|c|p{3.2cm}<{\centering}|p{3.2cm}<{\centering}|}
						\hline
						\textbf{Setups}        & \textbf{Error Terms}        & \textbf{HMD Remains Still} & \textbf{HMD Moves Normally} \\ \hline
						\multirow{4}{*}{\makecell[c]{*Poses from \\ ArUco Marker}} & Norm of Position Errors [cm] & 1.84             & 3.37             \\ \cline{2-4} 
						& Roll Errors [Deg]           & 0.27             & 0.57             \\ \cline{2-4} 
						& Pitch Errors [Deg]          & 1.94             & 3.70             \\ \cline{2-4} 
						& Yaw Errors [Deg]            & 2.35             & 5.16             \\ \hline\hline
						\multirow{4}{*}{\makecell[c]{Uncalibrated Biases \\  \(\mathbf{dp}\) + \(\mathbf{dq}\)}}& Norm of Position Errors [cm] & 1.64             & 2.73             \\ \cline{2-4} 
						& Roll Errors [Deg]           & 0.34             & 0.32             \\ \cline{2-4} 
						& Pitch Errors [Deg]          & 0.19             & 0.31             \\ \cline{2-4} 
						& Yaw Errors [Deg]            & 1.00             & 1.11             \\ \hline
						\multirow{4}{*}{\makecell[c]{ Uncalibrated Biases  \\  Only \(\mathbf{dp}\)}} & Norm of Position Errors [cm] & 1.61             & 2.65             \\ \cline{2-4} 
						& Roll Errors [Deg]           & 1.25             & 0.55             \\ \cline{2-4} 
						& Pitch Errors [Deg]          & 0.71             & 0.34             \\ \cline{2-4} 
						& Yaw Errors [Deg]            & 6.32             & 1.53             \\ \hline
						\multirow{4}{*}{\makecell[c]{Calibrated Biases \\  Only \(\mathbf{dp}\)}} & Norm of Position Errors [cm] & 1.58             & 2.64            \\ \cline{2-4} 
						& Roll Errors [Deg]           & 0.52             & 0.38             \\ \cline{2-4} 
						& Pitch Errors [Deg]          & 0.23             & 0.32             \\ \cline{2-4} 
						& Yaw Errors [Deg]            & 1.28             & 1.09             \\ \hline

					\end{tabular}
					
					With both \(\mathbf{dp}\) and \(\mathbf{dq}\), the errors of relative pose are all very small. With only \(\mathbf{dp}\) and uncalibrated biases, the RMSE of each rotation angle is greater when the HMD remains static compared to when it moves normally. And with only \(\mathbf{dp}\) but calibrated biases, the pose errors are also small. The first method (marked with *) in the first row differs from the other three as it calculates the pose (the \(\mathbf{dp}\) and \(\mathbf{dq}\) measurements) using only the ArUco marker.
				\end{threeparttable}
			}
			\label{RMS}
		\end{table*}
		
	\FloatBarrier
	
	\section{Conclusions \label{8-conclusions}}
	In this paper, we presented a Dual-IMU state estimation system, which jointly estimates the relative pose, velocity, and all four biases from the two IMUs.
	For this system, we proved that it is locally weakly observable when both IMUs undergo general motion, even with only relative position observation. 
	Furthermore, we have identified the system's unobservable directions across various special motions and categorized these motions into two groups: those resulting in the relative orientation becoming unobservable, and those potentially leading to only biases being unobservable.
	Through numerical simulations and real-world experiments, our observability findings were validated based on the estimation error and consistency, where the relative orientation estimates were degraded or unaffected, depending on the corresponding motion category.
	Lastly, in the real world, our system was able to localize an HMD with respect to a moving vehicle with high accuracy and computational efficiency.
	
	\section{Appendix\label{appendix}}

	\subsection*{Appendix A: Derivation of System State Equations \label{appendix-a}}
	
	For clarity, some subscripts and superscripts are omitted here. For
	instance, we use \(\mathbf{p}_1\), \(\mathbf{p}_2\), \(\mathbf{p}\) to denote 
	\(\prescript{{G}}{}{\mathbf{p}}_{I_1}\), \(\prescript{{G}}{}{\mathbf{p}}_{I_2}\), \(\prescript{I_1}{}{}{\mathbf{p}}_{I_2}\), \(\mathbf{C}_1\), \(\mathbf{C}_2\), \(\mathbf{C}\)
	for 
	\( \mathbf{C}\left(\prescript{{G}}{}{\mathbf{q}}_{I_1} \right)\), \(\mathbf{C}\left(\prescript{{G}}{}{\mathbf{q}}_{I_2} \right)\), \(\mathbf{C}\left(\prescript{I_1}{}{}{\mathbf{q}}_{I_2} \right)\),
	\(\mathbf{a}_1\), \(\mathbf{a}_2\) for
	\(\prescript{I_1}{}{}{\mathbf{a}}\), \(\prescript{{I_2}}{}{\mathbf{a}}\), \(\boldsymbol{\omega}_1\), \(\boldsymbol{\omega}_2\) for
	\(\prescript{I_1}{}{}{\boldsymbol{\omega}}\), \(\prescript{I_2}{}{\boldsymbol{\omega}}\). 
	
	The equation \eqref{dotp} - \eqref{1-2} become:
	\begin{align} 
		\dot{\mathbf{p}}(t) &= \mathbf{v}(t) \label{ndotp} \\
		\dot{\mathbf{v}}(t) &=\ \mathbf{C} \mathbf{a}_2(t) - \mathbf{a}_1(t) -2 \boldsymbol{\omega}_1(t) \times \mathbf{v}(t) \nonumber \\
		& - \boldsymbol{\omega}_1(t)  \times\left( \boldsymbol{\omega}_1(t) \times \mathbf{p}(t) \right) - \dot{\boldsymbol{\omega}}_1(t) \times \mathbf{p}(t) \label{ndotv} \\
		\dot{\mathbf{q}}(t) 
		&= \frac{1}{2}(\mathbf{q}(t) \otimes \bar{\boldsymbol{\omega}}_2(t) - \bar{\boldsymbol{\omega}}_1(t) \otimes \mathbf{q} (t))  \label{ndotq}\\
		\dot{\mathbf{b}}_{g_1}(t) &= \mathbf{n}_{w{g_1}},\  \dot{\mathbf{b}}_{g_2}(t) = \mathbf{n}_{w{g_2}} \\
		\dot{\mathbf{b}}_{a_1}(t) &= \mathbf{n}_{w{a_1}},\  \dot{\mathbf{b}}_{a_2}(t) = \mathbf{n}_{w{a_2}} 
	\end{align}
	where
	\(\bar{\boldsymbol{\omega}}_1 \triangleq \begin{bmatrix} 0 &\boldsymbol{\omega}_1^T \end{bmatrix}^T \) and \(\bar{\boldsymbol{\omega}}_2 \triangleq \begin{bmatrix} 0 &\boldsymbol{\omega}_2^T \end{bmatrix}^T \).
		
	The dervation process of \eqref{ndotp} and \eqref{ndotv}:
	
	\begin{align}
		\mathbf{p} &= \mathbf{C}_1^T(\mathbf{p}_2 - \mathbf{p}_1 ) \\
		\mathbf{v} = \dot{\mathbf{p}} &=  \frac{d\mathbf{C}_1^T}{dt}(\mathbf{p}_2 - \mathbf{p}_1 )  + \mathbf{C}_1^{T}(\mathbf{v}_2 - \mathbf{v}_1 ) \nonumber\\
		&= \lim_{dt\rightarrow 0} \frac{[\mathbf{C}_1(\mathbf{I} + [{\delta \boldsymbol{\theta}_1}_\times])]^T - \mathbf{C}_1^T}{dt}(\mathbf{p}_2 - \mathbf{p}_1 )  + {\mathbf{C}_1^T}(\mathbf{v}_2 - \mathbf{v}_1) \nonumber\\
		&= \lim_{dt\rightarrow 0} \frac{-[{\delta \boldsymbol{\theta}_1}_\times]\mathbf{C}_1^T}{dt}(\mathbf{p}_2 - \mathbf{p}_1 )  + {\mathbf{C}_1^T}(\mathbf{v}_2 - \mathbf{v}_1)  \nonumber\\
		&=  -[\boldsymbol{\omega}_{1\times}]({\mathbf{C}_1^T}(\mathbf{p}_2 - \mathbf{p}_1))  + {\mathbf{C}_1^T}(\mathbf{v}_2 - \mathbf{v}_1) \nonumber\\
		&=   -\boldsymbol{\omega}_1\times \mathbf{p} + {\mathbf{C}_1^T}(\mathbf{v}_2 - \mathbf{v}_1) \label{def-v} 
\end{align}
then:
\begin{align}
		\dot{\mathbf{v}} &= \frac{d \mathbf{C}_1^T(\mathbf{v}_2 - \mathbf{v}_1) }{dt}- \dot{\boldsymbol{\omega}}_1\times \mathbf{p} - \boldsymbol{\omega}_1 \times \dot{\mathbf{p}} \nonumber\\
		&= \mathbf{C}_1^T(\dot{\mathbf{v}}_2 - \dot{\mathbf{v}}_1)- \boldsymbol{\omega}_1 \times (\mathbf{C}_1^T(\mathbf{v}_2 - \mathbf{v}_1)) - \dot{\boldsymbol{\omega}}_1\times \mathbf{p} - \boldsymbol{\omega}_1 \times \dot{\mathbf{p}} \nonumber\\
		&= \mathbf{C}_1^T(\dot{\mathbf{v}}_2 - \dot{\mathbf{v}}_1)- \boldsymbol{\omega}_1 \times (\mathbf{C}_1^T(\mathbf{v}_2 - \mathbf{v}_1) -\boldsymbol{\omega}_1\times \mathbf{p} )  -\boldsymbol{\omega}_1\times(\boldsymbol{\omega}_1\times \mathbf{p}) -\dot{\boldsymbol{\omega}}_1\times \mathbf{p} - \boldsymbol{\omega}_1 \times \dot{\mathbf{p}} \nonumber\\
		&= \mathbf{C}_1^T(\dot{\mathbf{v}}_2 - \dot{\mathbf{v}}_1)- \boldsymbol{\omega}_1 \times \mathbf{v}  -\boldsymbol{\omega}_1\times(\boldsymbol{\omega}_1\times \mathbf{p}) -\dot{\boldsymbol{\omega}}_1\times \mathbf{p} - \boldsymbol{\omega}_1 \times \mathbf{v} \nonumber\\
		&= \mathbf{C}_1^T(\dot{\mathbf{v}}_2 - \dot{\mathbf{v}}_1)- 2\boldsymbol{\omega}_1 \times \mathbf{v}  -\boldsymbol{\omega}_1\times(\boldsymbol{\omega}_1\times \mathbf{p}) -\dot{\boldsymbol{\omega}}_1\times \mathbf{p} \nonumber\\
		&= \mathbf{C}_1^T(\mathbf{C}_2\mathbf{a}_2 + \mathbf{g} - \mathbf{C}_1\mathbf{a}_1 - \mathbf{g})- 2\boldsymbol{\omega}_1 \times \mathbf{v}  -\boldsymbol{\omega}_1\times(\boldsymbol{\omega}_1\times \mathbf{p}) -\dot{\boldsymbol{\omega}}_1\times \mathbf{p} \nonumber\\
		&= \mathbf{C}\mathbf{a}_2 - \mathbf{a}_1 - 2\boldsymbol{\omega}_1 \times \mathbf{v}  -\boldsymbol{\omega}_1\times(\boldsymbol{\omega}_1\times \mathbf{p}) -\dot{\boldsymbol{\omega}}_1\times \mathbf{p} 
	\end{align}

	The dervation process of \(\eqref{ndotq}\):
	
	\begin{align} 
		\dot{\mathbf{q}} = \left(\mathbf{q}_1^{-1}  \dot{\otimes} \mathbf{q}_2\right)
		& = \mathbf{q}_1^{-1} \otimes \dot{\mathbf{q}}_2  +  \dot{\mathbf{q}}_1^{-1} \otimes  \mathbf{q}_2 \nonumber\\
		& = \frac{1}{2}\left(\mathbf{q}_1^{-1} \otimes \mathbf{q}_2 \otimes \bar{\boldsymbol{\omega}}_2 - \bar{\boldsymbol{\omega}}_1\otimes \mathbf{q}_1^{-1}   \otimes \mathbf{q}_2 \right) \nonumber\\
		& = \frac{1}{2}\left(\mathbf{q} \otimes \bar{\boldsymbol{\omega}}_2 -  \bar{\boldsymbol{\omega}}_1 \otimes \mathbf{q}\right)  
	\end{align}

	\subsection*{Appendix B: Derivation of Continuous-Time Error-State Equations \label{appendix-b}}
	
	Using the same symbol abbreviation method as in Appendix A.
	The system kinematics equation in \eqref{dotp} to \eqref{1-2} can be divided into linear terms and error terms:
	
	Expand \(\dot{\mathbf{p}}\):
	
	\begin{align}
		\dot{\mathbf{p}} &= \dot{\hat{\mathbf{p}}} + \dot{\tilde{\mathbf{p}}} = \hat{\mathbf{v}} + \tilde{\mathbf{v}} \nonumber \\
		\dot{\hat{\mathbf{p}}} &= \hat{\mathbf{v}} \\
		\dot{\tilde{\mathbf{p}}} &= \tilde{\mathbf{v}} 
	\end{align}
	
	Expand all biases:
	\begin{align}
		\dot{\mathbf{b}}_{g_1} &= \dot {\widetilde{\mathbf{b}}}_{g_1} + \dot {\hat{\mathbf{b}}}_{g_1} =  \mathbf{n}_{w{g_1}} + 0 \\
		\dot{\mathbf{b}}_{g_2} &= \dot {\widetilde{\mathbf{b}}}_{g_2} + \dot {\hat{\mathbf{b}}}_{g_2} =  \mathbf{n}_{w{g_2}} + 0 \\
		\dot{\mathbf{b}}_{a_1} &= \dot {\widetilde{\mathbf{b}}}_{a_1} + \dot {\hat{\mathbf{b}}}_{a_1} =  \mathbf{n}_{w{a_1}} + 0 \\
		\dot{\mathbf{b}}_{a_2} &= \dot {\widetilde{\mathbf{b}}}_{a_2} + \dot {\hat{\mathbf{b}}}_{a_2} =  \mathbf{n}_{w{a_2}} + 0 
	\end{align}
	
	Expand \(\dot{\mathbf{v}}\) \eqref{ndotv}:
	
	- Term of \(-\mathbf{a}_1\):
	\begin{equation}
		-\hat{\mathbf{a}}_{1} - \delta \mathbf{a}_1 
	\end{equation}
	
	- Term of \(\mathbf{C}\mathbf{a}_2\):
	\begin{align}
		\hat{\mathbf{C}}(\mathbf{I}_3+[\delta \boldsymbol{\theta}_\times])(\hat{\mathbf{a}}_{2}+\delta \mathbf{a}_{2}) 
		& = \hat{\mathbf{C}}\hat{\mathbf{a}}_2 + \hat{\mathbf{C}}[\delta \boldsymbol{\theta}_\times](\hat{\mathbf{a}}_{2}+\delta \mathbf{a}_{2}) + \hat{\mathbf{C}}\delta \mathbf{a}_2 \nonumber \\
		&= \hat{\mathbf{C}}\hat{\mathbf{a}}_2-\hat{\mathbf{C}}[{\hat{\mathbf{a}}}_{2\times}]\delta \boldsymbol{\theta} + \hat{\mathbf{C}} \delta \mathbf{a}_2 + O(\delta^2) 
	\end{align}
	
	- Term of \(-2\boldsymbol{\omega}_1 \times \mathbf{v}\):
	\begin{align}
		-2((\hat{\boldsymbol{\omega}}_1 + \delta \boldsymbol{\omega}_1) \times (\hat{\mathbf{v}}+\tilde{\mathbf{v}}))
		&= -2( \hat{\boldsymbol{\omega}}_1\times \hat{\mathbf{v}} + \hat{\boldsymbol{\omega}}_1 \times \tilde{\mathbf{v}} + \delta {\boldsymbol{\omega}}_1 \times \hat{\mathbf{v}} + \delta {\boldsymbol{\omega}}_1 \times \tilde{\mathbf{v}}) \nonumber \\
		&= -2\hat{\boldsymbol{\omega}}_1 \times \hat{\mathbf{v}} - 2\hat{\boldsymbol{\omega}}_1 \times \tilde{\mathbf{v}} +   2\hat{\mathbf{v}} \times \delta\boldsymbol{\omega}_1 +O(\delta ^2) 
	\end{align}
	
	- Term of
	\(-\boldsymbol{\omega}_1 \times (\boldsymbol{\omega}_1 \times \mathbf{p})\):
	\begin{align} 
		-((\hat{\boldsymbol{\omega}}_1+\delta \boldsymbol{\omega}_1 ) \times ( (\hat{\boldsymbol{\omega}}_1 +\delta \boldsymbol{\omega}_1 ) \times(\hat{\mathbf{p}} + \tilde{\mathbf{p}} )))
		=&  -\hat{\boldsymbol{\omega}}_1 \times (\hat{\boldsymbol{\omega}}_1 \times \hat{\mathbf{p}})  -\delta \boldsymbol{\omega}_1 \times (\hat{\boldsymbol{\omega}}_1 \times \hat{\mathbf{p}} )  \nonumber \\
		&- \hat{\boldsymbol{\omega}}_1 \times (\delta \boldsymbol{\omega}_1 \times \hat{\mathbf{p}}) - \hat{\boldsymbol{\omega}}_1 \times (\hat{\boldsymbol{\omega}}_1 \times \tilde{\mathbf{p}} )  + O(\delta^2) \nonumber\\
		 = &
		-\hat{\boldsymbol{\omega}}_1 \times (\hat{\boldsymbol{\omega}}_1 \times \hat{\mathbf{p}}) 
		+[(\hat{\boldsymbol{\omega}}_1 \times \hat{\mathbf{p}})_\times ] \delta \boldsymbol{\omega}_1 \nonumber \\
		& + [\hat{\boldsymbol{\omega}}_{1\times}]  [  \hat{\mathbf{p}}_\times] \delta \boldsymbol{\omega}_1
		- [\hat{\boldsymbol{\omega}}_{1\times}]^2 \tilde{\mathbf{p}} + O(\delta^2) 
	\end{align}
	
	- Term of \(-\dot{\boldsymbol{\omega}}_1 \times \mathbf{p}\):
	\begin{align}
		-\left(\frac{d(\hat{\boldsymbol{\omega}}_1 + \delta \boldsymbol{\omega}_1)}{dt} \times (\hat{\mathbf{p}} + \tilde{\mathbf{p}})   \right) &= -\dot{\hat{\boldsymbol{\omega}}}_1 \times \hat{\mathbf{p}} -\dot{\hat{\boldsymbol{\omega}}}_1 \times \tilde{\mathbf{p}} + \hat{\mathbf{p}} \times \frac{d \delta \boldsymbol{\omega}_1}{dt} + O(\delta^2) \nonumber\\
		&= -\dot{\hat{\boldsymbol{\omega}}}_1 \times \hat{\mathbf{p}} -\dot{\hat{\boldsymbol{\omega}}}_1 \times \tilde{\mathbf{p}} - \hat{\mathbf{p}} \times \frac{d \left(\tilde{\mathbf{b}}_{g1} + \mathbf{n}_{g_1}\right)}{dt} + O(\delta^2) \nonumber \\
		&= -\dot{\hat{\boldsymbol{\omega}}}_1 \times \hat{\mathbf{p}} -\dot{\hat{\boldsymbol{\omega}}}_1 \times \tilde{\mathbf{p}} - \hat{\mathbf{p}} \times \mathbf{n}_{w{g_1}} - \hat{\mathbf{p}} \times \dot{\mathbf{n}}_{g_1} + O(\delta^2)
	\end{align}	
	with
	\(\delta \mathbf{a}_1 = -\tilde{\mathbf{b}}_{a_1}-\mathbf{n}_{a_1},\delta \mathbf{a}_2 = -\tilde{\mathbf{b}}_{a_2}-\mathbf{n}_{a_2},\delta \boldsymbol{\omega}_1 = -\tilde{\mathbf{b}}_{g_1}-\mathbf{n}_{g_1},\delta \boldsymbol{\omega}_2 = -\tilde{\mathbf{b}}_{g_2}-\mathbf{n}_{g_2}\)  according to \eqref{ww} and \eqref{aa}.
	
	Then:  
	\begin{align}
		\dot{\tilde {\mathbf{v}}} &= - \mathbf{U}\tilde {\mathbf{p}} - 2 \hat{\boldsymbol{\omega}}_1 \times  \tilde {\mathbf{v}} - \hat{\mathbf{C}}[\hat{\mathbf{a}}_{2\times} ]\delta \boldsymbol{\theta}  - \mathbf{K}\tilde{\mathbf{b}}_{g_1} - \mathbf{K} \mathbf{n}_{g_1} +\tilde{\mathbf{b}}_{a_1} + \mathbf{n}_{a_1}  - \hat{\mathbf{C}}\tilde{\mathbf{b}}_{a_2} - \hat{\mathbf{C}} \mathbf{n}_{a_2}  - \hat{\mathbf{p}} \times \mathbf{n}_{wg_1} - \hat{\mathbf{p}} \times \dot{\mathbf{n}}_{g_1} \label{ddotv}
	\end{align}
	where
	\begin{align}
		\mathbf{K} &=2[ \hat{\mathbf{v}}_\times ]+[ \hat{\boldsymbol{\omega}}_{1\times} ][\hat{\mathbf{p}}_\times] + [( \hat{\boldsymbol{\omega}}_1 \times \hat{\mathbf{p}}  )_\times] \label{VVV}\\
		\mathbf{U} &= [ \hat{\boldsymbol{\omega}}_{1\times} ]^2 + [ \dot{\hat{\boldsymbol{\omega}}}_{1\times} ] \label{WWW}
	\end{align}
	same as \(\eqref{FF}\).
	
	Expand \(\dot{\mathbf{q}}\):

	\begin{align}
		\dot{\mathbf{q}} = (\hat{\mathbf{q}} \dot{ \otimes}  \delta \mathbf{q}) &= \frac{1}{2}(\mathbf{q} \otimes \bar{\boldsymbol{\omega}}_2 -   \bar{\boldsymbol{\omega}}_1 \otimes \mathbf{q})  \nonumber\\
		\hat{\mathbf{q}} \otimes \dot{\delta \mathbf{q}} +   \dot{\hat{\mathbf{q}}} \otimes \delta \mathbf{q} &= \frac{1}{2}(\mathbf{q}  \otimes \bar{\boldsymbol{\omega}}_2  - \bar{\boldsymbol{\omega}}_1 \otimes  \mathbf{q} ) \nonumber\\
		\hat{\mathbf{q}} \otimes \dot{\delta \mathbf{q}}  &= \frac{1}{2}(\mathbf{q}  \otimes \bar{\boldsymbol{\omega}}_2  -  \bar{\boldsymbol{\omega}}_1 \otimes  \mathbf{q}  -  (\hat{\mathbf{q}}  \otimes \bar{\hat{\boldsymbol{\omega}}}_{2} - \bar{\hat{\boldsymbol{\omega}}}_1  \otimes  \hat{\mathbf{q}} ) \otimes \delta\mathbf{q}  )\\
		 \nonumber \\
		\dot{\delta \mathbf{q}} &= \frac{1}{2}(\delta \mathbf{q}  \otimes \bar{\boldsymbol{\omega}}_2 -  \hat{\mathbf{q}}^{-1} \otimes  \bar{\boldsymbol{\omega}}_1 \otimes  \mathbf{q} -  \bar{\hat{\boldsymbol{\omega}}}_{2}  \otimes \delta \mathbf{q}  +   \hat{\mathbf{q}}^{-1} \otimes \bar{\hat{\boldsymbol{\omega}}}_{1} \otimes  \mathbf{q} )  \nonumber\\
		&= \frac{1}{2}(\delta \mathbf{q}  \otimes  \bar{\boldsymbol{\omega}}_2 -  \hat{\mathbf{q}}^{-1}  \otimes \begin{bmatrix}
			0 \\ \delta \boldsymbol{\omega}_1
		\end{bmatrix} \otimes  \mathbf{q} - \bar{\hat{\boldsymbol{\omega}}}_{2} \otimes \delta \mathbf{q} ) \nonumber\\
		&= \frac{1}{2}(\delta \mathbf{q}  \otimes  \bar{\boldsymbol{\omega}}_2 - \bar{\hat{\boldsymbol{\omega}}}_{2} \otimes \delta \mathbf{q} -    \begin{bmatrix}
			0 \\ \hat{\mathbf{C}}^T \delta \boldsymbol{\omega}_1
		\end{bmatrix} \otimes \delta \mathbf{q} )
	\end{align}
	where:
	\begin{align}
		\delta \mathbf{q} \otimes \bar{\boldsymbol{\omega}}_2 - \bar{\hat{\boldsymbol{\omega}}}_2\otimes \delta \mathbf{q} 
		&= 
		\begin{bmatrix}
			0 & -\delta \boldsymbol{\omega}_2^T \nonumber\\
			\delta \boldsymbol{\omega}_2 & -[(2 \hat{\boldsymbol{\omega}}_2+\delta \boldsymbol{\omega}_2)_\times] 
		\end{bmatrix}
		\begin{bmatrix} 1 \\ \delta \boldsymbol{\theta} / 2 \end{bmatrix} \\
		&=\begin{bmatrix}
			O(\delta^2) \\ \delta \boldsymbol{\omega}_2 - [\hat{\boldsymbol{\omega}}_{2\times}]\delta \boldsymbol{\theta} + O(\delta^2)
		\end{bmatrix} 
		\\
		\begin{bmatrix}
			0 \\ \hat{\mathbf{C}}^T \delta \boldsymbol{\omega}_1
		\end{bmatrix} \otimes \delta \mathbf{q}  &= \begin{bmatrix}
			0 & - (\hat{\mathbf{C}}^T \delta \boldsymbol{\omega}_1)^T \nonumber\\
			\hat{\mathbf{C}}^T \delta \boldsymbol{\omega}_1 & -[(\hat{\mathbf{C}}^T \delta \boldsymbol{\omega}_1 )_\times] 
		\end{bmatrix}
		\begin{bmatrix} 1 \\ \delta \boldsymbol{\theta} / 2 \end{bmatrix} \nonumber\\
		&= \begin{bmatrix}
			O(\delta^2) \\ \hat{\mathbf{C}}^T \delta \boldsymbol{\omega}_1  + O(\delta^2) 
		\end{bmatrix}
	\end{align}
	then:
	\begin{equation}
		\dot{\delta \boldsymbol{\theta}} =  
	-[\hat{\boldsymbol{\omega}}_{2\times}] \cdot \delta \boldsymbol{\theta} - \tilde{\mathbf{b}}_{g_2} - \mathbf{n}_{g_2} +\hat{\mathbf{C}}^T\tilde{\mathbf{b}}_{g_1} + \hat{\mathbf{C}}^T\mathbf{n}_{g_1}
	\end{equation}

	Therefore, the continuous-time error-state kinematics are:
	\begin{equation}
		\dot{\tilde{\mathbf{x}}}(t) = \mathbf{F}(t)\mathbf{x}(t) + \mathbf{G}(t)\mathbf{n}(t)
	\end{equation}
	where:
	\begin{align}
		\mathbf{F} &= 
		\begin{bmatrix}
			\mathbf{0}_3 & \mathbf{I}_3 & \mathbf{0}_3 & \mathbf{0}_3 & \mathbf{0}_3 & \mathbf{0}_3 & \mathbf{0}_3 \\
			-\mathbf{U} & -2[\hat{\boldsymbol{\omega}}_{1\times} ] & -\hat{\mathbf{C}}[\hat{\mathbf{a}}_{2\times} ]
			& -\mathbf{K} & \mathbf{0}_3 & \mathbf{I}_3 & -\hat{\mathbf{C}}\\
			\mathbf{0}_3 & \mathbf{0}_3 & -[\hat{\boldsymbol{\omega}}_{2\times} ]  & \hat{\mathbf{C}}^T & - \mathbf{I}_3 & \mathbf{0}_3 & \mathbf{0}_3\\
			\mathbf{0}_3 & \mathbf{0}_3 & \mathbf{0}_3 & \mathbf{0}_3 & \mathbf{0}_3 & \mathbf{0}_3 & \mathbf{0}_3 \\
			\mathbf{0}_3 & \mathbf{0}_3 & \mathbf{0}_3 & \mathbf{0}_3 & \mathbf{0}_3 & \mathbf{0}_3 & \mathbf{0}_3\\
			\mathbf{0}_3 & \mathbf{0}_3 & \mathbf{0}_3 & \mathbf{0}_3 & \mathbf{0}_3 & \mathbf{0}_3 & \mathbf{0}_3\\
			\mathbf{0}_3 & \mathbf{0}_3 & \mathbf{0}_3 & \mathbf{0}_3 & \mathbf{0}_3 & \mathbf{0}_3 & \mathbf{0}_3\\
		\end{bmatrix} \\
		\mathbf{G} &= \begin{bmatrix}
			\mathbf{0}_3 & \mathbf{0}_3 & \mathbf{0}_3 & \mathbf{0}_3 & \mathbf{0}_3 & \mathbf{0}_3 & \mathbf{0}_3 & \mathbf{0}_3 & \mathbf{0}_3\\
			-\mathbf{K} & \mathbf{0}_3 & \mathbf{I}_3 & -\hat{\mathbf{C}} & -[\hat{\mathbf{p}}_\times] & \mathbf{0}_3 & \mathbf{0}_3 & \mathbf{0}_3 & -[\hat{\mathbf{p}}_\times]\\
			\hat{\mathbf{C}}^T & -\mathbf{I}_3 & \mathbf{0}_3 & \mathbf{0}_3 & \mathbf{0}_3 & \mathbf{0}_3 & \mathbf{0}_3 & \mathbf{0}_3 & \mathbf{0}_3\\
			\mathbf{0}_3 & \mathbf{0}_3 & \mathbf{0}_3 & \mathbf{0}_3 & \mathbf{I}_3 & \mathbf{0}_3 & \mathbf{0}_3 & \mathbf{0}_3 & \mathbf{0}_3\\
			\mathbf{0}_3 & \mathbf{0}_3 & \mathbf{0}_3 & \mathbf{0}_3 & \mathbf{0}_3 & \mathbf{I}_3 & \mathbf{0}_3 & \mathbf{0}_3 & \mathbf{0}_3\\
			\mathbf{0}_3 & \mathbf{0}_3 & \mathbf{0}_3 & \mathbf{0}_3 & \mathbf{0}_3 & \mathbf{0}_3 & \mathbf{I}_3 & \mathbf{0}_3 & \mathbf{0}_3\\
			\mathbf{0}_3 & \mathbf{0}_3 & \mathbf{0}_3 & \mathbf{0}_3 & \mathbf{0}_3 & \mathbf{0}_3 & \mathbf{0}_3 & \mathbf{I}_3 & \mathbf{0}_3\\
		\end{bmatrix} \\
		\mathrm{with}&\ \hat{\mathbf{C}} \triangleq \mathbf{C}(\prescript{I_1}{}{\hat{\mathbf{q}}}_{I_2}), \ \ \ \ \ 	\mathbf{U} \triangleq  [{\prescript{I_1}{}{\hat{\boldsymbol{\omega}}}}_\times ]^2 + [^{I_1}{\dot{\hat{\boldsymbol{\omega}}}}_\times ]  \nonumber \\
		&\  \mathbf{K} \triangleq [\prescript{I_1}{}{ \hat{\boldsymbol{\omega}}}_\times ][{\prescript{I_1}{}{\hat{\mathbf{p}}}_{I_2}}_\times] + [(\prescript{I_1}{}{ \hat{\boldsymbol{\omega}}} \times \prescript{I_1}{}{\hat{\mathbf{p}}}_{I_2} + 2{\prescript{I_1}{}{\hat{\mathbf{v}}}_{I_2}}   )_\times] \nonumber
	\end{align}
	and
\begin{equation}
	\mathbf{n} = \begin{bmatrix} \mathbf{n}_{g_1}^T,\mathbf{n}_{g_2}^T,\mathbf{n}_{a_1}^T,\mathbf{n}_{a_2}^T,\mathbf{n}_{\omega g_1}^T,\mathbf{n}_{\omega g_2}^T, \mathbf{n}_{\omega a_1}^T,\mathbf{n}_{\omega a_2}^T, \dot{\mathbf{n}}_{g_1}^T \end{bmatrix}^T
\end{equation}

	Since \(\mathbf{n}_{g_1}\) is a stochastic process of white gaussian noise, the modelling of \(\dot{\mathbf{n}}_{g_1}\) is complicated. However, the noise term has no effect on the observability analysis. Hence, this term has no impact on the analysis and results of Sect.~\ref{unobservable-directions-under-special-motions}.
	
	\subsection*{Appendix C: Derivation of Discrete-Time Error-State Transition Matrix \label{appendix-c}}
	
	This section describes the derivation process of the \(\mathbf{\Phi}\) matrix.
	We have knowledge of the derivative of \(\mathbf{\Phi}\):
	
	\begin{equation}
		\dot{\mathbf{\Phi}}(t,t_0) = \mathbf{F}\mathbf{\Phi}(t,t_0)
	\end{equation}

	The structure of \(\mathbf{\Phi}(t,t_0)\) is:
	
	\begin{equation}
		\mathbf{\Phi}(t,t_0) = \begin{bmatrix}
			\mathbf{\Phi}_{11} & \mathbf{\Phi}_{12} & \mathbf{\Phi}_{13} & \mathbf{\Phi}_{14} & \mathbf{\Phi}_{15} & \mathbf{\Phi}_{16} &  \mathbf{\Phi}_{17}\\
			\mathbf{\Phi}_{21} & \mathbf{\Phi}_{22} & \mathbf{\Phi}_{23} & \mathbf{\Phi}_{24} & \mathbf{\Phi}_{25} & \mathbf{\Phi}_{26} &  \mathbf{\Phi}_{27} \\
			\mathbf{0}_3 & \mathbf{0}_3 & \mathbf{\Phi}_{33} & \mathbf{\Phi}_{34} & \mathbf{\Phi}_{35} & \mathbf{0}_3 & \mathbf{0}_3 \\
			\mathbf{0}_3 & \mathbf{0}_3 & \mathbf{0}_3 & \mathbf{I}_3 & \mathbf{0}_3 & \mathbf{0}_3 & \mathbf{0}_3  \\
			\mathbf{0}_3 & \mathbf{0}_3 & \mathbf{0}_3 & \mathbf{0}_3 & \mathbf{I}_3 & \mathbf{0}_3 & \mathbf{0}_3 \\
			\mathbf{0}_3 & \mathbf{0}_3 & \mathbf{0}_3 & \mathbf{0}_3 & \mathbf{0}_3 & \mathbf{I}_3 & \mathbf{0}_3  \\
			\mathbf{0}_3 & \mathbf{0}_3 & \mathbf{0}_3 & \mathbf{0}_3 & \mathbf{0}_3 & \mathbf{0}_3 & \mathbf{I}_3 
		\end{bmatrix}
	\end{equation}

	Blocks \(\mathbf{\Phi}_{ij}(t,t_0)\) when \(i>3\):
	\begin{align}
		\dot{\mathbf{\Phi}}_{ij}(t,t_0) &= \mathbf{F}_{i,:}(t,t_0)\mathbf{\Phi}_{:,j}(t,t_0) = \mathbf{0}_3 \\
		\mathbf{\Phi}_{ij}(t,t_0) &= \mathbf{\Phi}_{ij}(t_0,t_0) = \mathbf{I}_3 \ or \ \mathbf{0}_3
	\end{align}
	if \(i=j\):
	\begin{align}
		\mathbf{\Phi}_{ij}(t,t_0) &= \mathbf{\Phi}_{ij}(t_0,t_0) = \mathbf{I}_3
	\end{align}
	and if \(\ i \neq j\):
		\begin{align}
		\mathbf{\Phi}_{ij}(t,t_0) &= \mathbf{\Phi}_{ij}(t_0,t_0) = \mathbf{0}_3
	\end{align}
	
	Blocks \(\mathbf{\Phi}_{31}\), \(\mathbf{\Phi}_{32}\), \(\mathbf{\Phi}_{36}\) and \(\mathbf{\Phi}_{37}\):
	\begin{align}
		\dot{\mathbf{\Phi}}_{3j}(t,t_0) &= \mathbf{F}_{3,:}(t,t_0) \mathbf{\Phi}_{:,j}(t,t_0) = -[\hat{\boldsymbol{\omega}}_{2\times}]\mathbf{\Phi}_{3j}(t,t_0) \\
		\mathbf{\Phi}_{3j}(t_0,t_0) &= \mathbf{0}_3 
	\end{align}
Thus:
	\begin{align}
		\mathbf{\Phi}_{3j}(t,t_0) &= \mathbf{0}_3,~j = 1, 2, 6, 7
	\end{align}
	
	Block \(\mathbf{\Phi}_{33}\):
	\begin{align}
		\dot{\mathbf{\Phi}}_{33}(t,t_0)  &=\mathbf{F}_{3,:}(t,t_0) \mathbf{\Phi}_{:,3}(t,t_0)=  \begin{bmatrix} \mathbf{0}_3 &  \mathbf{0}_3 & -[\prescript{{I^{t}_2}}{}{\hat{\boldsymbol{\omega}}}_\times] & \mathbf{C}(\prescript{{I^{t}_2}}{}{\hat{\mathbf{q}}}_{I^{t}_1}) & - \mathbf{I}_3 & \mathbf{0}_3 & \mathbf{0}_3 \end{bmatrix} \begin{bmatrix} \mathbf{\Phi}_{13} \\ \mathbf{\Phi}_{23} \\ \mathbf{\Phi}_{33} \\ \mathbf{0}_3 \\ \mathbf{0}_3 \\ \mathbf{0}_3 \\ \mathbf{0}_3 \end{bmatrix} \nonumber\\
		&= -[\prescript{{I^{t}_2}}{}{\hat{\boldsymbol{\omega}}}_\times] \mathbf{\Phi}_{33}(t,t_0)
	\end{align}
Thus:
	\begin{align}
		\mathbf{\Phi}_{33}(t,t_0) &= \mathbf{e}^{\int_{t_0}^{t} -[\prescript{{I^s_2}}{}{\hat{\boldsymbol{\omega}}}_\times] d s}\mathbf{\Phi}_{33}(t_0,t_0) =  \mathbf{C}(\prescript{{I_2^{t}}}{}{\hat{\mathbf{q}}}_{I_2^{t_0}}) 
	\end{align}
	
	Block \(\mathbf{\Phi}_{34}\):
	\begin{align}
		\dot{\mathbf{\Phi}}_{34}(t,t_0)  &=\mathbf{F}_{3,:}(t,t_0) \mathbf{\Phi}_{:,4}(t,t_0)=  \begin{bmatrix} \mathbf{0}_3 &  \mathbf{0}_3 & -[\prescript{{I_2^t}}{}{\hat{\boldsymbol{\omega}}}_\times] & \mathbf{C}(\prescript{{I^t_2}}{}{\hat{\mathbf{q}}}_{I^t_1}) & - \mathbf{I}_3 & \mathbf{0}_3 & \mathbf{0}_3 \end{bmatrix} \begin{bmatrix} \mathbf{\Phi}_{14} \\ \mathbf{\Phi}_{24} \\ \mathbf{\Phi}_{34} \\ \mathbf{I}_3 \\ \mathbf{0}_3 \\ \mathbf{0}_3 \\ \mathbf{0}_3 \end{bmatrix} \nonumber\\
		& = -[\prescript{{I_2^t}}{}{\hat{\boldsymbol{\omega}}}_\times]\mathbf{\Phi}_{34}(t,t_0) + \mathbf{C}(\prescript{{I^t_2}}{}{\hat{\mathbf{q}}}_{I^t_1}) 
	\end{align}
Thus:
	\begin{align}
		\mathbf{\Phi}_{34}(t,t_0) &= \mathbf{e}^{\int_{t_0}^{t} -[\prescript{{I^s_2}}{}{\hat{\boldsymbol{\omega}}}_\times] d s} \int_{t_0}^{t} \mathbf{e}^{\int_{t_0}^s [\prescript{{I^{\tau}_2}}{}{\hat{\boldsymbol{\omega}}}_\times] d \tau} \mathbf{C}(\prescript{{I_2^{s}}}{}{\hat{\mathbf{q}}}_{I_1^{s}}) ds \nonumber \\
		&= \mathbf{C}''(t)\int_{t_0}^{t} \mathbf{C}(\prescript{ {I_2^{t_0}}}{}{\hat{\mathbf{q}}}_{I_1^{s}}) ds
	\end{align}
where
	\(\mathbf{C}''(t) = \mathbf{C}(\prescript{{I_2^{t}}}{}{\hat{\mathbf{q}}}_{I_2^{t_0}})\).
	
	Block \(\mathbf{\Phi}_{35}\):
	\begin{align}
		\dot{\mathbf{\Phi}}_{35}(t,t_0)  &=\mathbf{F}_{3,:}(t,t_0) \mathbf{\Phi}_{:,5}(t,t_0)=  \begin{bmatrix} \mathbf{0}_3 &  \mathbf{0}_3 & -[\prescript{{I_2^t}}{}{\hat{\boldsymbol{\omega}}}_\times] & \mathbf{C}(\prescript{{I^t_2}}{}{\hat{\mathbf{q}}}_{I^t_1}) & - \mathbf{I}_3 & \mathbf{0}_3 & \mathbf{0}_3 \end{bmatrix} \begin{bmatrix} \mathbf{\Phi}_{15} \\ \mathbf{\Phi}_{25} \\ \mathbf{\Phi}_{35} \\ \mathbf{0}_3 \\ \mathbf{I}_3 \\ \mathbf{0}_3 \\ \mathbf{0}_3 \end{bmatrix} \nonumber \\
		& =  -[\prescript{{I_2^t}}{}{\hat{\boldsymbol{\omega}}}_\times]\mathbf{\Phi}_{35}(t,t_0) - \mathbf{I}_3 
	\end{align}
	Thus:
	\begin{align}
		\mathbf{\Phi}_{35}(t, t_0 ) & = \mathbf{C}''(t) \int_{t_0}^{t} -\mathbf{e}^{\int_{t_0}^s [\prescript{{I^{\tau}_2}}{}{\hat{\boldsymbol{\omega}}}_\times] d \tau} ds  \nonumber \\
		&= -\mathbf{C}''(t)\int_{t_0}^{t} \mathbf{C}( \prescript{{I_2^{t_0}}}{}{\hat{\mathbf{q}}}_{I_2^{s}}) ds
	\end{align}

	Blocks \(\mathbf{\Phi}_{11}\) and \(\mathbf{\Phi}_{21}\):
	\begin{align}
		\dot{\mathbf{\Phi}}_{21}(t,t_0) =\mathbf{F}_{2,:} (t,t_0)\mathbf{\Phi}_{:,1}(t,t_0) &=  \begin{bmatrix} -\mathbf{U}(t) 
			& -2[\prescript{{I^t_1}}{}{\hat{\boldsymbol{\omega}}}_\times ] 
			& -\mathbf{C}(\prescript{{I^t_1}}{}{\hat{\mathbf{q}}}_{I^t_2})[\prescript{{I^t_2}}{}{\hat{\mathbf{a}}}_\times] 
			& -\mathbf{K}(t)
			& \mathbf{0}_3 
			& \mathbf{I}_3 
			& -\mathbf{C}(\prescript{{I^t_1}}{}{\hat{\mathbf{q}}}_{I^t_2}) 
			\end{bmatrix}
		\begin{bmatrix} \mathbf{\Phi}_{11} \\ \mathbf{\Phi}_{21} \\ \mathbf{0}_3 \\ \mathbf{0}_3 \\ \mathbf{0}_3 \\ \mathbf{0}_3 \\ \mathbf{0}_3 \end{bmatrix} \nonumber\\
		&= -\mathbf{U}(t)\mathbf{\Phi}_{11}(t,t_0)-2[\prescript{{I^t_1}}{}{\hat{\boldsymbol{\omega}}}_\times ] \mathbf{\Phi}_{21}(t,t_0)\\
		\dot{\mathbf{\Phi}}_{11}(t,t_0) =\mathbf{F}_{1,:} (t,t_0)\mathbf{\Phi}_{:,1}(t,t_0) &= \mathbf{\Phi}_{21}(t,t_0) 
	\end{align}
	Construct a second-order differential equation:
	\begin{align}
		\ddot{\mathbf{\Phi}}_{11}(t,t_0) +\mathbf{U}(t)\mathbf{\Phi}_{11}(t,t_0)+2[\prescript{{I^t_1}}{}{\hat{\boldsymbol{\omega}}}_\times ]\dot{\mathbf{\Phi}}_{11}(t,t_0) = \mathbf{0}_3
	\end{align}
	Initial values:
	\begin{align}
		\mathbf{\Phi}_{11}(t_0,t_0) &= \mathbf{I}_3 \\
		\mathbf{\Phi}_{21}(t_0,t_0) &= \mathbf{0}_3 
	\end{align}
	Thus:
	\begin{align}
		\mathbf{\Phi}_{11}(t,t_0) &= \mathbf{e}^{\int_{t_0}^{t} -[\prescript{{I^s_1}}{}{\hat{\boldsymbol{\omega}}}_\times ] d s}(\mathbf{I}+[\prescript{{I^{t_0}_1}}{}{\hat{\boldsymbol{\omega}}}_\times](t-{t_0})) =  \mathbf{C}'(t) (\mathbf{I}+[\prescript{{I^{t_0}_1}}{}{\hat{\boldsymbol{\omega}}}_\times](t-{t_0})) \\
		\mathbf{\Phi}_{21}(t,t_0) &= \dot{\mathbf{\Phi}}_{11}(t,t_0) = -[\prescript{{I^t_1}}{}{\hat{\boldsymbol{\omega}}}_\times] \mathbf{C}'(t) (\mathbf{I}+[\prescript{{I^{t_0}_1}}{}{\hat{\boldsymbol{\omega}}}_\times](t-{t_0}))  + \mathbf{C}'(t) [\prescript{{I^{t_0}_1}}{}{\hat{\boldsymbol{\omega}}}_\times] 
	\end{align}	
	where \(\mathbf{C}'(t) = \mathbf{C}(\prescript{{I_1^{t}}}{}{\hat{\mathbf{q}}}_{I_1^{t_0}})\).
	
	Blocks \(\mathbf{\Phi}_{12}\) and \(\mathbf{\Phi}_{22}\):
	\begin{align}
		\dot{\mathbf{\Phi}}_{22}(t,t_0) =\mathbf{F}_{2,:}(t,t_0) \mathbf{\Phi}_{:,2}(t,t_0) &=  \begin{bmatrix} -\mathbf{U}(t) 
			& -2[\prescript{{I^t_1}}{}{\hat{\boldsymbol{\omega}}}_\times ] 
			& -\mathbf{C}(\prescript{{I^t_1}}{}{\hat{\mathbf{q}}}_{I^t_2})[\prescript{{I^t_2}}{}{\hat{\mathbf{a}}}_\times] 
			& -\mathbf{K}(t)
			& \mathbf{0}_3 
			& \mathbf{I}_3 
			& -\mathbf{C}(\prescript{{I^t_1}}{}{\hat{\mathbf{q}}}_{I^t_2}) 
		\end{bmatrix}
		\begin{bmatrix} \mathbf{\Phi}_{12} \\ \mathbf{\Phi}_{22} \\ \mathbf{0}_3 \\ \mathbf{0}_3 \\ \mathbf{0}_3 \\ \mathbf{0}_3 \\ \mathbf{0}_3 \end{bmatrix} \nonumber\\
		&= -\mathbf{U}(t) \mathbf{\Phi}_{12}(t,t_0)-2[\prescript{{I^t_1}}{}{\hat{\boldsymbol{\omega}}}_\times ] \mathbf{\Phi}_{22}(t,t_0)\\
		\dot{\mathbf{\Phi}}_{12}(t,t_0) =\mathbf{F}_{1,:} (t,t_0)\mathbf{\Phi}_{:,2}(t,t_0) &= \mathbf{\Phi}_{22}(t,t_0)
	\end{align}
	Construct a second-order differential equation:
	\begin{align}
		\ddot{\mathbf{\Phi}}_{12}(t,t_0) +\mathbf{U}(t)\mathbf{\Phi}_{12}(t,t_0)+2[\prescript{{I^t_1}}{}{\hat{\boldsymbol{\omega}}}_\times ] \dot{\mathbf{\Phi}}_{12}(t,t_0) = \mathbf{0}_3
	\end{align}
	Initial values:
	\begin{align}
		\mathbf{\Phi}_{12}(t_0) &= \mathbf{0}_3 \\
		\mathbf{\Phi}_{22}(t_0) &= \mathbf{I}_3 
	\end{align}
	Thus:
	\begin{align}
		\mathbf{\Phi}_{12}&=  \mathbf{C}'(t) (t - t_0) \\
		\mathbf{\Phi}_{22}&=  -[\prescript{{I^t_1}}{}{\hat{\boldsymbol{\omega}}}_\times] \mathbf{C}'(t) (t - t_0) +\mathbf{C}'(t) 
	\end{align}
	
	Blocks \(\mathbf{\Phi}_{13}\) and \(\mathbf{\Phi}_{23}\):
	\begin{align}
		\dot{\mathbf{\Phi}}_{23}(t,t_0) =\mathbf{F}_{:,2}(t,t_0) \mathbf{\Phi}_{3,:}(t,t_0) &=  \begin{bmatrix} -\mathbf{U}(t) 
			& -2[\prescript{{I^t_1}}{}{\hat{\boldsymbol{\omega}}}_\times ] 
			& -\mathbf{C}(\prescript{{I^t_1}}{}{\hat{\mathbf{q}}}_{I^t_2})[\prescript{{I^t_2}}{}{\hat{\mathbf{a}}}_\times] 
			& -\mathbf{K}(t)
			& \mathbf{0}_3 
			& \mathbf{I}_3 
			& -\mathbf{C}(\prescript{{I^t_1}}{}{\hat{\mathbf{q}}}_{I^t_2}) 
		\end{bmatrix}
		\begin{bmatrix} \mathbf{\Phi}_{13} \\ \mathbf{\Phi}_{23} \\ \mathbf{\Phi}_{33} \\ \mathbf{0}_3 \\ \mathbf{0}_3 \\ \mathbf{0}_3 \\ \mathbf{0}_3 \end{bmatrix} \nonumber\\
		&= -\mathbf{U}(t) -2[\prescript{{I^t_1}}{}{\hat{\boldsymbol{\omega}}}_\times ] \mathbf{\Phi}_{23}(t,t_0) -\mathbf{C}(\prescript{{I^t_1}}{}{\hat{\mathbf{q}}}_{I^t_2})[\prescript{{I^t_2}}{}{\hat{\mathbf{a}}}_\times] \mathbf{\Phi}_{33}(t,t_0) \\
		\dot{\mathbf{\Phi}}_{13}(t,t_0) =\mathbf{F}_{:,1}(t,t_0) \mathbf{\Phi}_{3,:}(t,t_0) &={\mathbf{\Phi}}_{23} (t,t_0)
	\end{align}
	Construct a second-order differential equation:
	\begin{align}
		\ddot{\mathbf{\Phi}}_{13}(t,t_0) 
		+\mathbf{U}(t)\mathbf{\Phi}_{13}(t,t_0)+2[\prescript{{I^t_1}}{}{\hat{\boldsymbol{\omega}}}_\times ]\dot{\mathbf{\Phi}}_{13}(t,t_0) 
		+\mathbf{C}(\prescript{{I^t_1}}{}{\hat{\mathbf{q}}}_{I^t_2})[\prescript{{I^t_2}}{}{\hat{\mathbf{a}}}_\times] \mathbf{C}(\prescript{{I_2^{t}}}{}{\hat{\mathbf{q}}}_{I_2^{t_0}}) = \mathbf{0}_3
	\end{align}
	Initial values:
	\begin{align}
		\mathbf{\Phi}_{13}(t_0,t_0) &= \mathbf{0}_3 \\
		\mathbf{\Phi}_{23}(t_0,t_0) &= \mathbf{0}_3 
	\end{align}
	Thus:
	\begin{align}
		\mathbf{\Phi}_{13}(t,t_0)
		=& -\mathbf{C}'(t) \int_{t_0}^{t} \int_{t_0}^{s} \mathbf{C}(\prescript{{I_1^{t_0}}}{}{\hat{\mathbf{q}}_{I_1^{\tau}}})[(\mathbf{C}(\prescript{{I_1^{\tau}}}{}{\hat{\mathbf{q}}}_{I_2^{\tau}})\prescript{{I^\tau_2}}{}{\hat{\mathbf{a}}})_\times] \mathbf{C}(\prescript{{I_1^{\tau}}}{}{\hat{\mathbf{q}}}_{I_2^{t_0}})  d\tau ds \\
		\mathbf{\Phi}_{23} (t,t_0)
		=& [\prescript{{I^t_1}}{}{\hat{\boldsymbol{\omega}}}_\times] \mathbf{C}'(t) \int_{t_0}^{t} \int_{t_0}^{s} \mathbf{C}(\prescript{{I_1^{t_0}}}{}{\hat{\mathbf{q}}_{I_1^{\tau}}})[(\mathbf{C}(\prescript{{I_1^{\tau}}}{}{\hat{\mathbf{q}}}_{I_2^{\tau}})\prescript{{I^\tau_2}}{}{\hat{\mathbf{a}}})_\times] \mathbf{C}(\prescript{{I_1^{\tau}}}{}{\hat{\mathbf{q}}}_{I_2^{t_0}})  d\tau ds  \nonumber \\
		&-\mathbf{C}'(t)\int_{t_0}^{t}\mathbf{C}(\prescript{{I_1^{t_0}}}{}{\hat{\mathbf{q}}_{I_1^{\tau}}})[(\mathbf{C}(\prescript{{I_1^{\tau}}}{}{\hat{\mathbf{q}}}_{I_2^{\tau}})\prescript{{I^\tau_2}}{}{\hat{\mathbf{a}}})_\times] \mathbf{C}(\prescript{{I_1^{\tau}}}{}{\hat{\mathbf{q}}}_{I_2^{t_0}})  d\tau 
	\end{align}
	
	Blocks \(\mathbf{\Phi}_{14}\) and \(\mathbf{\Phi}_{24}\):
	\begin{align}
		\dot{\mathbf{\Phi}}_{24}(t,t_0) =\mathbf{F}_{2,:} (t,t_0)\Phi_{:,4}(t,t_0) &=  \begin{bmatrix} -\mathbf{U}(t) 
			& -2[\prescript{{I^t_1}}{}{\hat{\boldsymbol{\omega}}}_\times ] 
			& -\mathbf{C}(\prescript{{I^t_1}}{}{\hat{\mathbf{q}}}_{I^t_2})[\prescript{{I^t_2}}{}{\hat{\mathbf{a}}}_\times] 
			& -\mathbf{K}(t)
			& \mathbf{0}_3 
			& \mathbf{I}_3 
			& -\mathbf{C}(\prescript{{I^t_1}}{}{\hat{\mathbf{q}}}_{I^t_2}) 
		\end{bmatrix}
		\begin{bmatrix} \mathbf{\Phi}_{14} \\ \mathbf{\Phi}_{24} \\ \mathbf{\Phi}_{34} \\ \mathbf{I}_3 \\ \mathbf{0}_3 \\ \mathbf{0}_3 \\ \mathbf{0}_3 \end{bmatrix} \nonumber\\
		&= -\mathbf{U}(t) \mathbf{\Phi}_{14}(t,t_0)-2[\prescript{{I^t_1}}{}{\hat{\boldsymbol{\omega}}}_\times ] \mathbf{\Phi}_{24}(t,t_0) -\mathbf{C}(\prescript{{I^t_1}}{}{\hat{\mathbf{q}}}_{I^t_2}) \mathbf{\Phi}_{34}(t,t_0) - \mathbf{K}(t) \\
		\dot{\mathbf{\Phi}}_{14}(t,t_0) =\mathbf{F}_{1,:}(t,t_0) \mathbf{\Phi}_{:,4}(t,t_0) &={\mathbf{\Phi}}_{24}(t,t_0)  
	\end{align}
	
	Construct a second-order differential equation:
	\begin{align}
		&\ddot{\mathbf{\Phi}}_{14}(t,t_0) + \mathbf{U}(t)\mathbf{\Phi}_{14}(t,t_0)+2[\prescript{{I^t_1}}{}{\hat{\boldsymbol{\omega}}}_\times ] \dot{\mathbf{\Phi}}_{14}(t,t_0)+\mathbf{C}(\prescript{{I^t_1}}{}{\hat{\mathbf{q}}}_{I^t_2})[\prescript{{I^t_2}}{}{\hat{\mathbf{a}}}_\times] \mathbf{C}''(t)\int_{t_0}^{t} \mathbf{C}(\prescript{ {I_2^{t_0}}}{}{\hat{\mathbf{q}}}_{I_1^{s}})ds  +\mathbf{K}(t)= \mathbf{0}_3
	\end{align}
	Initial values:
	\begin{align}
		\mathbf{\Phi}_{14}\left(t_0,t_0\right) &= \mathbf{0}_3 \\
		\mathbf{\Phi}_{24}\left(t_0,t_0\right) &= \mathbf{0}_3 
	\end{align}
	Thus:
	\begin{align}
		\mathbf{\Phi}_{14} (t,t_0)
		=& -\mathbf{C}'(t) \int_{t_0}^{t} \int_{t_0}^{s} \mathbf{C}(\prescript{{I_1^{t_0}}}{}{\hat{\mathbf{q}}_{I_1^{\tau}}}) [(\mathbf{C}(\prescript{{I_1^{\tau}}}{}{\hat{\mathbf{q}}}_{I_2^{\tau}})\prescript{{I^\tau_2}}{}{\hat{\mathbf{a}}})_\times] 
		\int_{t_0}^{\tau} \mathbf{C}(\prescript{{I_1^{\tau}}}{}{\hat{\mathbf{q}}}_{I_1^{\mu}}) d\mu d\tau ds \nonumber \\
		& - \mathbf{C}'(t) \int_{t_0}^{t} \int_{t_0}^{s} \mathbf{C}(\prescript{{I_1^{t_0}}}{}{\hat{\mathbf{q}}}_{I_1^{\tau}}) \mathbf{K}(\tau) d\tau ds \\
		\mathbf{\Phi}_{24} (t,t_0)
		=&~[\prescript{{I^t_1}}{}{\hat{\boldsymbol{\omega}}}_\times] 
		\mathbf{C}'(t) \int_{t_0}^{t} \int_{t_0}^{s} \mathbf{C}(\prescript{{I_1^{t_0}}}{}{\hat{\mathbf{q}}_{I_1^{\tau}}}) [(\mathbf{C}(\prescript{{I_1^{\tau}}}{}{\hat{\mathbf{q}}}_{I_2^{\tau}})\prescript{{I^\tau_2}}{}{\hat{\mathbf{a}}})_\times] 
		\int_{t_0}^{\tau} \mathbf{C}(\prescript{{I_1^{\tau}}}{}{\hat{\mathbf{q}}}_{I_1^{\mu}}) d\mu d\tau ds \nonumber \\
		 &+ [\prescript{{I^t_1}}{}{\hat{\boldsymbol{\omega}}}_\times]\mathbf{C}'(t) \int_{t_0}^{t} \int_{t_0}^{s} \mathbf{C}(\prescript{{I_1^{t_0}}}{}{\hat{\mathbf{q}}}_{I_1^{\tau}}) \mathbf{K}(\tau) d\tau ds   \nonumber\\
		&-\mathbf{C}'(t) \int_{t_0}^{t} \mathbf{C}(\prescript{{I_1^{t_0}}}{}{\hat{\mathbf{q}}_{I_1^{\tau}}}) [(\mathbf{C}(\prescript{{I_1^{\tau}}}{}{\hat{\mathbf{q}}}_{I_2^{\tau}})\prescript{{I^\tau_2}}{}{\hat{\mathbf{a}}})_\times] 
		\int_{t_0}^{\tau} \mathbf{C}(\prescript{{I_1^{\tau}}}{}{\hat{\mathbf{q}}}_{I_1^{\mu}}) d\mu d\tau - \mathbf{C}'(t) \int_{t_0}^{t} \mathbf{C}(\prescript{{I_1^{t_0}}}{}{\hat{\mathbf{q}}}_{I_1^{\tau}}) \mathbf{K}(\tau) d\tau 
	\end{align}
	
	Blocks \(\mathbf{\Phi}_{15}\) and \(\mathbf{\Phi}_{25}\):
	\begin{align}
		\dot{\mathbf{\Phi}}_{25}(t,t_0) =\mathbf{F}_{2,:}(t,t_0) \mathbf{\Phi}_{:,5}(t,t_0) &=  \begin{bmatrix} -\mathbf{U}(t) 
			& -2[\prescript{{I^t_1}}{}{\hat{\boldsymbol{\omega}}}_\times ] 
			& -\mathbf{C}(\prescript{{I^t_1}}{}{\hat{\mathbf{q}}}_{I^t_2})[\prescript{{I^t_2}}{}{\hat{\mathbf{a}}}_\times] 
			& -\mathbf{K}(t)
			& \mathbf{0}_3 
			& \mathbf{I}_3 
			& -\mathbf{C}(\prescript{{I^t_1}}{}{\hat{\mathbf{q}}}_{I^t_2}) 
		\end{bmatrix}
		\begin{bmatrix} \mathbf{\Phi}_{15} \\ \mathbf{\Phi}_{25} \\ \mathbf{\Phi}_{35} \\ \mathbf{0}_3 \\ \mathbf{I}_3 \\ \mathbf{0}_3 \\ \mathbf{0}_3 \end{bmatrix} \nonumber\\
		&= -\mathbf{U}(t)\mathbf{\Phi}_{15}(t,t_0)-2[\prescript{{I^t_1}}{}{\hat{\boldsymbol{\omega}}}_\times ] \mathbf{\Phi}_{25}(t,t_0) -\mathbf{C}(\prescript{{I^t_1}}{}{\hat{\mathbf{q}}}_{I^t_2})[\prescript{{I^t_2}}{}{\hat{\mathbf{a}}}_\times]  \mathbf{\Phi}_{35}(t,t_0) \\
		\dot{\mathbf{\Phi}}_{15}(t,t_0) =\mathbf{F}_{1,:}(t,t_0) \mathbf{\Phi}_{:,5}(t,t_0) &={\mathbf{\Phi}}_{25} (t,t_0)
	\end{align}
	Construct a second-order differential equation:
	\begin{align}
		\ddot{\mathbf{\Phi}}_{15}(t,t_0) + \mathbf{U}(t) \mathbf{\Phi}_{15}(t,t_0)+2[\prescript{{I^t_1}}{}{\hat{\boldsymbol{\omega}}}_\times ] \dot{\mathbf{\Phi}}_{15}(t,t_0) 
		-\mathbf{C}(\prescript{{I^t_1}}{}{\hat{\mathbf{q}}}_{I^t_2})[\prescript{{I^t_2}}{}{\hat{\mathbf{a}}}_\times]\mathbf{C}''(t)\int_{t_0}^{t} \mathbf{C}( \prescript{{I_2^{t_0}}}{}{\hat{\mathbf{q}}}_{I_2^{s}}) ds = \mathbf{0}_3
	\end{align}
	Initial values:
	\begin{align}
		\mathbf{\Phi}_{15}(t_0,t_0) &= \mathbf{0}_3 \\
		\mathbf{\Phi}_{25}(t_0,t_0) &= \mathbf{0}_3
	\end{align}
	Thus:
	\begin{align}
		\mathbf{\Phi}_{15} (t,t_0)
		&= \mathbf{C}'(t)\int_{t_0}^{t} \int_{t_0}^{s} \mathbf{C}(\prescript{{I_1^{t_0}}}{}{\hat{\mathbf{q}}}_{I_1^{\tau}}) [(\mathbf{C}(\prescript{{I_1^{\tau}}}{}{\hat{\mathbf{q}}}_{I_2^{\tau}})\prescript{{I^\tau_2}}{}{\hat{\mathbf{a}}})_\times] \int_{t_0}^{\tau} \mathbf{C}(\prescript{{I_1^{\tau}}}{}{\hat{\mathbf{q}}}_{I_2^{\mu}}) d\mu  d\tau ds  \\
		\mathbf{\Phi}_{25} (t,t_0)
		&= -[\prescript{{I^t_1}}{}{\hat{\boldsymbol{\omega}}}_\times] \mathbf{C}'(t)\int_{t_0}^{t} \int_{t_0}^{s} \mathbf{C}(\prescript{{I_1^{t_0}}}{}{\hat{\mathbf{q}}}_{I_1^{\tau}}) [(\mathbf{C}(\prescript{{I_1^{\tau}}}{}{\hat{\mathbf{q}}}_{I_2^{\tau}})\prescript{{I^\tau_2}}{}{\hat{\mathbf{a}}})_\times] \int_{t_0}^{\tau} \mathbf{C}(\prescript{{I_1^{\tau}}}{}{\hat{\mathbf{q}}}_{I_2^{\mu}}) d\mu  d\tau ds \nonumber \\ 
		&+ \mathbf{C}'(t)\int_{t_0}^{t} \mathbf{C}(\prescript{{I_1^{t_0}}}{}{\hat{\mathbf{q}}}_{I_1^{\tau}}) [(\mathbf{C}(\prescript{{I_1^{\tau}}}{}{\hat{\mathbf{q}}}_{I_2^{\tau}})\prescript{{I^\tau_2}}{}{\hat{\mathbf{a}}})_\times] \int_{t_0}^{\tau} \mathbf{C}(\prescript{{I_1^{\tau}}}{}{\hat{\mathbf{q}}}_{I_2^{\mu}}) d\mu  d\tau 
	\end{align}
	
	Blocks \(\mathbf{\Phi}_{16}\) and \(\mathbf{\Phi}_{26}\):
	\begin{align}
		\dot{\mathbf{\Phi}}_{26}(t,t_0) =\mathbf{F}_{2,:}(t,t_0)\mathbf{\Phi}_{:,6}(t,t_0) &=  \begin{bmatrix} -\mathbf{U}(t) 
			& -2[\prescript{{I^t_1}}{}{\hat{\boldsymbol{\omega}}}_\times ] 
			& -\mathbf{C}(\prescript{{I^t_1}}{}{\hat{\mathbf{q}}}_{I^t_2})[\prescript{{I^t_2}}{}{\hat{\mathbf{a}}}_\times] 
			& -\mathbf{K}(t)
			& \mathbf{0}_3 
			& \mathbf{I}_3 
			& -\mathbf{C}(\prescript{{I^t_1}}{}{\hat{\mathbf{q}}}_{I^t_2}) 
		\end{bmatrix}
		\begin{bmatrix} \mathbf{\Phi}_{16} \\ \mathbf{\Phi}_{26} \\ \mathbf{0}_3 \\ \mathbf{0}_3 \\ \mathbf{0}_3 \\ \mathbf{I}_3 \\ \mathbf{0}_3 \end{bmatrix} \nonumber\\
		&= -\mathbf{U}(t)\mathbf{\Phi}_{16}(t,t_0)-2[\prescript{{I^t_1}}{}{\hat{\boldsymbol{\omega}}}_\times ]\mathbf{\Phi}_{26}(t,t_0) +\mathbf{I}_3\\
		\dot{\mathbf{\Phi}}_{16}(t,t_0) =\mathbf{F}_{1,:}(t,t_0) \mathbf{\Phi}_{:,6}(t,t_0) &={\mathbf{\Phi}}_{26}(t,t_0) 
	\end{align}
	Construct a second-order differential equation:
	\begin{align}
		\ddot{\mathbf{\Phi}}_{16}(t,t_0) + \mathbf{U}(t)\mathbf{\Phi}_{16}(t,t_0)+2[\prescript{{I^t_1}}{}{\hat{\boldsymbol{\omega}}}_\times ]\dot{\mathbf{\Phi}}_{16}(t,t_0) -\mathbf{I}_3 = \mathbf{0}_3
	\end{align}
	Initial values:
	\begin{align}
		\mathbf{\Phi}_{16}(t_0,t_0) &= \mathbf{0}_3 \\
		\mathbf{\Phi}_{26}(t_0,t_0) &= \mathbf{0}_3 
	\end{align}
	Thus:
	\begin{align}
		\mathbf{\Phi}_{16}(t,t_0)
		&=  \mathbf{C}'(t)\int_{t_0}^{t} \int_{t_0}^{s} \mathbf{C}(\prescript{{I_1^{t_0}}}{}{\hat{\mathbf{q}}}_{I_1^{\tau}}) d\tau ds \\
		\mathbf{\Phi}_{26}(t,t_0)
		&=  -[\prescript{{I^t_1}}{}{\hat{\boldsymbol{\omega}}}_\times]\mathbf{C}'(t)\int_{t_0}^{t} \int_{t_0}^{s} \mathbf{C}(\prescript{{I_1^{t_0}}}{}{\hat{\mathbf{q}}}_{I_1^{\tau}}) d\tau ds +  \mathbf{C}'(t)\int_{t_0}^{t} \mathbf{C}(\prescript{{I_1^{t_0}}}{}{\hat{\mathbf{q}}}_{I_1^{\tau}}) d\tau  
	\end{align}
	
	Blocks \(\mathbf{\Phi}_{17}\) and \(\mathbf{\Phi}_{27}\):
	\begin{align}
		\dot{\mathbf{\Phi}}_{27}(t,t_0) =\mathbf{F}_{2,:} \mathbf{\Phi}_{:,7} &=  \begin{bmatrix} -\mathbf{U}(t) 
			& -2[\prescript{{I^t_1}}{}{\hat{\boldsymbol{\omega}}}_\times ] 
			& -\mathbf{C}(\prescript{{I^t_1}}{}{\hat{\mathbf{q}}}_{I^t_2})[\prescript{{I^t_2}}{}{\hat{\mathbf{a}}}_\times] 
			& -\mathbf{K}(t)
			& \mathbf{0}_3 
			& \mathbf{I}_3 
			& -\mathbf{C}(\prescript{{I^t_1}}{}{\hat{\mathbf{q}}}_{I^t_2}) 
		\end{bmatrix}
		\begin{bmatrix} \mathbf{\Phi}_{17} \\ \mathbf{\Phi}_{27} \\ \mathbf{0}_3 \\ \mathbf{0}_3 \\ \mathbf{0}_3 \\ \mathbf{0}_3 \\ \mathbf{I}_3 \end{bmatrix} \nonumber\\
		&= -\mathbf{U}(t,t_0)\mathbf{\Phi}_{17}-2[\prescript{{I^t_1}}{}{\hat{\boldsymbol{\omega}}}_\times ]  \mathbf{\Phi}_{27} -\mathbf{C}(\prescript{{I^t_1}}{}{\hat{\mathbf{q}}}_{I^t_2}) \\
		\dot{\mathbf{\Phi}}_{17}(t,t_0) =\mathbf{F}_{1,:} \mathbf{\Phi}_{:,7} &={\mathbf{\Phi}}_{27}(t,t_0) 
	\end{align}
	Construct a second-order differential equation:
	\begin{align}
		\ddot{\mathbf{\Phi}}_{17}(t,t_0) +\mathbf{U}(t) \mathbf{\Phi}_{17}(t,t_0) +2[\prescript{{I^t_1}}{}{\hat{\boldsymbol{\omega}}}_\times ]\dot{\mathbf{\Phi}}_{17}(t,t_0)  +\mathbf{C}(\prescript{{I^t_1}}{}{\hat{\mathbf{q}}}_{I^t_2})  = \mathbf{0}_3
	\end{align}
	Initial values:
	\begin{align}
		\mathbf{\Phi}_{17}(t_0,t_0) &= \mathbf{0}_3 \\
		\mathbf{\Phi}_{27}(t_0,t_0) &= \mathbf{0}_3 
	\end{align}
	Thus:
	\begin{align}
		\mathbf{\Phi}_{17} (t,t_0)
		&=  -\mathbf{C}'(t)\int_{t_0}^{t} \int_{t_0}^{s} \mathbf{C}( \prescript{{I_1^{t_0}}}{}{\hat{\mathbf{q}}}_{I_2^{\tau}}) d\tau ds \\
		\mathbf{\Phi}_{27}(t,t_0)
		&=[\prescript{{I^t_1}}{}{\hat{\boldsymbol{\omega}}}_\times] \mathbf{C}'(t)\int_{t_0}^{t} \int_{t_0}^{s} \mathbf{C}( \prescript{{I_1^{t_0}}}{}{\hat{\mathbf{q}}}_{I_2^{\tau}}) d\tau ds - \mathbf{C}'(t)\int_{t_0}^{t}  \mathbf{C}( \prescript{{I_1^{t_0}}}{}{\hat{\mathbf{q}}}_{I_2^{\tau}}) d\tau	\end{align}

	\subsection*{Appendix D: Proof of Observability Under General Motion}\label{appendix-d}
	To simplify the expression, in this section, we have adopted the same subscript and superscript simplification method as in Appendix A.
	
	\vspace{0.3cm}
	\subsubsection*{1) Derivation of System Dynamic Model}~{}
	
	The system state in \eqref{nolinear-state} becomes:
	\begin{equation}
		\mathbf{x}' =\begin{bmatrix} \mathbf{p}^T & \ \mathbf{v'}^T & {\mathbf{q}} & \mathbf{b}_{g_1}^T& \mathbf{b}_{g_2}^T& \mathbf{b}_{a_1}^T& \mathbf{b}_{a_2}^T \end{bmatrix}^T 
	\end{equation}
	where \(\mathbf{v}' = \mathbf{v} + {\boldsymbol{\omega}_1} \times \mathbf{p} =  \mathbf{C}_1^T( \mathbf{v}_2 - \mathbf{v}_1 )\) \eqref{def-v}. 
	
	We derive the terms of \(\dot{\mathbf{p}}\) and \(\dot{\mathbf{v}}'\) in system equation, as follow:
	\begin{align}
		\dot{\mathbf{p}} =& \mathbf{v} = \mathbf{v}' - {\boldsymbol{\omega}_1} \times \mathbf{p} \\
		\dot{\mathbf{v}}' =& \frac{d (\mathbf{C}_1^T(\mathbf{v}_2 - \mathbf{v}_1))}{dt}  \nonumber \\
					=& \frac{d\mathbf{C}_1^T}{dt}(\mathbf{v}_2-\mathbf{v}_1) + \mathbf{C}^T(\dot{\mathbf{v}}_2 - \dot{\mathbf{v}}_1 ) \nonumber \\
					=& \frac{d\mathbf{C}_1^T}{dt}(\mathbf{v}_2-\mathbf{v}_1) + \mathbf{C}_1^T(\mathbf{C}_2\mathbf{a}_2 - \mathbf{C}_1 \mathbf{a}_1 )  \nonumber \\
					=& -\boldsymbol{\omega_{1}}  \times \mathbf{v}' + \mathbf{C} \mathbf{a}_2 - \mathbf{a}_1 
	\end{align}
	
	Thus, the system equation become (same as \eqref{nolinear-tran}):
	\begin{align}
	\dot{\mathbf{x}}' = \underbrace{\begin{bmatrix} \mathbf{v}' + \mathbf{b}_{g_1} \times \mathbf{p} \\ 
			-\mathbf{C} \mathbf{b}_{a_2} + \mathbf{b}_{a_1}+\mathbf{b}_{g_1}\times \mathbf{v}' \\ 
			-\frac{1}{2}(\left[\mathbf{q}\right]_{L} \begin{bmatrix}0 \\	\mathbf{b}_{g_2} \end{bmatrix}  - \left[\mathbf{q}\right]_R \begin{bmatrix}0 \\	\mathbf{b}_{g_1} \end{bmatrix} )  \\ 
			\mathbf{0}_{3 \times 1} \\\mathbf{0}_{3 \times 1}  \\\mathbf{0}_{3 \times 1} \\\mathbf{0}_{3 \times 1} \end{bmatrix}}_{\mathbf{f}_0}
	+ \underbrace{\begin{bmatrix} [\mathbf{p}_\times] \\ [\mathbf{v}'_\times] \\ -\frac{1}{2} \left[\mathbf{q}\right]_{R(:,2:4)} \\ \mathbf{0}_3\\ \mathbf{0}_3\\ \mathbf{0}_3\\ \mathbf{0}_3 \end{bmatrix}}_{\mathbf{f}_1} \boldsymbol{\omega}_{1m}
	+ \underbrace{\begin{bmatrix} \mathbf{0}_3 \\ \mathbf{0}_3\\\frac{1}{2}\left[\mathbf{q}\right]_{L(:,2:4)} \\\mathbf{0}_3\\\mathbf{0}_3\\\mathbf{0}_3 \\\mathbf{0}_3\end{bmatrix}}_{\mathbf{f_2}}  \boldsymbol{\omega}_{2m}
	+ \underbrace{\begin{bmatrix} \mathbf{0}_3 \\-\mathbf{I}_3 \\\mathbf{0}_3\\\mathbf{0}_3\\\mathbf{0}_3\\\mathbf{0}_3\\\mathbf{0}_3  \end{bmatrix}}_{\mathbf{f}_3} \mathbf{a}_{1m} 
	+ \underbrace{\begin{bmatrix} \mathbf{0}_3 \\ \mathbf{C}\\\mathbf{0}_3\\\mathbf{0}_3\\\mathbf{0}_3\\\mathbf{0}_3\\\mathbf{0}_3 \end{bmatrix}}_{\mathbf{f_4}} \mathbf{a}_{2m} \label{nolinearxdot}
	\end{align}
	where \(\boldsymbol{\omega}_{1m} = \boldsymbol{\omega}_1 + \mathbf{b}_{g_1}\), \(\boldsymbol{\omega}_{2m} = \boldsymbol{\omega}_2 + \mathbf{b}_{g_2}\),  \(\mathbf{a}_{1m} = \mathbf{a}_1 + \mathbf{b}_{a_1}\) and \(\mathbf{a}_{2m} = \mathbf{a}_2 + \mathbf{b}_{a_2}\).

	\subsubsection*{2) Proof of the Full Column Rank Property of the Nonlinear Observability Matrix}~{}
	
	Combining the Lie derivatives set from \(\eqref{XX}\) into matrix \(\boldsymbol{\Xi} \). 
	\begin{equation}
		\boldsymbol{\Xi} = \begin{bmatrix} 
			\boldsymbol{L}^0\boldsymbol{h}_1 \\ 
			\boldsymbol{L}_{f_0}^1\boldsymbol{h}_1 \\ 
			\boldsymbol{L}^2_{f_0f_{41}}\boldsymbol{h}_1\\  
			\boldsymbol{L}^2_{f_0f_{42}}\boldsymbol{h}_1 \\  
			\boldsymbol{L}^2_{f_0f_{43}}\boldsymbol{h}_1 \\
			\boldsymbol{L}^0\boldsymbol{h}_0 \\
			\boldsymbol{L}_{f_0f_0f_{31}}^3\boldsymbol{h}_1  \\
			\boldsymbol{L}_{f_0f_0f_{32}}^3\boldsymbol{h}_1 \\ 
			\boldsymbol{L}^3_{{f_0}f_{41}f_0}\boldsymbol{h}_1 \\  
			\boldsymbol{L}^3_{{f_0}f_{42}f_0}\boldsymbol{h}_1 \\  
			\boldsymbol{L}^3_{{f_0}f_{43}f_0}\boldsymbol{h}_1 \\ 
			\boldsymbol{L}^1_{f_0}\boldsymbol{h}_0  \\
			\boldsymbol{L}_{f_0f_0}^2\boldsymbol{h}_1 \\
			\boldsymbol{L}_{f_0f_0f_{21}}^3\boldsymbol{h}_1  \\
			\boldsymbol{L}_{f_0f_0f_{22}}^3\boldsymbol{h}_1
		\end{bmatrix} \label{XX1}
	\end{equation}

	Expand the Lie derivatives involved in \(\eqref{XX1}\), according to \eqref{nolinearxdot}:
	\begin{align}
		\boldsymbol{L}^0\boldsymbol{h}_1 &= \begin{bmatrix} \mathbf{I}_3 & \mathbf{0}_{3\times 19} \end{bmatrix} \label{n1}\\
		\boldsymbol{L}_{f_0}^1\boldsymbol{h}_1 &= \begin{bmatrix} \mathbf{X}_1 && \mathbf{I}_3 && \mathbf{0}_{{3\times4}} & -[\mathbf{p}_\times] &  \mathbf{0}_{3\times9} \end{bmatrix} \label{n21}  \\
		\boldsymbol{L}^2_{f_0f_{41}}\boldsymbol{h}_1 &= \begin{bmatrix} \mathbf{0}_3 && \mathbf{0}_3 && \mathbf{\Lambda}_1 &&& \mathbf{0}_3 & \ \  \mathbf{0}_{3\times9} \end{bmatrix}  \\
		\boldsymbol{L}^2_{f_0f_{42}}\boldsymbol{h}_1 &= \begin{bmatrix} \mathbf{0}_3 && \mathbf{0}_3 && \mathbf{\Lambda}_2 &&& \mathbf{0}_3 & \ \ \mathbf{0}_{3\times9} \end{bmatrix}  \\
		\boldsymbol{L}^2_{f_0f_{43}}\boldsymbol{h}_1 &= \begin{bmatrix} \mathbf{0}_3 && \mathbf{0}_3 && \mathbf{\Lambda}_3 &&& \mathbf{0}_3 & \ \ \mathbf{0}_{3\times9} \end{bmatrix}  \\
		\boldsymbol{L}^0\boldsymbol{h}_0 &= \begin{bmatrix} \mathbf{0}_{1\times3} & \mathbf{0}_{1\times3} & \mathbf{S} &&& \mathbf{0}_{1\times3} & \mathbf{0}_{1\times9} \end{bmatrix}  \\
		\boldsymbol{L}_{f_0f_0f_{31}}^3\boldsymbol{h}_1 &=\begin{bmatrix} \mathbf{0}_3 && \mathbf{0}_{3} && \mathbf{0}_{3\times4} &\  -2\mathbf{P}_1 &  \mathbf{0}_{3\times9} \end{bmatrix}  \\
		\boldsymbol{L}_{f_0f_0f_{32}}^3\boldsymbol{h}_1 &=\begin{bmatrix} \mathbf{0}_3 && \mathbf{0}_{3} && \mathbf{0}_{3\times4} &\  -2\mathbf{P}_2 & \mathbf{0}_{3\times9}  \end{bmatrix} \label{n22} \\
		\boldsymbol{L}^3_{{f_0}f_{41}f_0}\boldsymbol{h}_1  &= \begin{bmatrix} \mathbf{0}_3&& \mathbf{0}_{3}  && \mathbf{Z}_1  & \mathbf{Z}_2 & -\frac{1}{2}\mathbf{\Lambda}_1 \left[\mathbf{q}\right]_{L(:,2:4)}  &\mathbf{0}_{3} && \mathbf{0}_{3} \end{bmatrix} \label{n31} \\
		\boldsymbol{L}^3_{{f_0}f_{42}f_0}\boldsymbol{h}_1  &= \begin{bmatrix} \mathbf{0}_3&& \mathbf{0}_{3}  && \mathbf{Z}_3  & \mathbf{Z}_4 & -\frac{1}{2}\mathbf{\Lambda}_2 \left[\mathbf{q}\right]_{L(:,2:4)}  &\mathbf{0}_{3} && \mathbf{0}_{3} \end{bmatrix}  \\
		\boldsymbol{L}^3_{{f_0}f_{43}f_0}\boldsymbol{h}_1  &= \begin{bmatrix} \mathbf{0}_3&& \mathbf{0}_{3}  && \mathbf{Z}_5  & \mathbf{Z}_6 & -\frac{1}{2}\mathbf{\Lambda}_3 \left[\mathbf{q}\right]_{L(:,2:4)}  & \mathbf{0}_{3}  && \mathbf{0}_{3} \end{bmatrix}  \\
		\boldsymbol{L}_{f_0}^1 \boldsymbol{h}_0 &= \begin{bmatrix} \mathbf{0}_{1\times3}  & \mathbf{0}_{1\times3}  & \mathbf{Z}_7  & \mathbf{Z}_8 & -\frac{1}{2}\mathbf{S}\left[\mathbf{q}\right]_{L(:,2:4)}  & \mathbf{0}_{1\times3} & \mathbf{0}_{1\times3} \end{bmatrix}  \\
		\boldsymbol{L}_{f_0f_0}^2\boldsymbol{h}_1 &= \begin{bmatrix} \mathbf{Y}_1 &\ \  \mathbf{Z}_9 &&  \mathbf{Z}_{10} & \mathbf{Z}_{11} &\ \ \ \ \  \mathbf{0}_3\ \ \ \ \ \  && \mathbf{I}_3 && -\mathbf{C} \end{bmatrix}  \\
		\boldsymbol{L}_{f_0f_0f_{21}}^3\boldsymbol{h}_1 &=  \begin{bmatrix} \mathbf{0}_{3}  && \mathbf{0}_{3} && \mathbf{Z}_{12} & \mathbf{0}_3 &\ \ \ \ \ \ \  \mathbf{0}_{3}\ \ \ \ \ \   && \mathbf{0}_{3} & -\mathbf{C}\mathbf{P}_1  \end{bmatrix}  \\
		\boldsymbol{L}_{f_0f_0f_{22}}^3\boldsymbol{h}_1 &=  \begin{bmatrix} \mathbf{0}_{3}  && \mathbf{0}_{3} && \mathbf{Z}_{13} & \mathbf{0}_3 &\ \ \ \ \ \ \   \mathbf{0}_{3}\ \ \ \ \ \   &&  \mathbf{0}_{3} & -\mathbf{C}\mathbf{P}_2  \end{bmatrix} \label{n32}
 	\end{align}
	where:
	\begin{align}
		\mathbf{\Lambda}_1 &= 2\begin{bmatrix} q_w & q_x  & -q_y & -q_z \\  q_z & q_y & q_x & q_w \\ -q_y & q_z &  -q_w & q_x \end{bmatrix} \label{lambda_1}\\
		\mathbf{\Lambda}_2 &= 2\begin{bmatrix} -q_z &  q_y & q_x & -q_w \\  q_w & -q_x & q_y & -q_z \\ q_x & q_w &   q_z &  q_y \end{bmatrix} \label{lambda_2}\\
		\mathbf{\Lambda}_3 &= 2\begin{bmatrix}  q_y & q_z & q_w & q_x \\  -q_x & -q_w & q_z & q_y \\ q_w & -q_x  & -q_y   &  q_z \end{bmatrix}  \label{lambda_3} \\
		\mathbf{S} &= 2 \begin{bmatrix}q_w & q_x & q_y & q_z \end{bmatrix} \label{lambda_q} \\
		\mathbf{P}_1 &= \begin{bmatrix} 0 & 0 & 0 \\ 0 & 0 & 1 \\ 0 & -1&0 \end{bmatrix} \\
		\mathbf{P}_2 &= \begin{bmatrix} 0 & 0 & -1 \\ 0 & 0 & 0 \\ 1 & 0&0 \end{bmatrix}  
	\end{align}

	The matrix \(\mathbf{\Xi}\) can be represented in the following
	structure:
	\begin{equation}
		\boldsymbol{\Xi} = 
		\begin{bmatrix}
			\mathbf{I}_3 & \mathbf{0}_{3\times10} & \mathbf{0}_{3\times9} \\
			\mathbf{X}  & \mathbf{A}_{19\times10} & \mathbf{0}_{19\times9} \\
			\mathbf{Y}  & \mathbf{Z} & \mathbf{B}_{19\times9} 
		\end{bmatrix} 
	\end{equation}
	where the first row of \(\boldsymbol{\Xi}\) corresponds to \eqref{n1}, the second row corresponds  to \eqref{n21} to \eqref{n22} and the last row corresponds  to \eqref{n31} to \eqref{n32}.
	
	We prove that the rank of matrix \(\mathbf{A}\) is 10 and the rank of matrix \(\mathbf{B}\) is 9 to show that matrix \(\mathbf{\Xi}\) has full column rank.	Matrix \(\mathbf{A}\) takes the form of an block upper triangular matrix:
	\begin{equation}
		\mathbf{A} = \begin{bmatrix} 
			\mathbf{I}_3 & \mathbf{0}_{3\times4} & -[\mathbf{p}_\times] \\ 
			\mathbf{0}_{10 \times 3} & \mathbf{\Lambda} & \mathbf{0}_{10 \times 3} \\
			\mathbf{0}_3 &\mathbf{0}_{3\times4} & -2\mathbf{P}_1 \\
			\mathbf{0}_3 &\mathbf{0}_{3\times4} & -2\mathbf{P}_2 
		\end{bmatrix}
	\end{equation}
	where
	\begin{equation}
		\mathbf{\Lambda} = \begin{bmatrix} \mathbf{\Lambda}_1 \\ \mathbf{\Lambda}_2 \\ \mathbf{\Lambda}_3  \\ \mathbf{S}\end{bmatrix}
	\end{equation}
	
	The matrix \(\mathbf{\Lambda}\) can be decomposed into two matrices by QR decomposition:
	\begin{equation}
	 \mathbf{\Lambda} = \mathbf{Q_{\Lambda}} \cdot \mathbf{R_{\Lambda}}
	 \label{qrd}
	\end{equation}
	where
	\begin{equation}
	\mathbf{R_{\Lambda}} = \begin{bmatrix}
		r_{11} & r_{12} & r_{13} & r_{14} \\
		0 & r_{22} & r_{23} & r_{24} \\
		0 & 0 & r_{33} & r_{34} \\
		0 & 0 & 0 & r_{44}  \\
		\mathbf{0}_{6\times 1} & \mathbf{0}_{6\times 1} & \mathbf{0}_{6\times 1} & \mathbf{0}_{6\times 1}
		\end{bmatrix}_{[10 \times 4 ]}
	\end{equation}
	where
	\begin{align}
	r_{11} &= 2 \sqrt{2} \sqrt{2 q_w^2+q_x^2+q_y^2+q_z^2} \\
	r_{22} &= 2 \sqrt{2} \sqrt{\frac{2 q_w^4+q_w^2 \left(4 q_x^2+3 \left(q_y^2+q_z^2\right)\right)+2 q_x^4+3 q_x^2 \left(q_y^2+q_z^2\right)+\left(q_y^2+q_z^2\right)^2}{2 q_w^2+q_x^2+q_y^2+q_z^2}} \\
	r_{33} &= 2 \sqrt{2} \sqrt{\frac{2 q_w^4+q_w^2 \left(4 q_x^2+4 q_y^2+3 q_z^2\right)+2 q_x^4+q_x^2 \left(4 q_y^2+3 q_z^2\right)+2 q_y^4+3 q_y^2 q_z^2+q_z^4}{2 q_w^2+2 q_x^2+q_y^2+q_z^2}} \\
	r_{44} &= \frac{4 \left(q_w^2+q_x^2+q_y^2+q_z^2\right)}{\sqrt{2 q_w^2+2 q_x^2+2 q_y^2+q_z^2}} 
	\end{align}
	Since \(q_w\), \(q_x\), \(q_y\), \(q_z\) are not equal to 0 simultaneously, hence
	\begin{equation}
	\mathbf{Rank}\left(\mathbf{R_{\Lambda}}\right) = 4
	\end{equation}
	And
	\begin{equation}
		\mathbf{Rank}\left(\mathbf{\Lambda}\right)= \mathbf{Rank}\left(\mathbf{R_{\Lambda}}\right) = 4 \label{rankoflambdaall}
	\end{equation}

%
%
	
	Regarding matrix
	\( \mathbf{P} = \begin{bmatrix} \mathbf{P}_1 \\ \mathbf{P}_2\end{bmatrix} \),
	delete rows with all elements equal to 0 does not affect the
	rank. Clearly:
	
	\begin{equation}
		\mathbf{Rank}\left(\mathbf{P}\right) = \mathbf{Rank}\left(\mathbf{P}'\right) =  \mathbf{Rank}\left(\begin{bmatrix}  0 & 0 & 1 \\ 0 & -1&0 \\ 0 & 0 & -1 \\ 1 & 0&0 \end{bmatrix}\right) = 3  \label{rankofrhoall} \\
	\end{equation}
	
	Combining \(\eqref{rankoflambdaall}\) and \(\eqref{rankofrhoall}\) we
	can infer that \(\mathbf{Rank}\left(\mathbf{A}\right) = 10\).
	
	Matrix \(\mathbf{B}\) is also a block upper triangular matrix.
	\begin{equation}
		\mathbf{B} = \begin{bmatrix} 
			-\frac{1}{2}\mathbf{\Lambda}\left[\mathbf{q}\right]_{L(:,2:4)} & \mathbf{0}_{10\times3} & \mathbf{0}_{10\times3} \\ 
			\mathbf{0}_3 & \mathbf{I}_3 & -\mathbf{C} \\
			\mathbf{0}_{3} &\mathbf{0}_{3} & -\mathbf{C}\mathbf{P}_1 \\
			\mathbf{0}_{3} &\mathbf{0}_{3} & -\mathbf{C}\mathbf{P}_2 \\
		\end{bmatrix}
	\end{equation}
	where the \(\mathbf{\Lambda}\) is column full rank, thus:
	\begin{equation}
		\mathbf{Rank}\left(\mathbf{\Lambda} \left[\mathbf{q}\right]_{L(:,2:4)}\right) =  \mathbf{Rank}\left(\left[\mathbf{q}\right]_{L(:,2:4)}\right) = 3
	\end{equation}
	And, \(\begin{bmatrix}
		\mathbf{C}\mathbf{P}_1 \\
		\mathbf{C}\mathbf{P}_2
	\end{bmatrix}= \begin{bmatrix}
	\mathbf{C} & \mathbf{0}_3\\
	\mathbf{0}_3& \mathbf{C}
	\end{bmatrix} \begin{bmatrix} \mathbf{P}_1 \\ \mathbf{P}_2\end{bmatrix} \), then:
	\begin{align}
		\mathbf{Rank}\left(\begin{bmatrix}
			\mathbf{C}\mathbf{P}_1 \\
			\mathbf{C}\mathbf{P}_2
		\end{bmatrix}\right) = \mathbf{Rank}\left(\mathbf{P}\right) = 3
	\end{align}
	
	Hence, the rank of matrix \(\mathbf{B}\) is 9.
	
	The above results indicate that the rank of matrix \(\mathbf{\Xi}\) is 22, which establishes the observability of the system under general motion.

	\subsection*{Appendix E: Proof of Unobservable Directions Under Special Motion}\label{appendix-e}
	 
	\subsubsection*{Observability Matrix \label{observable-matrix}}
	As the definition in \cite{refobsm}, to obtain the linear observability matrix \(\mathbf{M}\), it is essential to compute the discrete-time error-state transition matrix \(\mathbf{\Phi}\) and observation matrix \(\mathbf{H}\). The null spaces of \(\mathbf{M}\) corresponds to the unobservable directions.
	\begin{equation}
		\mathbf{M}=\begin{bmatrix}
			\mathbf{H}_{1} \\
			\mathbf{H}_{2} \boldsymbol{\mathbf{\Phi}}_{2,1} \\
			\vdots \\
			\mathbf{H}_{k} \boldsymbol{\mathbf{\Phi}}_{k, 1}
		\end{bmatrix}
	\end{equation}
	where \(\mathbf{\Phi}_{k,1} = \mathbf{\Phi}_{k-1}\cdots\mathbf{\Phi}_{1}\), is the error-state transition matrix $\eqref{phi_1}$ from time-step 1 to k, and \(\mathbf{H}_k\) is the Jacobian of the measurement model $\eqref{H1}$ and $\eqref{H2}$. We readily obtain the linear observability matrix for the two types of measurements at time step \(t\), which are as follows:
	\begin{align}
		\mathbf{M}_{dp}(k ) &= \begin{bmatrix} \mathbf{\Phi}_{11} & \mathbf{\Phi}_{12} & \mathbf{\Phi}_{13} & \mathbf{\Phi}_{14} & \mathbf{\Phi}_{15} & \mathbf{\Phi}_{16} & \mathbf{\Phi}_{17}  \end{bmatrix} \label{Mp} \\
		\mathbf{M}_{dq}(k ) &= \mathbf{C}(\prescript{I^t_1}{}{\hat{\mathbf{q}}}_{I_2^t}) \begin{bmatrix} \mathbf{0}_3 &  \mathbf{0}_3  & \mathbf{\Phi}_{33} & \mathbf{\Phi}_{34} &  \mathbf{\Phi}_{35}  &  \mathbf{0}_3  &  \mathbf{0}_3   \label{Mq} \end{bmatrix}
	\end{align}	
	Here are some submatrices that will be utilized in the upcoming context, from Appendix C:
	\begin{align}
		\mathbf{\Phi}_{13}(t,t_0)
		=& -\mathbf{C}'(t) \int_{t_0}^{t} \int_{t_0}^{s} \mathbf{C}(\prescript{{I_1^{t_0}}}{}{\hat{\mathbf{q}}_{I_1^{\tau}}})[(\mathbf{C}(\prescript{{I_1^{\tau}}}{}{\hat{\mathbf{q}}}_{I_2^{\tau}})\prescript{{I^\tau_2}}{}{\hat{\mathbf{a}}})_\times] \mathbf{C}(\prescript{{I_1^{\tau}}}{}{\hat{\mathbf{q}}}_{I_2^{t_0}})  d\tau ds \\
		\mathbf{\Phi}_{14} (t,t_0)
		=& -\mathbf{C}'(t) \int_{t_0}^{t} \int_{t_0}^{s} \mathbf{C}(\prescript{{I_1^{t_0}}}{}{\hat{\mathbf{q}}_{I_1^{\tau}}}) [(\mathbf{C}(\prescript{{I_1^{\tau}}}{}{\hat{\mathbf{q}}}_{I_2^{\tau}})\prescript{{I^\tau_2}}{}{\hat{\mathbf{a}}})_\times] 
			\int_{t_0}^{\tau} \mathbf{C}(\prescript{{I_1^{\tau}}}{}{\hat{\mathbf{q}}}_{I_1^{\mu}}) d\mu d\tau ds \nonumber \\
		& - \mathbf{C}'(t) \int_{t_0}^{t} \int_{t_0}^{s} \mathbf{C}(\prescript{{I_1^{t_0}}}{}{\hat{\mathbf{q}}}_{I_1^{\tau}}) \mathbf{K}(\tau) d\tau ds \\
		\mathbf{\Phi}_{15} (t,t_0)
		&= \mathbf{C}'(t)\int_{t_0}^{t} \int_{t_0}^{s} \mathbf{C}(\prescript{{I_1^{t_0}}}{}{\hat{\mathbf{q}}}_{I_1^{\tau}}) [(\mathbf{C}(\prescript{{I_1^{\tau}}}{}{\hat{\mathbf{q}}}_{I_2^{\tau}})\prescript{{I^\tau_2}}{}{\hat{\mathbf{a}}})_\times] \int_{t_0}^{\tau} \mathbf{C}(\prescript{{I_1^{\tau}}}{}{\hat{\mathbf{q}}}_{I_2^{\mu}}) d\mu  d\tau ds  \\
		\mathbf{\Phi}_{16}(t,t_0)
		&=  \mathbf{C}'(t)\int_{t_0}^{t} \int_{t_0}^{s} \mathbf{C}(\prescript{{I_1^{t_0}}}{}{\hat{\mathbf{q}}}_{I_1^{\tau}}) d\tau ds \\
		\mathbf{\Phi}_{17} (t,t_0)
&=  -\mathbf{C}'(t)\int_{t_0}^{t} \int_{t_0}^{s} \mathbf{C}( \prescript{{I_1^{t_0}}}{}{\hat{\mathbf{q}}}_{I_2^{\tau}}) d\tau ds \\
		\mathbf{\Phi}_{34}(t,t_0) &= \mathbf{C}''(t)\int_{t_0}^{t} \mathbf{C}(\prescript{ {I_2^{t_0}}}{}{\hat{\mathbf{q}}}_{I_1^{s}}) ds \\
		\mathbf{\Phi}_{35}(t,t_0) &= -\mathbf{C}''(t)\int_{t_0}^{t} \mathbf{C}( \prescript{{I_2^{t_0}}}{}{\hat{\mathbf{q}}}_{I_2^{s}}) ds
	\end{align}
	where \(\mathbf{K}\) same as
	\(\eqref{FF}\), \(\mathbf{C}'(t) = \mathbf{C}(\prescript{{I_1^{t}}}{}{\hat{\mathbf{q}}}_{I_1^{t_0}})\)
	and 
	\(\mathbf{C}''(t) = \mathbf{C}(\prescript{{I_2^{t}}}{}{\hat{\mathbf{q}}}_{I_2^{t_0}})\).

	\vspace{0.4cm}
	\subsubsection*{1) Unobservable Directions with \(\mathbf{dp}\) and \(\mathbf{dq}\) Measurements \label{observable-matrix-dpq}}~{}
	
	Linear observability matrix at time step \(t\) is composed of the combination of \(\eqref{Mp}\) and \(\eqref{Mq}\):
\begin{equation}
	\mathbf{M}_{dpdq}(k ) = \begin{bmatrix} \mathbf{M}_{dp}(k) \\ \mathbf{M}_{dq}(k) 
	\end{bmatrix} 
\end{equation}
	The proofs for each unobservable direction are as follows: 
\begin{itemize}	
	\item
	The unobservable direction \( {\mathbf{b}_{a+}} \):
	
	\begin{itemize}
	\item[-] Sufficient Conditions:  \(^{I_1}\boldsymbol{\omega}_{I_2} = \mathbf{0}_{3\times1} \).
	\item[-] Proof: 
	\begin{align}
		\mathbf{M}_{dq}{\mathbf{b}_{a+}} =&~\mathbf{0}_{3} 
	\end{align}
	and
	\begin{align}
		\mathbf{M}_{dp}{\mathbf{b}_{a+}} =&~\mathbf{C}'(t)\int_{t_0}^{t} \int_{t_0}^{s} \mathbf{C}(\prescript{{I_1^{t_0}}}{}{\hat{\mathbf{q}}}_{I_1^{\tau}})(\mathbf{I}_3 -  \mathbf{C}(\prescript{I^\tau_1}{}{\hat{\mathbf{q}}}_{I^\tau_2})  \mathbf{C}^T(\prescript{I^{t_0}_1}{}{\hat{\mathbf{q}}}_{I^{t_0}_2})  ) d\tau ds 
	\end{align}
	Since \(^{I_1}\boldsymbol{\omega}_{I_2} = \mathbf{0}_{3\times1}\), \(\prescript{I^t_1}{}{\dot{\mathbf{q}}}_{I^t_2}= \frac{1}{2} \prescript{I^t_1}{}{\bar{\boldsymbol{\omega}}}_{I^t_2} \otimes \prescript{I^t_1}{}{\mathbf{q}}_{I^t_2} \equiv \mathbf{0}_{4\times1} \), hence, \(\prescript{I^\tau_1}{}{\hat{\mathbf{q}}}_{I^\tau_2}  \equiv  \prescript{I^{t_0}_1}{}{\hat{\mathbf{q}}}_{I^{t_0}_2}  \), where  \(^{I_1}\boldsymbol{\omega}_{I_2} =  \mathbf{C}(\prescript{I_1}{}{\mathbf{q}}_{I_2})\prescript{I_2}{}{\boldsymbol{\omega}} -  \prescript{I_1}{}{{\boldsymbol{\omega}}} \), and:
	\begin{align}
		\prescript{I_1}{}{\dot{\mathbf{q}}}_{I_2} =&~\frac{1}{2}(\prescript{I_1}{}{\mathbf{q}}_{I_2} \otimes \prescript{I_2}{}{\bar{\boldsymbol{\omega}}} -\prescript{I_1}{}{\bar{\boldsymbol{\omega}}} \otimes \prescript{I_1}{}{\mathbf{q}}_{I_2}) \nonumber \\
		=&~\frac{1}{2}(\prescript{I_1}{}{\mathbf{q}}_{I_2} \otimes \prescript{I_2}{}{\bar{\boldsymbol{\omega}}} \otimes  \prescript{I_1}{}{\mathbf{q}}^{-1}_{I_2} -\prescript{I_1}{}{\bar{\boldsymbol{\omega}}}) \otimes \prescript{I_1}{}{\mathbf{q}}_{I_2} \nonumber \\
		=&~\frac{1}{2}(\begin{bmatrix}
			0 \\ \mathbf{C}(\prescript{I_1}{}{\mathbf{q}}_{I_2})\prescript{I_2}{}{\boldsymbol{\omega}} 
		\end{bmatrix} -\prescript{I_1}{}{\bar{\boldsymbol{\omega}}} ) \otimes \prescript{I_1}{}{\mathbf{q}}_{I_2} \nonumber \\
		=&~\frac{1}{2} \prescript{I_1}{}{\bar{\boldsymbol{\omega}}}_{I_2} \otimes \prescript{I_1}{}{\mathbf{q}}_{I_2} \label{wq12}
	\end{align}
	Then:
	\begin{align}
		\mathbf{M}_{dp}{\mathbf{b}_{a+}} =&~\mathbf{C}'(t)\int_{t_0}^{t} \int_{t_0}^{s} \mathbf{C}(\prescript{{I_1^{t_0}}}{}{\hat{\mathbf{q}}}_{I_1^{\tau}})(\mathbf{I}_3 -\mathbf{I}_3 ) d\tau ds = \mathbf{0}_3
	\end{align}
	\end{itemize}

	\item 
	The unobservable direction \( {\mathbf{b}_{a+}^{\boldsymbol{\omega}}}\)

	\begin{itemize}
		\item[-] Sufficient Conditions: The relative angular velocity direction between the two IMUs remains constant, i.e., \(\prescript{I_1^{t}}{}{\boldsymbol{\omega}}_{I^{t}_2} = k(t) \cdot \boldsymbol{\xi}  \), where, \(\boldsymbol{\xi}\) is unit vector of constant direction of \(\prescript{I_1}{}{\boldsymbol{\omega}}_{I_2}\).
		\item[-] Proof:
		\begin{equation}
			\mathbf{b}_{a+}^{\boldsymbol{\omega}}  = \mathbf{b}_{a+} \cdot \prescript{{I_1}}{}{\boldsymbol{\omega}}_{I_2} 
		\end{equation}
		\begin{align}
			\mathbf{M}_{dq}\mathbf{b}_{a+} \cdot \prescript{{I_1}}{}{\boldsymbol{\omega}}_{I_2} =&~\mathbf{0}_{3\times1} 
		\end{align}	
		and
		\begin{align}
			\mathbf{M}_{dp}{\mathbf{b}_{a+}} =&~\mathbf{C}'(t)\int_{t_0}^{t} \int_{t_0}^{s} \mathbf{C}(\prescript{{I_1^{t_0}}}{}{\hat{\mathbf{q}}}_{I_1^{\tau}})(\prescript{{I_1}}{}{\boldsymbol{\omega}}_{I_2}  -  \mathbf{C}(\prescript{I^\tau_1}{}{\hat{\mathbf{q}}}_{I^\tau_2})  \mathbf{C}^T(\prescript{I^{t_0}_1}{}{\hat{\mathbf{q}}}_{I^{t_0}_2}) \prescript{{I_1}}{}{\boldsymbol{\omega}}_{I_2} ) d\tau ds 
		\end{align}
		Since \(\prescript{I_1^{t}}{}{\boldsymbol{\omega}}_{I^{t}_2} = k(t) \cdot \boldsymbol{\xi} \), and \(\prescript{I^t_1}{}{\dot{\mathbf{q}}}_{I^t_2}= \frac{1}{2} \prescript{I^t_1}{}{\bar{\boldsymbol{\omega}}}_{I^t_2} \otimes \prescript{I^t_1}{}{\mathbf{q}}_{I^t_2} \) (using \eqref{wq12}), hence:
		\begin{align}
			\mathbf{C}(\prescript{I^t_1}{}{\mathbf{q}}_{I^t_2}) =&~\left(\mathbf{I}_3 + \sin(\int_{t_0}^{t}  k(s) ds ) [\boldsymbol{\xi}_\times] + (1-\cos(\int_{t_0}^{t}  k(s) ds ))[\boldsymbol{\xi}_\times]^2\right)\mathbf{C}(\prescript{I^{t_0}_1}{}{\mathbf{q}}_{I^{t_0}_2}) \label{dirc1}
		\end{align}
		\begin{align}
			\mathbf{C}(\prescript{I^t_1}{}{\mathbf{q}}_{I^t_2}) \mathbf{C}^T(\prescript{I^{t_0}_1}{}{\hat{\mathbf{q}}}_{I^{t_0}_2}) \prescript{{I_1}}{}{\boldsymbol{\omega}}_{I_2}  =&~\prescript{{I_1}}{}{\boldsymbol{\omega}}_{I_2} \label{dirc2}
		\end{align}
		Then:
		\begin{align}
				\mathbf{M}_{dp}{\mathbf{b}_{a+}} =&~ \mathbf{C}'(t)\int_{t_0}^{t} \int_{t_0}^{s} \mathbf{C}(\prescript{{I_1^{t_0}}}{}{\hat{\mathbf{q}}}_{I_1^{\tau}})(\prescript{{I_1}}{}{\boldsymbol{\omega}}_{I_2} - \prescript{{I_1}}{}{\boldsymbol{\omega}}_{I_2}) d\tau ds = \mathbf{0}_{3\times1}
		\end{align}
	
	\end{itemize}

	\item 
	The unobservable direction \({\mathbf{b}_{g+}}\)

\begin{itemize}
	\item[-] Sufficient conditions:
\begin{enumerate}[itemindent=25pt]
	\item [c1.1]
	\(\prescript{I_1}{}{}{\boldsymbol{\omega}}_{I_2} = \mathbf{0}_{3 \times 1} \);
	\item [and c1.2]
	\(\prescript{I_1}{}{}{\mathbf{p}}_{I_2} = \mathbf{0}_{3 \times 1} \) or \( \prescript{I_1}{}{}{\boldsymbol{\omega}} = \mathbf{0}_{3 \times 1} \); 
	\item [and c1.3] 
	\(\prescript{I_1}{}{}{\mathbf{v}}_{I_2}= \mathbf{0}_{3 \times 1}\); 
\end{enumerate}
	\item[-]Proof:
	\begin{align}
		\mathbf{M}_{dq}{\mathbf{b}_{g+}} =&~ \mathbf{C}(\prescript{I^t_1}{}{\hat{\mathbf{q}}}_{I_2^t}) \mathbf{C}''(t)\int_{t_0}^{t} \mathbf{C}(\prescript{{I_2^{t_0}}}{}{\hat{\mathbf{q}}}_{I_1^{s}})(\mathbf{I}_3 - \mathbf{C}(\prescript{{I_1^{s}}}{}{\hat{\mathbf{q}}}_{I_2^{s}})\mathbf{C}^T(\prescript{I^{t_0}_1}{}{\hat{\mathbf{q}}}_{I^{t_0}_2})  ) ds 
	\end{align}
		Since c1.1, \(\mathbf{C}(\prescript{I^{s}_1}{}{\hat{\mathbf{q}}}_{I^{s}_2})\mathbf{C}(\prescript{I^{t_0}_1}{}{\hat{\mathbf{q}}}_{I^{t_0}_2}) = \mathbf{I}_3\), using \eqref{wq12}, then:
	\begin{align}
		\mathbf{M}_{dq}{\mathbf{b}_{g+}} =&~ \mathbf{C}(\prescript{I^t_1}{}{\hat{\mathbf{q}}}_{I_2^t}) \mathbf{C}''(t)\int_{t_0}^{t} \mathbf{C}(\prescript{{I_2^{t_0}}}{}{\hat{\mathbf{q}}}_{I_1^{s}})(\mathbf{I}_3 - \mathbf{I}_3  ) ds	= \mathbf{0}_3
	\end{align}
	Furthermore:
	\begin{align}	
		\mathbf{M}_{dp}{\mathbf{b}_{g+}} =&  -\mathbf{C}'(t) \int_{t_0}^{t_0} \int_{t_0}^{s} \mathbf{C}(\prescript{{I_1^{t_0}}}{}{\hat{\mathbf{q}}}_{I_1^{\tau}}) [(\mathbf{C}(\prescript{{I_1^{\tau}}}{}{\hat{\mathbf{q}}}_{I_2^{\tau}})\prescript{{I^\tau_2}}{}{\hat{\mathbf{a}}})_\times] \int_{t_0}^{\tau} \mathbf{C}(\prescript{{I_1^{\tau}}}{}{\hat{\mathbf{q}}}_{I_1^{\mu}}) (-\mathbf{I}_3 +\mathbf{C}(\prescript{I^{\mu}_1}{}{\hat{\mathbf{q}}}_{I^{\mu}_2})\mathbf{C}(\prescript{I^{t_0}_1}{}{\hat{\mathbf{q}}}_{I^{t_0}_2}) )  d\mu d\tau ds  \nonumber \\
		&- \mathbf{C}'(t) \int_{t_0}^{t} \int_{t_0}^{s} \mathbf{C}(\prescript{{I_1^{t_0}}}{}{\hat{\mathbf{q}}}_{I_1^{\tau}}) \mathbf{K}(\tau) d\tau ds 
	\end{align}
	Since c1.1, \(\mathbf{C}(\prescript{I^{\mu}_1}{}{\hat{\mathbf{q}}}_{I^{\mu}_2})\mathbf{C}(\prescript{I^{t_0}_1}{}{\hat{\mathbf{q}}}_{I^{t_0}_2}) = \mathbf{I}_3\). 
	
	Since c1.2 and c1.3, \(\mathbf{K} = 2[ \hat{\mathbf{v}}_\times ]+[ \hat{\boldsymbol{\omega}}_{1\times} ][\hat{\mathbf{p}}_\times] + [( \hat{\boldsymbol{\omega}}_1 \times \hat{\mathbf{p}}  )_\times] = \mathbf{0}_3\).
	
	Therefore:
	\begin{align}
		\mathbf{M}_{dp}{\mathbf{b}_{g+}} =& -\mathbf{C}'(t) \int_{t_0}^{t_0} \int_{t_0}^{s} \mathbf{C}(\prescript{{I_1^{t_0}}}{}{\hat{\mathbf{q}}}_{I_1^{\tau}}) [(\mathbf{C}(\prescript{{I_1^{\tau}}}{}{\hat{\mathbf{q}}}_{I_2^{\tau}})\prescript{{I^\tau_2}}{}{\hat{\mathbf{a}}})_\times] \int_{t_0}^{\tau} \mathbf{C}(\prescript{{I_1^{\tau}}}{}{\hat{\mathbf{q}}}_{I_1^{\mu}}) (-\mathbf{I}_3 +\mathbf{I}_3)  d\mu d\tau ds = \mathbf{0}_3
	\end{align}
	
\end{itemize}

\item 	
The unobservable direction \({\mathbf{b}_{g+}^{\boldsymbol{\omega}}} \)
	
\begin{itemize}
	\item[-] Sufficient Conditions:
\begin{enumerate}[itemindent=25pt]
	\item [c2.1]
	The relative angular velocity  \(\prescript{I_1}{}{}{\boldsymbol{\omega}}_{I_2} \) direction remains constant, i.e., \(\prescript{I_1^{t}}{}{\boldsymbol{\omega}}_{I^{t}_2} = k(t) \cdot \boldsymbol{\xi}  \), where, \(\boldsymbol{\xi}\) is unit vector of constant direction of \(\prescript{I_1}{}{\boldsymbol{\omega}}_{I_2}\);
	\item [and c2.2]
	\(\prescript{I_1}{}{}{\mathbf{p}}_{I_2} \parallel  \prescript{I_1}{}{}{\boldsymbol{\omega}} \) or  \(\prescript{I_1}{}{}{\mathbf{p}}_{I_2} = \mathbf{0}_{3 \times 1} \) or \( \prescript{I_1}{}{}{\boldsymbol{\omega}} = \mathbf{0}_{3 \times 1}\);
	\item [and c2.3]
	\(\prescript{I_1}{}{}{\mathbf{p}}_{I_2} \parallel  \prescript{I_1}{}{}{\boldsymbol{\omega}}_{I_2} \) or  \(\prescript{I_1}{}{}{\mathbf{p}}_{I_2} = \mathbf{0}_{3 \times 1} \) or \( \prescript{I_1}{}{}{\boldsymbol{\omega}} = \mathbf{0}_{3 \times 1}\);
	\item [and c2.4]
	\(\prescript{I_1}{}{}{\mathbf{v}}_{I_2} \parallel  \prescript{I_1}{}{}{\boldsymbol{\omega}}_{I_2} \) or  \(\prescript{I_1}{}{}{\mathbf{v}}_{I_2}= \mathbf{0}_{3 \times 1} \);
\end{enumerate}
	\item[-]Proof:
	\begin{align}
		{\mathbf{b}_{g+}^{\boldsymbol{\omega}}} =&~ \mathbf{b}_{g+} \cdot \prescript{{I_1}}{}{\boldsymbol{\omega}}_{I_2} \\
		\mathbf{M}_{dq}\mathbf{b}_{g+}\cdot \prescript{{I_1}}{}{\boldsymbol{\omega}}_{I_2} 
		=&~ \mathbf{C}(\prescript{I^t_1}{}{\hat{\mathbf{q}}}_{I_2^t})\mathbf{C}''(t)\int_{t_0}^{t} \mathbf{C} (t) (\prescript{{I_2^{t_0}}}{}{\hat{\mathbf{q}}}_{I_1^{s}})(\prescript{I_1}{}{\boldsymbol{\omega}}_{I_2} - \mathbf{C}(\prescript{{I_1^{s}}}{}{\hat{\mathbf{q}}}_{I_2^{s}})\mathbf{C}^T(\prescript{I^{t_0}_1}{}{\hat{\mathbf{q}}}_{I^{t_0}_2})\prescript{I_1}{}{\boldsymbol{\omega}}_{I_2} ) ds   
	\end{align}
	Since, c2.1 (same as \eqref{dirc1}, \eqref{dirc2}), hence:
	\begin{align}
	\mathbf{M}_{dq}\mathbf{b}_{g+}\cdot \prescript{{I_1}}{}{\boldsymbol{\omega}}_{I_2} =&~ \mathbf{0}_{3\times1}
	\end{align}
	Furthermore:	
	\begin{align}
		\mathbf{M}_{dp}\mathbf{b}_{g+}\cdot \prescript{{I_1}}{}{\boldsymbol{\omega}}_{I_2}  =&~  \mathbf{C}' (t)\int_{t_0}^{t_0} \int_{t_0}^{s} \mathbf{C}(\prescript{{I_1^{t_0}}}{}{\hat{\mathbf{q}}}_{I_1^{\tau}}) [\mathbf{C}(\prescript{{I_1^{\tau}}}{}{\hat{\mathbf{q}}}_{I_2^{\tau}}) \prescript{{I_2}}{}{\mathbf{a}} \times] \nonumber \\ 
		&\int_{t_0}^{\tau} \mathbf{C}(\prescript{{I_1^{\tau}}}{}{\hat{\mathbf{q}}}_{I_1^{\mu}})( -\prescript{I_1}{}{\boldsymbol{\omega}}_{I_2} +   \mathbf{C}(\prescript{{I_1^{\mu}}}{}{\hat{\mathbf{q}}}_{I_2^{\mu}}) \mathbf{C}^T(\prescript{{I_1^{t_0}}}{}{\hat{\mathbf{q}}}_{I_2^{t_0}})\prescript{I_1}{}{\boldsymbol{\omega}}_{I_2} )   d\mu d\tau ds   \nonumber\\
		& -\mathbf{C}' (t)\int_{t_0}^{t} \int_{t_0}^{s} \mathbf{C}(\prescript{{I_1^{t_0}}}{}{\hat{\mathbf{q}}}_{I_1^{\tau}}) \mathbf{K}(\tau)\cdot \prescript{{I_1}}{}{\boldsymbol{\omega}}_{I_2} d\tau ds 
	\end{align}
	Since c2.1 (same as \eqref{dirc1}, \eqref{dirc2}), \(  -\prescript{I_1}{}{\boldsymbol{\omega}}_{I_2} +   \mathbf{C}(\prescript{{I_1^{\mu}}}{}{\hat{\mathbf{q}}}_{I_2^{\mu}}) \mathbf{C}^T(\prescript{{I_1^{t_0}}}{}{\hat{\mathbf{q}}}_{I_2^{t_0}})\prescript{I_1}{}{\boldsymbol{\omega}}_{I_2} = \mathbf{0}_{3\times1} \)
	
	Since c2.2, c2.3, and c2.4, hence:
	\begin{align}
		\mathbf{K} \prescript{I_1}{}{\boldsymbol{\omega}}_{I_2} = 2[\prescript{{I_1}}{}{\hat{\mathbf{v}}}_{I_2\times} ]\cdot \prescript{{I_1}}{}{\boldsymbol{\omega}}_{I_2}  +[\prescript{{I_1}}{}{\hat{\boldsymbol{\omega}}}_\times ][\prescript{{I_1}}{}{\hat{\mathbf{p}}}_{I_2\times} ] \cdot \prescript{{I_1}}{}{\boldsymbol{\omega}}_{I_2}  + [( \prescript{{I_1}}{}{\hat{\boldsymbol{\omega}}} \times \prescript{{I_1}}{}{\hat{\mathbf{p}}}_{I_2}  )_\times] \cdot \prescript{{I_1}}{}{\boldsymbol{\omega}}_{I_2} = \mathbf{0}_{3\times1}
	\end{align}
	
	Then:
	\begin{align}
		\mathbf{M}_{dp}\mathbf{b}_{g+}\cdot \prescript{{I_1}}{}{\boldsymbol{\omega}}_{I_2}  =&~  \mathbf{0}_{3\times1}
	\end{align}

\end{itemize}

\item 
The unobservable direction \({\mathbf{b}_{g+}^{\boldsymbol{\alpha}}} \)
	
	\begin{itemize}
		\item[-] Sufficient conditions:
	\begin{enumerate}[itemindent=25pt]
	\item [c3.1]
	\(\prescript{I_1}{}{}{\boldsymbol{\omega}}_{I_2} \parallel  \boldsymbol{\alpha} \);
	\item [and c3.2]
	\(\prescript{I_1}{}{}{\mathbf{p}}_{I_2} \parallel  \prescript{I_1}{}{}{\boldsymbol{\omega}} \) or  \(\prescript{I_1}{}{}{\mathbf{p}}_{I_2} =  \mathbf{0}_{3\times1} \) or \( \prescript{I_1}{}{}{\boldsymbol{\omega}} =   \mathbf{0}_{3\times1} \);
	\item [and c3.3]
	\(\prescript{I_1}{}{}{\mathbf{p}}_{I_2} \parallel  \boldsymbol{\alpha}\) or  \(\prescript{I_1}{}{}{\mathbf{p}}_{I_2} =  \mathbf{0}_{3\times1} \) or \( \prescript{I_1}{}{}{\boldsymbol{\omega}} =   \mathbf{0}_{3\times1} \);
	\item [and c3.4]
	\(\prescript{I_1}{}{}{\mathbf{v}}_{I_2} \parallel  \boldsymbol{\alpha} \)  or  \(\prescript{I_1}{}{}{\mathbf{v}}_{I_2}= \mathbf{0}_{3 \times 1} \);
\end{enumerate}
	
	\item[-] Proof: This proof is exactly same as the proof of \({\mathbf{b}_{g+}^{\boldsymbol{\omega}}} \),  where we only need to change the symbol \(\prescript{I_1}{}{\boldsymbol{\omega}}_{I_2}\) to \(\boldsymbol{\alpha}\).
	\end{itemize}

\end{itemize}
	
	\vspace{0.4cm}	
	\subsubsection*{2) Additional Unobservable Directions with Only \(\mathbf{dp}\) Measurement \label{observable-matrix-dp}}~{}
	
	The linear observability matrix has only block \(\mathbf{M}_{dp}(k)\). The proofs for each unobservable direction are as follows: 
	
	\begin{itemize}	
	\item 
	The unobservable directions \(\boldsymbol{\theta}^{\boldsymbol{\alpha}} \) and \({\mathbf{b}_{g_1}^{\boldsymbol{\alpha}}}\)

	\begin{itemize}
		\item[-] Sufficient Conditions:
	\begin{enumerate}[itemindent=25pt]
	\item [c4.1]
	\( \prescript{I_1}{}{}{\mathbf{v}}_{I_2} \parallel \boldsymbol{\alpha} \) or  \( \prescript{{I_1}}{}{\mathbf{v}}_{I_2}  =  \mathbf{0}_{3 \times 1} \); 
	\item [and c4.2]
	\(\prescript{I_1}{}{}{\mathbf{a}} \parallel \boldsymbol{\alpha} \);
	\item [and c4.3]
	\(\prescript{I_1}{}{}{\boldsymbol{\omega}} \parallel \boldsymbol{\alpha} \) or \(\prescript{I_1}{}{}{\boldsymbol{\omega}} = \mathbf{0}_{3 \times 1} \);
	\item [and c4.4]
	\(\prescript{I_1}{}{}{\mathbf{p}}_{I_2} \parallel \prescript{I_1}{}{}{\boldsymbol{\omega}} \) or \(\prescript{I_1}{}{}{\mathbf{p}}_{I_2} = \mathbf{0}_{3 \times 1}\) or \(\prescript{I_1}{}{}{\boldsymbol{\omega}}  = \mathbf{0}_{3 \times 1} \);
\end{enumerate}
		\item[-] Proof:
		\begin{align}
			\mathbf{M}_{dp} \boldsymbol{\theta}^{\boldsymbol{\alpha}} =& -\mathbf{C}'(t) \int_{t_0}^{t} \int_{t_0}^{s} \mathbf{C}(\prescript{{I_1^{t_0}}}{}{\hat{\mathbf{q}}}_{I_1^{\tau}})[(\mathbf{C}(\prescript{{I_1^{\tau}}}{}{\hat{\mathbf{q}}}_{I_2^{\tau}})\prescript{{I^\tau_2}}{}{\hat{\mathbf{a}}})_\times] \mathbf{C}(\prescript{{I_1^{\tau}}}{}{\hat{\mathbf{q}}}_{I_2^{t_0}})  d\tau ds \cdot \mathbf{C}(\prescript{{I_2^{t_0}}}{}{\hat{\mathbf{q}}}_{I_1^{t_0}})\cdot \boldsymbol{\alpha}  \nonumber\\
			=&  -\mathbf{C}' (t)\int_{t_0}^{t} \int_{t_0}^{s} \mathbf{C}(\prescript{{I_1^{t_0}}}{}{\hat{\mathbf{q}}}_{I_1^{\tau}})[(\mathbf{C}(\prescript{{I_1^{\tau}}}{}{\hat{\mathbf{q}}}_{I_2^{\tau}})\prescript{{I^\tau_2}}{}{\hat{\mathbf{a}}})_\times]\mathbf{C}(\prescript{{I^{\tau}_1}}{}{\hat{\mathbf{q}}}_{I^{t_0}_1})
			 \boldsymbol{\alpha}   d\tau ds
		\end{align}
		Since c4.3, hence \(\mathbf{C}(\prescript{{I^{\tau}_1}}{}{\hat{\mathbf{q}}}_{I^{t_0}_1})
		\boldsymbol{\alpha} = \boldsymbol{\alpha}\), then:
		\begin{align}
			\mathbf{M}_{dp} \boldsymbol{\theta}^{\boldsymbol{\alpha}} = -\mathbf{C}' (t)\int_{t_0}^{t} \int_{t_0}^{s} \mathbf{C}(\prescript{{I_1^{t_0}}}{}{\hat{\mathbf{q}}}_{I_1^{\tau}})[(\mathbf{C}(\prescript{{I_1^{\tau}}}{}{\hat{\mathbf{q}}}_{I_2^{\tau}})\prescript{{I^\tau_2}}{}{\hat{\mathbf{a}}})_\times]
			\boldsymbol{\alpha}   d\tau ds
		\end{align}
		where, using \eqref{dotv}:
		\begin{align}
			[(\mathbf{C}(\prescript{{I_1}}{}{\hat{\mathbf{q}}}_{I_2})\prescript{{I_2}}{}{\mathbf{a}})_ \times] \boldsymbol{\alpha} 
			=&~ [(\prescript{{I_1}}{}{\dot{\hat{\mathbf{v}}}}_{I_2} +\prescript{{I_1}}{}{\hat{\mathbf{a}}}
			+2\prescript{{I_1}}{}{\hat{\boldsymbol{\omega}}}\times\prescript{{I_1}}{}{\hat{\mathbf{v}}}_{I_2}
			+ \prescript{{I_1}}{}{\hat{\boldsymbol{\omega}}}  \times(\prescript{{I_1}}{}{\hat{\boldsymbol{\omega}}} \times \prescript{{I_1}}{}{\hat{\mathbf{p}}}_{I_2}) + \prescript{{I_1}}{}{\dot{\hat{\boldsymbol{\omega}}}} \times \prescript{{I_1}}{}{\hat{\mathbf{p}}}_{I_2})_\times ]\boldsymbol{\alpha}
		\end{align}
		Since c4.1, hence:
		\begin{align}
			\prescript{{I_1}}{}{\dot{\mathbf{v}}}_{I_2} \times \boldsymbol{\alpha} = \mathbf{0}_{3\times1}
		\end{align}
		Since c4.2, hence:
		\begin{align}
			\prescript{{I_1}}{}{\hat{\mathbf{a}}} \times \boldsymbol{\alpha} = \mathbf{0}_{3\times1}
		\end{align}
		Since c4.4, hence:
		\begin{align}
			2\prescript{{I_1}}{}{\hat{\boldsymbol{\omega}}}\times\prescript{{I_1}}{}{\hat{\mathbf{v}}}_{I_2} + \prescript{{I_1}}{}{\hat{\boldsymbol{\omega}}}  \times(\prescript{{I_1}}{}{\hat{\boldsymbol{\omega}}} \times \prescript{{I_1}}{}{\hat{\mathbf{p}}}_{I_2}) + \prescript{{I_1}}{}{\dot{\hat{\boldsymbol{\omega}}}} \times \prescript{{I_1}}{}{\hat{\mathbf{p}}}_{I_2} = \mathbf{0}_{3\times1}
		\end{align}
		Therefore:
		\begin{align}
			[(\mathbf{C}(\prescript{{I_1}}{}{\hat{\mathbf{q}}}_{I_2})\prescript{{I_2}}{}{\mathbf{a}})_ \times] \boldsymbol{\alpha} = \mathbf{0}_{3\times1} \label{ca2alpha}
		\end{align}
		Then:
		\begin{align}
			\mathbf{M}_{dp} \boldsymbol{\theta}^{\boldsymbol{\alpha}} = \mathbf{0}_{3\times1}
		\end{align}
		
		Furthermore:	
		\begin{align}
			\mathbf{M}_{dp} \mathbf{b}_{g_1}^{\boldsymbol{\alpha}}
			=& -\mathbf{C}' (t)\int_{t_0}^{t_0} \int_{t_0}^{s} \mathbf{C}(\prescript{{I_1^{t_0}}}{}{\hat{\mathbf{q}}}_{I_1^{\tau}}) [(\mathbf{C}(\prescript{{I_1^{\tau}}}{}{{\hat{\mathbf{q}}}}_{I_2^{\tau}})\prescript{{I^\tau_2}}{}{\hat{\mathbf{a}}})_\times]
			\int_{t_0}^{\tau} \mathbf{C}(\prescript{{I_1^{\tau}}}{}{\hat{\mathbf{q}}}_{I_1^{\mu}}) \cdot \boldsymbol{\alpha} d\mu d\tau ds  \nonumber \\
			& -\mathbf{C}' (t)\int_{t_0}^{t} \int_{t_0}^{s} \mathbf{C}( \prescript{{I_1^{t_0}}}{}{\hat{\mathbf{q}}}_{I_1^{\tau}}) \mathbf{K}(\tau) \cdot \boldsymbol{\alpha} d\tau ds 
		\end{align}
		Since c4.3, hence \(\int_{t_0}^{\tau}\mathbf{C}(\prescript{{I_1^{\tau}}}{}{\hat{\mathbf{q}}}_{I_1^{\mu}}) \cdot \boldsymbol{\alpha} d\mu = \boldsymbol{\alpha} \tau \).
		And since c4.1, c4.2, and c4.4 (same as \eqref{ca2alpha}):
		\begin{align}
			-\mathbf{C}' (t)\int_{t_0}^{t_0} \int_{t_0}^{s} \mathbf{C}(\prescript{{I_1^{t_0}}}{}{\hat{\mathbf{q}}}_{I_1^{\tau}}) [(\mathbf{C}(\prescript{{I_1^{\tau}}}{}{{\hat{\mathbf{q}}}}_{I_2^{\tau}})\prescript{{I^\tau_2}}{}{\hat{\mathbf{a}}})_\times]
			\int_{t_0}^{\tau} \mathbf{C}(\prescript{{I_1^{\tau}}}{}{\hat{\mathbf{q}}}_{I_1^{\mu}}) \cdot \boldsymbol{\alpha} d\mu d\tau ds = \mathbf{0}_{3\times1}	
		\end{align}
		For \(\mathbf{K}\boldsymbol{\alpha}\):
		\begin{align}
			\mathbf{K}\boldsymbol{\alpha} =  2[\prescript{{I_1}}{}{\hat{\mathbf{v}}}_{I_2\times} ]\cdot \boldsymbol{\alpha}  +[\prescript{{I_1}}{}{\hat{\boldsymbol{\omega}}}_\times ][\prescript{{I_1}}{}{\hat{\mathbf{p}}}_{I_2\times} ] \cdot \boldsymbol{\alpha}  + [( \prescript{{I_1}}{}{\hat{\boldsymbol{\omega}}} \times \prescript{{I_1}}{}{\hat{\mathbf{p}}}_{I_2}  )_\times] \cdot \boldsymbol{\alpha}
		\end{align}
		Since c4.1, \([\prescript{{I_1}}{}{\hat{\mathbf{v}}}_{I_2\times} ]\cdot \boldsymbol{\alpha} = \mathbf{0}_{3\times1}\). 
		
		Since c4.3 if \(\prescript{I_1}{}{\boldsymbol{\omega}} = \mathbf{0}_{3\times1}\), hence \( [\prescript{{I_1}}{}{\hat{\boldsymbol{\omega}}}_\times ][\prescript{{I_1}}{}{\hat{\mathbf{p}}}_{I_2\times} ] =\mathbf{0}_3\) and \([( \prescript{{I_1}}{}{\hat{\boldsymbol{\omega}}} \times \prescript{{I_1}}{}{\hat{\mathbf{p}}}_{I_2}  )_\times] =\mathbf{0}_3 \), or if \( \prescript{I_1}{}{}{\boldsymbol{\omega}} \parallel \boldsymbol{\alpha} \) combining c4.4, \([( \prescript{{I_1}}{}{\hat{\boldsymbol{\omega}}} \times \prescript{{I_1}}{}{\hat{\mathbf{p}}}_{I_2}  )_\times] \boldsymbol{\alpha} =\mathbf{0}_{3\times1}\),   \( [\prescript{{I_1}}{}{\hat{\boldsymbol{\omega}}}_\times ][\prescript{{I_1}}{}{\hat{\mathbf{p}}}_{I_2\times}] \boldsymbol{\alpha} =\mathbf{0}_{3\times1}\).
		
		Therefore:
		\begin{align}
			\mathbf{K}\boldsymbol{\alpha} = \mathbf{0}_{3\times1}
		\end{align}
		
		Then:
		\begin{align}
			\mathbf{M}_{dp} \mathbf{b}_{g_1}^{\boldsymbol{\alpha}} = \mathbf{0}_{3\times1}
		\end{align}
	\end{itemize}

	\item 	
	The unobservable direction \(\boldsymbol{\theta}^{\boldsymbol{\beta}} \)
	\begin{itemize}
		\item[-] Sufficient Conditions: 
			\begin{enumerate}[itemindent=25pt]
	\item [c5.1]
		\(\prescript{I_1}{}{\mathbf{a}} = \boldsymbol{\alpha}\);
	\item [and c5.2]
		\(\prescript{I_1}{}{\boldsymbol{\omega}} = \mathbf{0}_{3\times1}\);
	\item [and c5.3]
		\(\prescript{I_1}{}{\mathbf{v}}_{I_2} = \mathbf{0}_{3\times1}\);
	\end{enumerate}
		\item[-]Proof:
		\begin{align}
			\mathbf{M}_{dp} \begin{bmatrix}\boldsymbol{\theta}^{\boldsymbol{\beta}_1}  & \boldsymbol{\theta}^{\boldsymbol{\beta}_2} \end{bmatrix} =& -\mathbf{C}'(t) \int_{t_0}^{t} \int_{t_0}^{s} \mathbf{C}(\prescript{I_1^{t_0}}{}{\hat{\mathbf{q}}}_{I_1^{\tau}})[(\mathbf{C}(\prescript{{I_1^{\tau}}}{}{\hat{\mathbf{q}}}_{I_2^{\tau}})\prescript{{I^\tau_2}}{}{\hat{\mathbf{a}}})_\times] \mathbf{C}(\prescript{I_1^{\tau}}{}{\hat{\mathbf{q}}}_{I_2^{t_0}})  d\tau ds \mathbf{C}(\prescript{I_2^{t_0}}{}{\hat{\mathbf{q}}}_{I_1^{t_0}})\begin{bmatrix}
				\boldsymbol{\beta}_1 & \boldsymbol{\beta}_2
			\end{bmatrix}  \nonumber\\
			& + \mathbf{C}'(t) \int_{t_0}^{t} \int_{t_0}^{s} \mathbf{C}( \prescript{I_1^{t_0}}{}{\hat{\mathbf{q}}}_{I_1^{\tau}}) d\tau ds \begin{bmatrix}
			\boldsymbol{\alpha} \times	\boldsymbol{\beta}_1 & \boldsymbol{\alpha} \times \boldsymbol{\beta}_2
			\end{bmatrix} \nonumber \\
			=&~\mathbf{C}'(t) \int_{t_0}^{t} \int_{t_0}^{s} \mathbf{C}( \prescript{I_1^{t_0}}{}{\hat{\mathbf{q}}}_{I_1^{\tau}}) ( -[(\mathbf{C}(\prescript{{I_1^{\tau}}}{}{\hat{\mathbf{q}}}_{I_2^{\tau}})\prescript{{I^\tau_2}}{}{\hat{\mathbf{a}}})_\times] \mathbf{C}(\prescript{{I^{\tau}_1}}{}{\hat{\mathbf{q}}}_{I^{t_0}_1}) \begin{bmatrix}
			\boldsymbol{\beta}_1 & \boldsymbol{\beta}_2
			\end{bmatrix}  + \begin{bmatrix}
			\boldsymbol{\alpha} \times	\boldsymbol{\beta}_1 & \boldsymbol{\alpha} \times \boldsymbol{\beta}_2
			\end{bmatrix}  ) d\tau ds 
		\end{align}		
	
		Since c5.1, c5.2, and c5.3, hence, \(\mathbf{C}(\prescript{{I_1}}{}{\hat{\mathbf{q}}}_{I_2}) \prescript{{I_2}}{}{\mathbf{a}} = \prescript{I_1}{}{\mathbf{a}} = \boldsymbol{\alpha}\) (using \eqref{dotv}). And,  since c5.2, \(\mathbf{C}(\prescript{{I^{t}_1}}{}{\hat{\mathbf{q}}}_{I^{t_0}_1}) = \mathbf{I}_3 \). Therefore:
		\begin{align}
			\mathbf{M}_{dp} \begin{bmatrix}\boldsymbol{\theta}^{\boldsymbol{\beta}_1}  & \boldsymbol{\theta}^{\boldsymbol{\beta}_2} \end{bmatrix} =&~ \mathbf{C}'(t) \int_{t_0}^{t} \int_{t_0}^{s} \mathbf{C}( \prescript{I_1^{t_0}}{}{\hat{\mathbf{q}}}_{I_1^{\tau}}) ( -\boldsymbol{\alpha} \times \begin{bmatrix}
				\boldsymbol{\beta}_1 & \boldsymbol{\beta}_2
			\end{bmatrix}  + \begin{bmatrix}
				\boldsymbol{\alpha} \times	\boldsymbol{\beta}_1 & \boldsymbol{\alpha} \times \boldsymbol{\beta}_2
			\end{bmatrix}  ) d\tau ds = \mathbf{0}_{3\times2}
		\end{align}
		 
	\end{itemize}

\item 
	The unobservable direction \(\boldsymbol{\theta}^{\boldsymbol{\beta}_1} \)
	
		\begin{itemize}
		\item[-] Sufficient Conditions:
	\begin{enumerate}[itemindent=25pt]
	\item [c6.1]
	\(\prescript{I_1}{}{}{\mathbf{v}} \parallel \boldsymbol{\beta}_1\) and \(\boldsymbol{\beta}_1 \nparallel \boldsymbol{\alpha} \);
	\item [and c6.2]
	\(\prescript{I_1}{}{}{\boldsymbol{\omega}}  = \mathbf{0}_{3\times1}\);
	\item [and c6.3]
	\(\prescript{I_1}{}{}{\mathbf{v}}_{I_2} =  \mathbf{0}_{3\times1} \);
\end{enumerate}
		\item[-]Proof:
			\begin{align}
			\mathbf{M}_{dp} \boldsymbol{\theta}^{\boldsymbol{\beta}_1} =& -\mathbf{C}' (t)\int_{t_0}^{t} \int_{t_0}^{s} \mathbf{C}(\prescript{I_1^{t_0}}{}{\hat{\mathbf{q}}}_{I_1^{\tau}})[(\mathbf{C}(\prescript{{I_1^{\tau}}}{}{\hat{\mathbf{q}}}_{I_2^{\tau}})\prescript{{I^\tau_2}}{}{\hat{\mathbf{a}}})_\times] \mathbf{C}(\prescript{I_1^{\tau}}{}{\hat{\mathbf{q}}}_{I_2^{t_0}})  d\tau ds \mathbf{C}(\prescript{I_2^{t_0}}{}{\hat{\mathbf{q}}}_{I_1^{t_0}})\boldsymbol{\beta}_1  \nonumber\\
			&+ \mathbf{C}'(t)\int_{t_0}^{t} \int_{t_0}^{s} \mathbf{C}( \prescript{I_1^{t_0}}{}{\hat{\mathbf{q}}}_{I_1^{\tau}}) d\tau ds \boldsymbol{\alpha} \times \boldsymbol{\beta}_1 \nonumber \\
			=&~\mathbf{C}' (t) \int_{t_0}^{t} \int_{t_0}^{s} \mathbf{C}(\prescript{I_1^{t_0}}{}{\hat{\mathbf{q}}}_{I_1^{\tau}}) ( -[(\mathbf{C}(\prescript{{I_1^{\tau}}}{}{\hat{\mathbf{q}}}_{I_2^{\tau}})\prescript{{I^\tau_2}}{}{\hat{\mathbf{a}}})_\times]\mathbf{C}(\prescript{{I^{\tau}_1}}{}{\hat{\mathbf{q}}}_{I^{t_0}_1}) \boldsymbol{\beta}_1 + \boldsymbol{\alpha} \times \boldsymbol{\beta}_1 ) d\tau ds 
		\end{align}	
		
		Since c6.1, \(^{I^{t}_1}\mathbf{a} = \boldsymbol{\alpha} + k(t) \boldsymbol{\beta}_1\), combining c6.2 and c6.3, \(\mathbf{C}(\prescript{{I_1}}{}{\hat{\mathbf{q}}}_{I_2}) \prescript{{I_2}}{}{\mathbf{a}} =  \prescript{I_1}{}{\mathbf{a}} = \boldsymbol{\alpha} + k(t) \boldsymbol{\beta}_1 \), and since c6.2, \(\mathbf{C}(\prescript{{I^{t}_1}}{}{\hat{\mathbf{q}}}_{I^{t_0}_1}) = \mathbf{I}_3 \), then,  \([(\mathbf{C}(\prescript{{I_1^{\tau}}}{}{\hat{\mathbf{q}}}_{I_2^{\tau}})\prescript{{I^\tau_2}}{}{\hat{\mathbf{a}}})_\times]\mathbf{C}(\prescript{{I^{\tau}_1}}{}{\hat{\mathbf{q}}}_{I^{t_0}_1}) \boldsymbol{\beta}_1 = \boldsymbol{\alpha} \times \boldsymbol{\beta}_1 + \mathbf{0}_{3\times1}\).

		Therefore:		
		\begin{align}
			\mathbf{M}_{dp} \boldsymbol{\theta}^{\boldsymbol{\beta}_1} =&~ \mathbf{C}' (t) \int_{t_0}^{t} \int_{t_0}^{s} \mathbf{C}(\prescript{I_1^{t_0}}{}{\hat{\mathbf{q}}}_{I_1^{\tau}}) ( -\boldsymbol{\alpha} \times \mathbf{C}(\prescript{{I^{\tau}_1}}{}{\hat{\mathbf{q}}}_{I^{t_0}_1}) \boldsymbol{\beta}_1 + \boldsymbol{\alpha} \times \boldsymbol{\beta}_1 ) d\tau ds = \mathbf{0}_{3\times1}
		\end{align}	
		\end{itemize}
		
	\end{itemize}

\end{document}